%% file: relational_transformation.tex
\documentclass{KERauth}
\usepackage[bookmarks=true]{hyperref}
\hypersetup{%
    bookmarks=false,    
    pdftitle={Transforming Graph Representations for Statistical Relational Learning},    
    pdfauthor={R. Rossi et al.},                     
    pdfsubject={Machine learning, statistical relational learning, relational representation transformation},                        
    pdfkeywords={Machine learning, artificial intelligence, data mining, graph transformations, relational representation transformation, link transformation, node transformation, link transformation, link prediction, link weighting, link labeling, link feature construction, node prediction, node weighting, node labeling, node feature construction, knowledge discovery}, 
}

\usepackage{theapa}

\usepackage{graphicx,epsfig}
\usepackage{graphics,dcolumn,bm,epic,eepic,float}
\usepackage{amssymb,amsmath,multirow,rotate,color}
\usepackage{subfigure}
\usepackage{comment}
\usepackage{multirow}
\usepackage{graphicx}
\usepackage{subfigure}
\usepackage{graphics}
\usepackage{longtable}
\usepackage{color}
\usepackage{lscape}
\usepackage{rotating}
\usepackage{tabularx}
\usepackage{booktabs}
\usepackage{paralist}

\usepackage{amssymb}   

\usepackage{color}
\definecolor{light-gray}{gray}{0.45}
\definecolor{invisible}{gray}{1}
\newcommand{\mycheck}{{$\checkmark$}}
\newcommand{\mylight}{{\color{light-gray} $\checkmark$}}

\newcommand{\qed}{\nobreak \ifvmode \relax \else
      \ifdim\lastskip<1.5em \hskip-\lastskip
      \hskip1.5em plus0em minus0.5em \fi \nobreak
      \vrule height0.75em width0.5em depth0.25em\fi}

\newcommand\T{\rule{0pt}{3.8ex}}
\newcommand\B{\rule[-1.8ex]{0pt}{0pt}}

\newcommand{\eg}{{e.g.}}
\newcommand{\ie}{{i.e.}}
\newcommand{\eat}[1]{}

\newcommand{\graph}{{$\tilde{G}$}}
\newcommand{\links}{{$\tilde{E}$}}
\newcommand{\nodes}{{$\tilde{V}$}}

\newcommand{\nfeatures}{{$\mathbf{\tilde{X}}^{V}$}}
\newcommand{\lfeatures}{{$\mathbf{\tilde{X}}^{E}$}}

\newcommand{\nfeature}{{$\tilde{X}^{V}_{k}$}}
\newcommand{\lfeature}{{$\tilde{X}^{E}_{k}$}}

\newcommand{\fullgraph}{{$\tilde{G} = \langle \tilde{V}, \tilde{E}, \mathbf{\tilde{X}}^{V}, \mathbf{\tilde{X}}^{E} \rangle$}}

\newcommand{\ifeatures}{{$\mathbf{X} = (\mathbf{X}^{E},\mathbf{X}^{V})$}}

\newcommand{\infeatures}{{$\mathbf{X}^{V}$}}
\newcommand{\ilfeatures}{{$\mathbf{X}^{E}$}}

\newcommand{\ifullgraph}{{$G = \langle V,E,\mathbf{X}^{V},\mathbf{X}^{E} \rangle$}}

\providecommand{\mat}[1]{\boldsymbol{\mathrm{#1}}}%
\renewcommand{\vec}[1]{\boldsymbol{\mathrm{#1}}}
\providecommand{\eye}{\mat{I}}
\providecommand{\mA}{\ensuremath{\mat{A}}}

\providecommand{\mD}{\ensuremath{\mat{D}}}

\providecommand{\mL}{\ensuremath{\mat{L}}}

\providecommand{\mP}{\ensuremath{\mat{P}}}

\begin{document}

\KER{1}{24}{00}{0}{2010}{S000000000000000}
\runningheads{R. Rossi, L. McDowell, D. Aha, \& J. Neville}
{Transforming Graph Representations for Statistical Relational Learning}

\title{Transforming Graph Representations\\for Statistical Relational Learning}

 \author{RYAN A. ROSSI\affilnum{1}, LUKE K. MCDOWELL\affilnum{2}, DAVID W. AHA\affilnum{3} and \\ JENNIFER NEVILLE\affilnum{1}}

\address{\affilnum{1}Department of Computer Science, Purdue University, West Lafayette, IN 47907, USA\\
\affilnum{2}Department of Computer Science, U.S. Naval Academy, Annapolis, MD 21402, USA\\
\affilnum{3}Navy Center for Applied Research in Artificial Intelligence,\\
 Naval Research Laboratory (Code 5514), Washington, DC 20375, USA\\
\email{\{rrossi, neville\}@cs.purdue.edu, lmcdowel@usna.edu, david.aha@nrl.navy.mil}}

\begin{abstract}
Relational data representations have become an increasingly important topic
 due to the recent proliferation of network datasets (e.g., social, biological, information networks) and a corresponding increase in
the application of statistical relational learning (SRL) algorithms to these domains. 
In this article, we examine a range of representation issues for graph-based relational data.
Since the choice of relational data representation---for the nodes, links, and
features---can dramatically affect the capabilities
of SRL algorithms, we survey approaches
and opportunities for \textit{relational representation transformation} designed to
improve the performance of these algorithms.
This leads us to introduce an intuitive taxonomy for data representation transformations in relational domains that incorporates \textit{link transformation} and \textit{node transformation} as symmetric representation tasks.
In particular, 
the transformation tasks for both nodes and links include (i) predicting their existence, (ii) predicting their label or type, (iii) estimating their weight or importance, and (iv) systematically constructing their relevant features.
We motivate our taxonomy through detailed examples and use it to survey and compare competing approaches for 
each of these tasks. We also discuss general conditions for transforming links, nodes, and
features. Finally, we highlight challenges that remain to be
addressed.
\end{abstract}

\input{introduction}

\input{overview}

\input{link-pred}
\input{link-interp}

\input{node-pred}

\input{node-interp}

\input{joint-disc}

\input{discussion}

\input{conclusion}

\eat{
\section*{Acknowledgements}
We thank the anonymous reviewers for helpful comments that improved the paper.
The majority of the research was completed at the Naval Research
Laboratory, where Ryan Rossi was supported by a Graduate NREIP
Fellowship. This research was also made with Government support under
and awarded by DoD, Air Force Office of Scientific Research, National
Defense Science and Engineering Graduate (NDSEG) Fellowship, 32 CFR
168a.
}

\small
\vskip 0.2in
\bibliography{rossi}
\bibliographystyle{theapa}

\end{document}

%% file: introduction.tex
\section{Introduction}
 The majority of research in machine learning assumes independently
 and identically distributed data. This independence assumption is
 often violated in relational data, which
 encode dependencies among data instances.
 For instance, people are often linked by business associations,
 and information about one person can be highly informative 
 for a prediction task involving an associate of that person.
 More generally, relational data can be described as a set of nodes, 
 which can be connected by one or more types of relations (or ``links'').
Relational information is seemingly ubiquitous; it is present in domains such as the Internet and the world-wide web~\shortcite{faloutsos1999power,broder2000graph,albert1999diameter}, scientific citation and collaboration~\shortcite{mcgovern:04,newman2001structure}, epidemiology~\shortcite{pastor2001epidemic,moore2000epidemics,may2001infection,kleczkowski1999mean} communication analysis~\shortcite{rossi:10}, metabolism~\shortcite{jeong2000large,wagner2001small}, ecosystems~\shortcite{dunne2002food,camacho2002robust}, bioinformatics~\shortcite{maslov2002specificity,jeong2001lethality}, fraud and terrorist analysis~\shortcite{neville2005using,krebs2002mapping}, and many others. The links in these data may represent citations, friendships, associations, metabolic functions, communications, co-locations, shared mechanisms, or many other explicit or implicit relationships.

 Statistical relational learning (SRL) 
 methods have been developed to address the problems of 
 reasoning and learning in domains with complex relations and
 probabilistic structure \shortcite{introSRL07}.
 In particular, SRL algorithms leverage relational information in an attempt to learn models with higher predictive accuracy. 
 A key characteristic of
 many relational datasets is a correlation or statistical
 dependence between the values of the same attribute across linked instances (e.g.,
 two friends are more likely to share political views than two
 randomly selected people). This \textit{relational autocorrelation}
 provides a unique opportunity to increase the accuracy of statistical
 inferences \shortcite{jensen:kdd04}. Similarly, relational information can
 be exploited for many other reasoning tasks such as identifying
 useful patterns or optimizing systems \shortcite{kleinberg10networks}.

Representation issues---including knowledge, model, and data representation---have been at the heart of the artificial
 intelligence community for decades \shortcite{Amarel1968rep,minsky74rep,RussellNorvig09}.
All of these are important, but here we focus on {\em data representation} issues, simple examples of which include the choices of whether to discretize continuous features
or to add higher-order polynomial features.
Such decisions can have a significant effect on the accuracy and efficiency of AI algorithms.
They are especially critical for the performance
 of SRL algorithms because, in relational domains, there is an even larger space of potential data representations to consider.  The complex structure of relational data can often be represented in a variety of ways and the choice of specific data representation can impact both the applicability of particular models/algorithms and their performance.
Specifically, there are two categories of decisions that need to be considered in the context of relational data representation.

First, we have to consider the {\em type} of data representation to use
(cf., the hierarchy of \shortciteR{de2008logical}, Chapter 4). For instance, relational data can be propositionalized
for the application of standard, non-relational learning algorithms.  More often, in order to fully exploit the relational information, SRL researchers have chosen to represent
the data either using an attributed graph in a relational database (see e.g.,~\shortciteR{Friedman99:PRMs}), or via logic programs (see e.g.,~\shortciteR{kersting:02}).\footnote{In the latter case, 
the applicable SRL algorithms are often referred to as probabilistic inductive logic programming (ILP) \shortcite{de2008probabilistic}.}
Each choice has different strengths.
In this article, we focus on the graph-based representation,
which has been a common choice for addressing the growing interest in network data and applications
analyzing electronic communication and online social networks such as Facebook, Twitter, Flickr, and LinkedIn~\shortcite{mislove2007measurement,ahmed2010time}. 
Specifically, we assume a graph-based data representation \ifullgraph ~where the nodes $V$ are entities (e.g., people, places, events) and the links $E$ represent relationships among those entities (e.g., friendships, citations).
\infeatures\ is a set of features about the entities in $V$. Likewise, the set of features \ilfeatures\ provides information about the relation links in $E$.

Next, given the type of representation, 
we must consider the specific {\em content} of the data representation, for which there is a large space of choices.
For instance, features for the nodes and links of a graph can be constructed using a wide range of aggregation functions, based on multiple kinds of links and
paths. SRL researchers have already recognized the importance of such data representation choices (e.g., \shortciteR{getoor:expl05}), and many separate studies have examined techniques for feature construction \shortcite{neville:kdd03}, 
node weighting \shortcite{tang2009temporal}, link prediction \shortcite{Taskar03linkprediction}, etc.  
However, this article is the first to comprehensively survey and assess approaches to relational representation transformation for graph-based data.

Given a set of (graph-based) relational data, we define \textit{relational representation transformation} as any change to the space of links,
nodes, and/or features used to represent the data.  Typically, the goal of this transformation is to improve the
performance of some subsequent SRL application.  For instance, in Figure~\ref{fig:rep-trans} the original graph
representation $G$ is transformed into a new representation $\tilde{G}$ where links, nodes, and features (such as link
weights) have been added, and some links have been removed.  Some SRL algorithm or analysis is then applied to the new
representation, for instance to classify the nodes or to identify anomalous links.  The particular transformations that
are used to produce $\tilde{G}$ will vary depending upon the intended application, but can sometimes substantially
improve the accuracy, speed, or complexity of the final application.  For instance, \shortciteA{gallagher:08} found that
adding links between similar nodes could increase node classification accuracy by up to 15\% on some tasks.  Similarly,
\shortciteA{neville:icdm05} demonstrated that adding nodes which represent underlying groups enabled both simpler inference
and increased accuracy.

\subsection{Scope of this Article}
\label{sec:scope}

\begin{figure}[t]
\centering
\includegraphics[width=\linewidth]{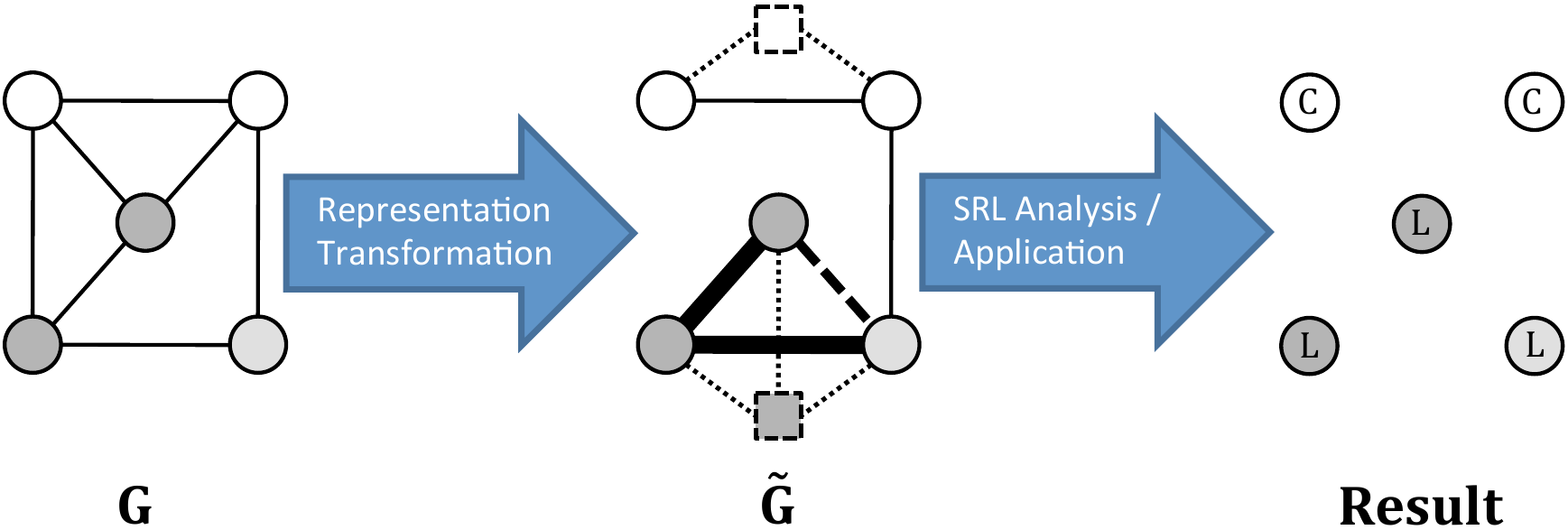} 
\caption{\textsc{Example Transformation and Subsequent Analysis:} 
The original relational representation $G$ is transformed into $\tilde{G}$ where dotted lines represent predicted links, squares represent predicted nodes, and bold links represent link weighting.  Changes may be based on link structure, link features, and node features (here, similar node shadings indicate similar feature values). Some SRL analysis is then applied to the new representation. In this example, the SRL analysis produces a label ($C$ or $L$) for each node, as with the example task discussed in Section~\ref{sec:motivating-example}. This article focuses on the representation transformation (left side of the figure), not the subsequent analysis.}
\label{fig:rep-trans}
\end{figure}

This article focuses on examining and categorizing various techniques for changing the
representation of graph-based relational data.  
As shown in Figure~\ref{fig:rep-trans}, we typically view these 
changes as a pre-processing step that enables increased
accuracy or speed for some other task, such as object classification.
However, an output of these techniques can itself be valuable.
For instance, the administrators of a social network may be interested in link
prediction so that predicted links can be presented to their users as potential new ``friendship'' links. 
Alternatively, these techniques may also be applied to improve the comprehensibility of a model.
For example, the prediction of protein-protein interactions provides insights into protein function
\shortcite{ben2005kernel}.  
Thus, the techniques we survey may be used for multiple purposes, and relevant publications
may have used them in different contexts.
Regardless of the original context, we will examine the general applicability and benefits
of each technique. 
After such techniques have been applied, the
transformed data can be used as is (e.g., for friendship suggestions), 
examined for greater understanding, used for some other task (e.g.,
for object classification), or used recursively as the input for another representation change
(e.g., as in object/node prediction followed by link prediction).

We do {\em not} attempt to survey the many methods that could be used for SRL analysis (e.g., the right side of
Figure~\ref{fig:rep-trans}), although the relevant set of methods for such analysis overlaps with the set
of methods that facilitate the transformations we consider.  For instance, collective classification \shortcite{neville:srl00,taskardpm:02} is an important SRL application
that we define in Section~\ref{sec:overview} and use as a running example of an SRL analysis task.
The output of such classification could also be used to create new attributes for the nodes (a data representation
change).  We discuss this possibility in Section~\ref{sec:node-label}, but
focus on a few cases where such node labeling is particularly useful as a pre-processing step
(e.g., before applying certain ``stacked'' algorithms), rather than surveying the wide range of possible classification algorithms, whether
collective or not.  
Likewise, we do not survey issues in model and knowledge representation, such as whether the statistical dependencies 
between nodes, links, and features should be modeled with Structural Logistic Regression \shortcite{Popescul03SLR}
or with a Markov Logic Network \shortcite{Domingos04markovlogic}.
We consider such issues only briefly, in Section~\ref{sec:modelrep}.

Furthermore, we focus on transformations to graph-based data that modify the set of links or nodes, or modify their features.
We do not consider changing the graph-based data to a different type of representation,
e.g., by propositionalizing the data or by changing to a logic program.
However, some of the transformations we discuss, such as node or link feature aggregation, are a form of propositionalization.
In addition, Section~\ref{sec:search-select-features} describes a number of techniques for structure learning of logic programs, 
because these techniques are closely related to the analogous problem of feature construction for graph-based representations.
Finally, many of the other techniques that we discuss are also applicable to logical representations.
For instance, link weighting could be applied to weight the known relations before using a logic program to detect anomalous objects.
We focus, however, on the methods most useful for transforming graph-based representations.

\subsection{Approach and Organization of this Article}

There are many dimensions of relational data transformation, which complicate 
the task of understanding and selecting the most appropriate techniques.
To assist in this process, we introduce a simple and intuitive taxonomy for representation transformation that identifies
\textit{link transformation} and \textit{node transformation} as symmetric representation tasks.  
More specifically, the
transformation tasks for both nodes and links include (i) predicting their existence, (ii) predicting
their label or type, (iii) estimating their weight or importance, and (iv) constructing their relevant
features. In addition, we propose a taxonomy for constructing both link and node
features that consists of non-relational features, topology features,
relational node-value features, and relational link-value features.  For each relational transformation task,
we survey the applicable techniques, examine necessary conditions, and provide detailed examples and comparisons.

This article is organized as follows.
The next section presents our taxonomy for relational representation transformation and discusses a motivating example.  
In Section \ref{sec:link-pred}, we review the algorithms for link prediction, while 
Section  \ref{sec:link-interp} examines the task of link interpretation (i.e., constructing link labels, link weights, and link features).
Sections  \ref{sec:node-pred} and  \ref{sec:node-interp} consider the corresponding prediction and interpretation tasks for nodes instead of links.
In Section  \ref{sec:joint-disc}, we summarize algorithms that jointly transform nodes and links. 
Section \ref{sec:discussion} discusses methods for evaluating representation transformations and challenges for future work,
and Section \ref{sec:conclusion} concludes.

%% file: overview.tex
\section{Overview and Motivating Example}\label{sec:overview}

In this section we first introduce a running example based on the classification of data from Facebook, then describe how
relational algorithms could be used to perform this task.  
Next, we introduce a taxonomy for relational representation transformation and explain
how each type of transformation could aid the Facebook classification task.
Finally, we formally define each type of relational representation transformation.

\subsection{Motivating SRL Analysis Example: A Classification Task}\label{sec:motivating-example}

As an example, we consider hypothetical data inspired by Facebook (www.facebook.com), one of the most popular online social networks.
We assume that we are given a graph \ifullgraph ~where the nodes $V$ are users 
\footnote{In general, there may be more than one type of node. For instance, nodes in a citation network may represent papers or authors.} 
and the links $E$ represent friendships in Facebook.
\infeatures\ is a set of features about the users in $V$ such as their gender, relationship status, 
school, favorite movies, or musical preference (though information may be missing for some users). 
Likewise, the set of features \ilfeatures\ provides information about the friendship links in $E$
such as the time of formation or possibly the contents of the message that was sent
when the link formation was requested by one of the users.

The example SRL analysis task (see Figure~\ref{fig:rep-trans}) is to predict the political affiliation (liberal, moderate, or conservative) of every node (person) in 
$G$. We assume that this affiliation, which we call the {\em class label} of a node, is known for some but not all of the people in 
$G$.\footnote{Later, we discuss the representation change of {\em node labeling}, which also constructs an estimated label for every node.
As discussed in Section~\ref{sec:scope}, representation changes can sometimes resemble the output of SRL analysis, but we focus on 
changes that are particularly useful as pre-processing before some subsequent SRL analysis.}
Moreover, we assume that a user's political affiliation is likely to be correlated with the characteristics of that user and
(to a lesser degree) that user's friends.
The next section summarizes how these correlations can be used for classification. 

For this example, we assume that links are simple, binary friendship connections.  
However, other link types could be used to represent other kinds of relationships.
For instance, a link might indicate that two people have communicated via a ``wall-post'' message, or that two
people have chosen to join the same Facebook group.
In addition, the notion of friendship in Facebook is very weak and thus 
a significant portion of a person's ``friends'' are often only casual acquaintances. 
Thus, representation changes such as link deletion or weighting
may have a significant impact on classification accuracy.  
For notational purposes, we add a tilde to the top of each graph component's symbol to indicate that it has 
undergone some transformation (e.g., the modified link set $E$ is denoted by \links).

\subsection{Background: Features and Methods for Classification}\label{sec:bg-classification}
\label{sec:methodsForCC}

To predict the political affiliation of Facebook users, conventional classification approaches would
ignore the links and classify each user using only information known about that user,
such as their gender or location.  
We assume that such information is represented in the form of {\em non-relational features}, which are those features that can be computed
directly from \infeatures\ without considering the links $E$.  We refer to classification based only
on these features as \textit{non-relational classification}.  
Alternatively, in \textit{relational classification}, the 
links are explicitly used to construct additional {\em relational features} to capture information about each user's friends.
For instance, a relational feature could compute, for each user, the proportion of friends that
are male or that live in a particular region.  Using such relational
information can potentially increase classification accuracy, though may sometimes decrease accuracy
as well \shortcite{chakra:98}. 
Finally, even greater (and usually more reliable) increases can occur
when the class labels (e.g., political affiliations) of the linked users are used instead to derive
relevant features \shortcite{jensen:kdd04}.  For instance, a ``class-label'' relational feature could compute,
for each user, the proportion of friends that have liberal views. However, using such features is
challenging since some or all of the labels are initially unknown, and thus typically must be
estimated and then iteratively refined in some way. This process of jointly inferring the labels of
interrelated nodes is known as \textit{collective classification} (CC).

CC requires both models and inference procedures that use inferences about one user to affect inferences about related users.
Many such algorithms have been considered for CC, including Gibbs Sampling \shortcite{jensen:kdd04}, relaxation labeling \shortcite{chakra:98}, belief propagation \shortcite{taskardpm:02}, ICA \shortcite{neville:srl00,getoorlinkbased2003}, and weighted neighbor techniques \shortcite{macskassy2007classification}. See \shortciteA{sen2008collective} for a survey.

As a concrete example of SRL analysis, we explain many of the techniques in this survey in terms of the Facebook
classification task, with a special emphasis on CC.  
However, the features and the transformation
techniques apply to many other SRL tasks and data sets such as relationship classification, anomalous link detection, entity resolution, or group discovery~\shortcite{getoor:expl05}.

\subsection{Representation transformation Tasks for Improving SRL}\label{sec:taxonomy}

\input{fig-taxonomy}

Figure~\ref{fig:rep-disc-taxonomy} shows our proposed taxonomy for relational representation transformation.
The two main tasks in this taxonomy are \textbf{link transformation} and \textbf{node transformation}.We find that there is a powerful and elegant symmetry between these two tasks.  
In particular, the link and node representation transformation tasks can be decomposed into prediction and interpretation tasks.
The former task involves predicting the existence of new nodes and links.
The latter task of interpretation involves three parts: constructing the weights, labels, or features of nodes or links.
Together, this yields eight distinct transformation tasks as shown in the leaves of the taxonomy in Figure~\ref{fig:rep-disc-taxonomy}.
Underneath these eight tasks in the figure, 
we list the primary graph component that is modified by each task (i.e., \nodes,
\links, \nfeatures, or \lfeatures), followed by an illustration of a possible representation change for that task.
In the text below, we summarize Figure~\ref{fig:rep-disc-taxonomy}, organized around 
the four larger categories of link prediction, link interpretation, node prediction, and node interpretation.

First, \textbf{link prediction} adds new links to the graph.
The sample graph for this task (Figure~\ref{fig:rep-disc-taxonomy}A) shows a link being predicted where the similarity between two nodes has been used to predict a new link between them.  
Intuitively, Facebook users that share the values of many non-relational features may also share the same political affiliation. 
Thus, adding links between such people should increase autocorrelation and improve the accuracy of collective classification. 
There are many simple link prediction algorithms based on similarity, neighbor properties, shortest path distances, infinite sums over paths (i.e. random walks), and other strategies.
Section \ref{sec:link-pred} provides more detail on these techniques.

Second, there are several types of \textbf{link interpretation}, which involves constructing weights, labels, or features for the existing links.  
For instance, in many graphs (including our Facebook data), not
all links (or friendships) are of equal importance.  Thus, Figure~\ref{fig:rep-disc-taxonomy}B shows the result of performing {\em link weighting}. In this case, weights are based on the similarity between
the feature values of each pair of linked nodes, under the assumption that high similarity 
may indicate stronger relationships. (Link prediction techniques may also use such similarity measures, but for identifying probable new links, 
rather than weighting existing links.)
Alternatively, {\em link labeling} may be used to
assign some kind of discrete label to each link.  For instance, Figure~\ref{fig:rep-disc-taxonomy}C shows how
links might be labeled as either ``personal'' (p) or ``work'' (w) related, e.g., based on known feature values or an
analysis of communication events between the linked users.  On the other hand, links might instead be labeled
as having positive or negative influence (i.e., labeled as $+$/$-$).  
Finally, Figure~\ref{fig:rep-disc-taxonomy}D shows how {\em link
feature construction} can be used to add more general kinds of feature values to each link.  For
instance, a link feature might count the number of communication events that occurred between
two people or the number of friends in common.  
Link weighting and labeling could perhaps be viewed as special cases of link feature construction, but we separate them
because later sections will show how the most useful techniques for each task differ.
All three of these link interpretation tasks could
help with our example classification problem.  In particular, a model learned to predict
political affiliation might choose to place special emphasis on links that are highly weighted
or that are labeled as personal. Other link features might be used to represent more complex dependencies,
for instance modeling influence from a user's ``work'' friendships, but only for friendship links between nodes where there are a large number of
friends in common. More details on these techniques are provided in Section~\ref{sec:link-interp}.

Third, \textbf{node prediction} adds additional nodes (and associated links) to the
graph.  For instance, Figure~\ref{fig:rep-disc-taxonomy}E shows the result after relational clustering has
been applied to discover two latent groups in the graph, where each user is now connected to one
latent group node. A discovered node in Facebook might represent types of social processes, influences,
or a tightly knit group of friends. The clustering or other techniques used to identify the
new nodes could be designed to identify people that are particularly similar
with respect to a relevant characteristic, such as their political affiliation. The new nodes and
associated links could then be used in several ways.  For instance, though not present in the small
example of Figure~\ref{fig:rep-disc-taxonomy}E, some nodes that were far away (in terms of shortest path length) in the original graph may be
much closer in the new graph.  Thus, links to a latent node may allow influence to propagate more
effectively when an algorithm such as CC is applied.  Alternatively, 
identification of distinct latent groups may even enable more efficient or accurate algorithms to be applied 
separately to each group~\shortcite{neville:icdm05}. Node prediction is discussed further in Section~\ref{sec:node-pred}.

Finally, there are several types of \textbf{node interpretation}, which involves constructing weights, labels,
or feature values for existing nodes.  For instance, as with links, some nodes may be more
influential than others and thus should have more weight.  
Figure~\ref{fig:rep-disc-taxonomy}F demonstrates {\em node weighting}, where the weights might be assigned based on the numbers of friends or
via the PageRank/eigenvector techniques.  See Section \ref{sec:node-weight} for more details.
Alternatively, Figure~\ref{fig:rep-disc-taxonomy}G shows an example
of {\em node labeling}.  
Here the graph represents a training graph,
and each node has been given an estimated label of conservative (C), liberal (L), or moderate (M).
Such labels might be estimated using only the non-relational features or via textual analysis.
While most classification algorithms learn a model based on
true labels in the training graph, some approaches instead first compute such estimated labels, then
learn a model from this new representation~\shortcite{kou2007stacked}.  Section~\ref{sec:node-label} discusses how this can simplify inference. 
Finally, Figure~\ref{fig:rep-disc-taxonomy}H shows the result of {\em node feature construction,} where 
arbitrary feature values are added to each node.  For instance, suppose we find that users with
relatively few Facebook friends are often moderate while those with many friends are often liberal.
In this case, a feature counting the number of friends for each node would be useful.
To more directly exploit autocorrelation, a different feature 
might count the proportion of a user's friends that are conservative, or the most common
political affiliation of a user's friends.  
Any feature that is correlated with political
affiliation could be used to improve the performance of a classification algorithm for our example
problem.  Identifying and/or computing such features is essential to the performance of most SRL
algorithms but can be very challenging; Section \ref{sec:node-feature} considers this
process. 

In Table~\ref{table:relational-discovery-summary}, we summarize some of the most prominent techniques for performing these tasks of link prediction, link interpretation, node prediction, and node interpretation.
Sections \ref{sec:link-pred}-\ref{sec:node-interp} provide more detail about each category in turn.

\input{tab-summary}

\input{tab-symbols}

\subsection{Relational Representation Transformation: Definitions and Terminology}\label{sec:definitions}

We assume that the initial relational data is represented as a graph \ifullgraph\ such that each $v_i \in V$ corresponds to node $i$ and each edge $e_{ij} \in E$ corresponds to a 
(directed) link between nodes $i$ and $j$. 
\infeatures\ is a set of features about the nodes in $V$, and $X^{V}_{k} \in \mathbf{X}^{V}$ is the $k^{th}$ such feature.
Likewise, \ilfeatures\ is a set of features about the links in $E$, and $X^{E}_{k} \in \mathbf{V}^{E}$ is the $k^{th}$ such feature.
The features $\mathbf{X}^{E}$ could refer to link weights, distances, or types, among other possibilities. 
The preceding notation lets us identify, for instance, the values of a particular feature $X^{V}_{k}$ for all nodes.  Alternatively, 
 $\vec{x^{v}_{i}}$ refers to a vector containing all of the feature values for a particular node $v_i$, and 
$\vec{x^{e}_{ij}}$ contains all of the feature values for a particular edge $e_{ij}$.
Table~\ref{table:symbols} summarizes this notation.

Relational representation transformation is the process of transforming the original graph $G$ into some new graph \fullgraph\ by
an arbitrary set of transformation techniques.  During this process, nodes, links, weights, labels, and general features may be added,
and nodes and links may be removed.  In theory, the transformation seeks to optimize some objective function (for instance, 
to maximize the autocorrelation), although in practice the objective function may not be completely specified or guaranteed to be improved
by the transformation. We now define more specifically the four primary parts of relational representation transformation:

\medskip
\noindent
\textbf{Definition 2.1 (Link Prediction)}
Given the nodes $V$, observed links $E$ and/or the feature set \ifeatures, the link prediction task is defined as the creation of 
a modified link set \links\ such that $E \neq$ \links.  Usually, this involves adding new links that were not present in $E$, but links may also be deleted.

\medskip
\noindent
\textbf{Definition 2.2 (Link Interpretation)} Given the nodes $V$,
observed links $E$ and/or the feature set \ifeatures, the link
interpretation task is defined as the creation of a new link feature
\lfeature\ where \lfeature\ $\notin$ \ilfeatures.  
This task may estimate a feature value for every link.  Alternatively, the values of \lfeature\ may be only 
partially estimated, for example, if the original features have missing values or if 
additional links are also introduced during link prediction.

\medskip
\noindent
\textbf{Definition 2.3 (Node Prediction)}
Given the nodes $V$, links $E$ and/or the feature set \ifeatures, node transformation is defined as the creation of a 
modified node set \nodes\ such that $V \subset$ \nodes.  In addition, many node prediction tasks simultaneously 
create new links, e.g., between an initial node $v_{i} \in V$ and a predicted node $\tilde{v}_{j} \in \tilde{V}$. 
Thus, this task may also produce a modified link set \links.

\medskip
\noindent
\textbf{Definition 2.4 (Node Interpretation)}
Given the nodes $V$,
observed links $E$ and/or the feature set \ifeatures, the node
interpretation task is defined as the creation of a new node feature
\nfeature\ where \nfeature\ $\notin$ \infeatures.  
As with link interpretation, the values of \nfeature\ may be estimated
for only some of the nodes.
The node feature \nfeature\ could represent node weights, labels, or other general features. 

Section~\ref{sec:methodsForCC} introduced the notion of a non-relational feature, which is a node feature \nfeature\ that can be constructed without making use of the links
(i.e., without using $E$ or \ilfeatures).  Such features are sometimes referred to in other articles as {\em attributes} or {\em intrinsic features}.
Other important terms can also be referred to in multiple different ways.
To aid the reader, Table~\ref{table:synonyms} summarizes the key synonyms for the terms that are found most often
in the literature.

\input{tab-synonyms}

%% file: fig-taxonomy.tex
\begin{sidewaysfigure}
\hspace{-4mm}
\includegraphics[width=9.4in]{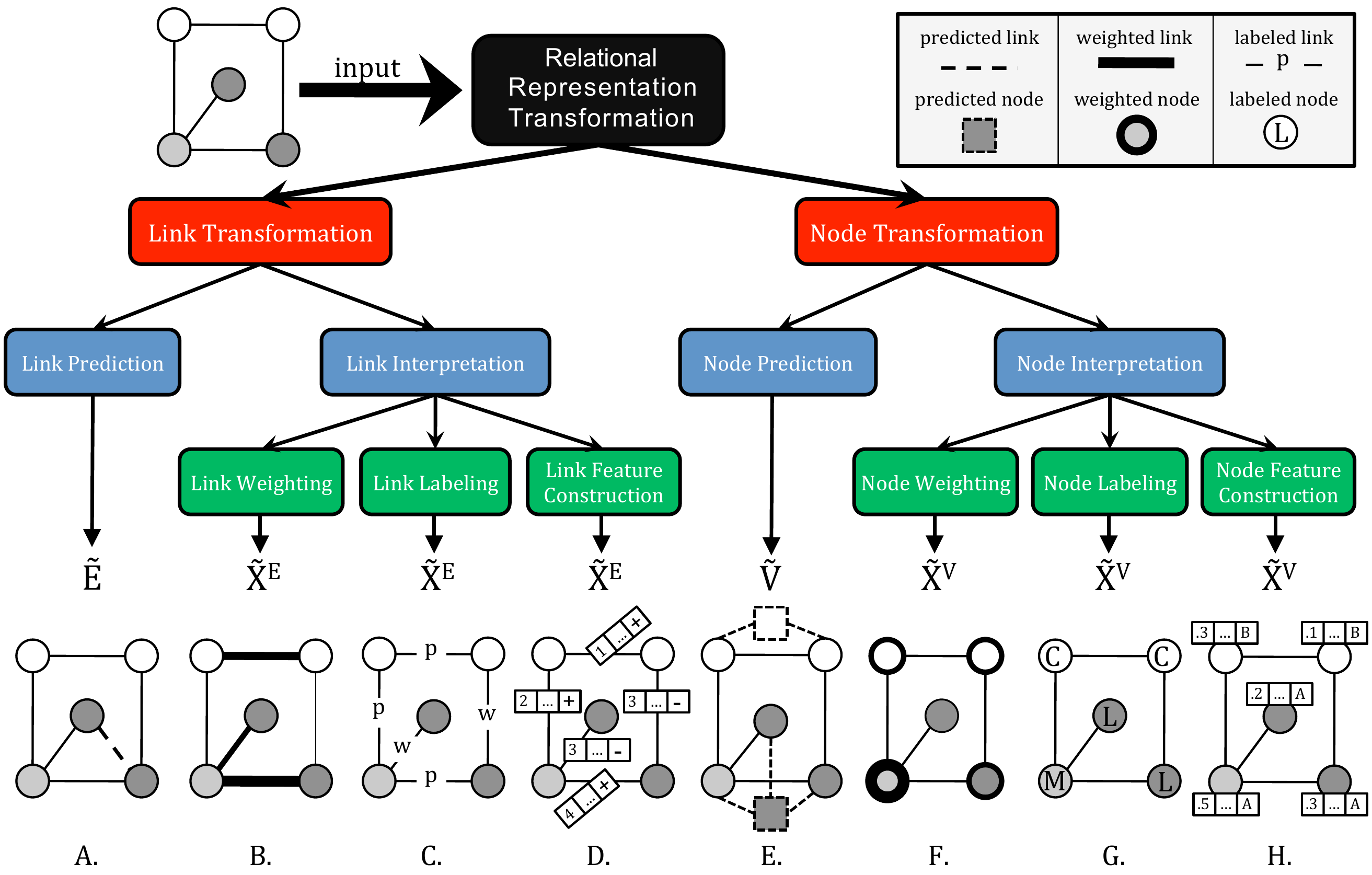} 
\caption{\textsc{Relational Representation Transformation Taxonomy}: Link and node transformation are formulated as symmetric tasks leading to four main transformation tasks: link prediction, link interpretation, node prediction, and node interpretation.  
Each task yields a modified graph component: \links, \lfeatures, \nodes, or \nfeatures, respectively.
Interpretation is further divided into weighting, labeling, or constructing features.
Examples of each of the tasks in relational representation transformation are shown under the leaves of the taxonomy. 
In these example graphs, nodes with similar shadings have similar feature values.}
\label{fig:rep-disc-taxonomy}
\end{sidewaysfigure}

%% file: tab-summary.tex
\definecolor{light-gray}{gray}{0.50}

\newcommand\TT{\rule{0pt}{3.8ex}}
\newcommand\BB{\rule[-1.8ex]{0pt}{0pt}}

\begin{table}[t!]
\caption{\textsc{Summary of techniques}:  A summary of prominent graph transformation techniques for the tasks of predicting the existence of nodes and links and interpreting them by weighting, labeling, and constructing general features.}
\label{table:relational-discovery-summary}
\vspace{-6mm}
\begin{center}
\begin{small}
\begin{tabular}{c||cp{50mm}||cp{50mm}}

\toprule
\bottomrule

\multicolumn{5}{c}{\T \B \Large \textbf{Relational Representation Transformation}}\\
\toprule
\bottomrule
\T \B
& \multicolumn{2}{c||}{\Large \textbf{Links}}
& \multicolumn{2}{c}{\Large \textbf{Nodes}}\\

\hline
\hline

\multirow{8}{*}{\large \textbf{Prediction}} & 
$\star$ & 
\textbf{Adamic/Adar} \shortcite{Adamic01friendsand}, \textbf{Katz} \shortcite{katz1953new}, \textbf{and others} \shortcite{linkpred:liben07}
&
$\star$ &
\textbf{Spectral Clustering} \shortcite{neville:icdm05}, \textbf{Mixed-Membership Relational Clustering} \shortcite{long2007probabilistic}
\\ 

&
$\star$ &
\textbf{Text or Feature Similarity} \shortcite{SofusMinedLinks}
& 
$\star$ &
\textbf{LDA} \shortcite{blei:03}, \textbf{PLSA} \shortcite{plsa99},  
\\

&
$\star$ &
\textbf{Classification via RMN} \shortcite{Taskar03linkprediction} \textbf{or SVM} \shortcite{Hasan06linkprediction}
&
$\star$ &
\textbf{Hierarchical Clustering via Edge-betweenness} \shortcite{newman2004finding}
\\ 

\hline 
\hline

\multirow{8}{*}{\large \textbf{Weighting}} & 
  \scriptsize
$\star$ &
\textbf{Latent Variable Estimation}~\shortcite{xiang:10} 
&
$\star$ & 
\textbf{Betweenness} \shortcite{betweennessfreeman}, \textbf{Closeness} \shortcite{sabidussi1966centrality} \\ 

\multirow{5}{*}{\textbf{}}& 
$\star$ &
\textbf{Linear Combination of Features} \shortcite{tiestrengthEricGilbert} 
& 
$\star$ &
\textbf{HITs} \shortcite{kleinberg1999authoritative}, \textbf{Prob. HITs} \shortcite{cohn2000learning}, \textbf{SimRank} \shortcite{jeh2002simrank} \\ 

& 
$\star$ &
\textbf{Aggregating Intrinsic Information}~\shortcite{barabasi:07} 
&
$\star$ &
\textbf{PageRank} \shortcite{page1998pagerank}, \textbf{Topical PageRank} \shortcite{haveliwala2003topic,Richardson02theintelligent} \\ 

\hline
\hline

\multirow{7}{*}{\large \textbf{Labeling}} & 
$\star$ &
\textbf{LDA} \shortcite{blei:03}, \textbf{PLSA} \shortcite{plsa99},   
& 
$\star$ &
\textbf{LDA} \shortcite{blei:03}, \textbf{PLSA} \shortcite{plsa99},  \\

\multirow{4}{*}{\textbf{}} & 
$\star$ &
\textbf{Link Classification via Logistic Regression} \shortcite{Les_predpos}, \textbf{Bagged Decision Trees} \shortcite{neville:aaai09},  
&
$\star$ & 

\textbf{Node Classification via Stacked Model} \shortcite{kou2007stacked} or \textbf{RN} \shortcite{macskassy:03} \\

\hline
\hline

\multirow{7}{*}{\large \textbf{Feature}} & 
$\star$ &
\textbf{Link Feature Similarity} \shortcite{rossi:10} 
& 
$\star$ &
\textbf{Database Query Search} \shortcite{Popescul03SLR}, \textbf{RPT} \shortcite{neville:kdd03}  \\ 

\multirow{3}{*}{\textbf{Construction}} & 
$\star$ &
\textbf{Link Aggregations} \shortcite{neville:aaai09} & 
$\star$ &
\textbf{MLN Structure Learning} \shortcite{Kok09structlearning,kokMotifs2010} \\ 

& 
$\star$ &
\textbf{Graph Features} \shortcite{niteshchawlakdd2010} & 
$\star$ &
\textbf{FOIL}, \textbf{nFOIL} \shortcite{landwehr2005nfoil}, \textbf{kFOIL} \shortcite{landwehr2010fast}, \textbf{Aleph} \shortcite{srinivasan1999aleph},  \\ 

\hline
\hline

\toprule
\bottomrule
\end{tabular}
\end{small}
\end{center}
\vspace{-6mm}
\end{table}

%% file: tab-symbols.tex
\newcommand\TC{\rule{0pt}{1.8ex}}
\newcommand\BC{\rule[-1.8ex]{0pt}{0pt}}

\begin{table}[t!]
\caption{\textsc{Summary of notation used in this survey}:  The top half of the table shows symbols that are sometimes written with a tilde on top of the symbol,
indicating the result of some transformation.  For conciseness, the table demonstrates this notation only for $G$ and $\tilde{G}$. }
\label{table:symbols}
\begin{center}
\small
\begin{tabularx}{\linewidth}{ cX}
\toprule
\textsc{\textbf{Symbol}} & \textsc{\textbf{Description}} \\
\midrule
\TC \BC \multirow{1}{*}{$G$} & Initial graph \\
\TC \BC \multirow{1}{*}{\graph} & Transformed graph \\
\TC \BC \multirow{1}{*}{$E$} & Initial link set \\
\TC \BC \multirow{1}{*}{$V$} & Initial node set \\
\TC \BC \multirow{1}{*}{\ilfeatures} &  Initial set of link features\\
\TC \BC \multirow{1}{*}{\infeatures} & Initial set of node features\\
\TC \BC \multirow{1}{*}{$X^{E}_{k}$} &  Initial link feature $k$ ($X^{E}_{k} \in $\ilfeatures) (for one feature, values for all links)\\ 
\TC \BC \multirow{1}{*}{$X^{V}_{k}$} &  Initial node feature $k$ ($X^{V}_{k} \in $\infeatures) (for one feature, values for all nodes)\\
\TC \BC \multirow{1}{*}{$\vec{x^{e}_{ij}}$} &  Initial feature vector for $e_{ij}$ (for one link, values for all link features)\\
\TC \BC \multirow{1}{*}{$\vec{x^{v}_{i}}$} &  Initial feature vector for $v_i$ (for one node, values for all node features)\\
\toprule
\textsc{\textbf{Other symbols}} & \textsc{\textbf{Description}}  \\
\midrule
\TC \BC \multirow{1}{*}{$\mA$} & \multicolumn{1}{l}{Adjacency matrix of the graph}\\
\TC \BC \multirow{1}{*}{$\Gamma(v_{i})$} &  \multicolumn{1}{l}{Neighbors of $v_{i}$} \\
\TC \BC \multirow{1}{*}{$\delta$} &  \multicolumn{1}{l}{Cut-off value} \\
\bottomrule
\end{tabularx}
\end{center}
\end{table}

%% file: tab-synonyms.tex
\newcommand\TK{\rule{0pt}{1.8ex}}
\newcommand\BK{\rule[-1.8ex]{0pt}{0pt}}

\begin{table}[t!]
\caption{\textsc{Possible synonyms for important terms related to relational data}.}
\vspace{2mm}
\label{table:synonyms}
\begin{center}
\begin{small}
\begin{tabularx}{\linewidth}{ lX}
\toprule
\textsc{\large\textbf{Term}} & \textsc{\large\textbf{Potential synonyms}} \\
\midrule
\TK \BK \multirow{1}{*}{\textbf{Nodes}} &  \multicolumn{1}{l}{Vertices, points, objects, entities, individuals, users, constants, ...} \\
\TK \BK \multirow{1}{*}{\textbf{Links}} &  \multicolumn{1}{l}{Edges, relationships, ties, arcs, events, interactions, predicates} \\ 
\TK \BK \multirow{1}{*}{\textbf{Topology}} &  \multicolumn{1}{l}{Link/network/graph structure, relational information} \\
\TK \BK \multirow{1}{*}{\textbf{Features}} &  \multicolumn{1}{l}{Attributes, variables, co-variates, queries, predicates, ...} \\
\TK \BK \multirow{1}{*}{\textbf{Graph Measures}} &  \multicolumn{1}{l}{Topology-based metrics (such as proximity, centrality, betweenness, ...)} \\
\TK \BK \multirow{1}{*}{\textbf{Similarity}} &  \multicolumn{1}{l}{Distance (the inverse of similarity), likeness} \\
\TK \BK \multirow{1}{*}{\textbf{Clusters}} &  \multicolumn{1}{l}{Classes, communities, groups, roles, topics} \\
\TK \BK \multirow{1}{*}{\textbf{Non-relational Features}} &  \multicolumn{1}{l}{Intrinsic attributes/features, local attributes/features, ...} \\
\TK \BK \multirow{1}{*}{\textbf{Relational Features}} &  \multicolumn{1}{l}{Features, link-based features, graph features, aggregates, queries, ...} \\
\TK \BK \multirow{1}{*}{\textbf{Structure Learning}} &  \multicolumn{1}{l}{Feature generation/construction, hypothesis learning} \\
\TK \BK \multirow{1}{*}{\textbf{Parameter Learning}} &  \multicolumn{1}{l}{Model selection, function learning} \\
\bottomrule
\end{tabularx}
\end{small}
\end{center}
\end{table}

%% file: link-pred.tex
\section{Link Prediction}\label{sec:link-pred}

This section focuses on predicting the existence of links while
Section~\ref{sec:link-interp} considers link interpretation.  Given
the initial graph \ifullgraph, we are interested in creating a
modified link set \links, usually through the prediction of new links that
were not present in $E$.  This task can be motivated in several ways.
For instance, there may be a need to predict \textit{missing links}
that are not present in $E$ because of incomplete data collection or other
problems.  Similarly, we may be interested in predicting
\textit{hidden links}, where we assume that there exists some
unobservable interactions and the goal is to discover and model these
interactions.  For example, in a network representing criminals or
terrorist activity, we may seek to predict a link between two people (nodes)
that are not directly connected but whose actions share some common
motivation or cause.  
For both missing and hidden links, predicting such links may improve the accuracy of a subsequent learned model.
Alternatively, we may seek to predict
\textit{future links} in an evolving network, such as new friendships
or connections that will be formed next year.  We might also be
interested in predicting links between objects that are spatially
related.  Finally, we may wish to predict \textit{beneficial links},
for instance, predicting pairs of individuals that are likely to be
successful working together.

Figure~\ref{fig:lp} summarizes one general approach that is often used 
for these link prediction tasks.  In summary, scores or weights are computed for every pair of 
nodes in the graph, as shown in Figure~\ref{fig:lp}(b).
Predicted links with a weight greater than some threshold $\delta$, along
with the original links, are used to create the new link set $\tilde{E}^+$ (shown in Figure~\ref{fig:lp}(e)).
(At this step, original links with very low weight could also be deleted if appropriate.)
As a final step, the weights of the predicted links are often discarded, yielding a new graph with uniform link
weights as shown in Figure~\ref{fig:lp}(f).

The key challenge in this approach is how to compute a weight or score for each possible link. The information used for this computation provides a natural way to categorize the link prediction techniques.
Below, Section~\ref{sec:feature-lp} describes techniques that use only the
non-relational features of the nodes (ignoring the initial links), while Section~\ref{sec:topology-lp}
describes ``topology-based'' techniques that use only the graph structure (i.e., the links or relations).
Finally, Section~\ref{sec:hybrid-lp} describes hybrid techniques that exploit both the node features and the graph structure.

\input{fig-lp}

\subsection{Non-relational (Feature-based) Link Prediction}\label{sec:feature-lp}

In this section, we consider link predictors that do not exploit the graph structure or relational features derived using the graph structure.
We are given an arbitrary pair of nodes $v_{i}$ and $v_{j}$ from the graph such that each node is represented by a feature vector $\vec{x^{v}_{i}}$ and $\vec{x^{v}_{j}}$, respectively. \textit{Feature-based link prediction} is defined as using an arbitrary similarity measure $S(x^v_{i}, x^v_{j})$ as a means to estimate the likelihood that a link should exist between $v_{i}$ and $v_{j}$.  Typically, a link is created if the similarity exceeds some fixed cut-off value;
another strategy is to predict links among the n\% of all such node pairs with highest similarity.

A traditional approach is to simply \textit{define} a measure of similarity between two objects, possibly based on knowledge of the application and/or problem-domain. There are many similarity metrics that have been proposed such as mutual information, cosine similarity, and many others \shortcite{Lin98aninformation-theoretic}. 
For instance, \shortciteA{SofusMinedLinks} represents the textual content of each node as a feature vector and uses cosine similarity to create new links between nodes in a graph. 
\shortciteauthor{SofusMinedLinks} showed that the combination of the initial links with the predicted text-based links increased classification accuracy compared to using only the initial links or the text-based links.
In addition to leveraging textual information to predict links, we might use any arbitrary set of features combined with a proper measure of similarity for link prediction. 
For instance, many recommender systems implicitly predict a link between two users based on the similarity between their ratings of items such as movies or books \shortcite{adomavicius2005toward,resnick1997recommender}.
In this case, cosine similarity or correlation are commonly used as similarity metrics.

Alternatively, a similarity measure can be \textit{learned} for predicting link existence. 
The link prediction problem can be transformed into a standard supervised classification problem where a binary classifier is trained to determine the similarity between two nodes based on their feature vectors.
One such approach by \shortciteA{Hasan06linkprediction} applied Support Vector Machines (SVMs) and found that a non-relational feature (keyword match count) was most useful for predicting links in a bibliographic network. There are many link prediction approaches 
~\shortcite{Taskar03linkprediction,getoor2003pmlink} that apply traditional machine learning algorithms. However, most of them use features based on the graph structure as well as the non-relational features that are the focus of this section.
Thus, we discuss such techniques further in Section~\ref{sec:hybrid-lp}.

Finally, variants of {\em topic models} can be used for link prediction.
These types of models traditionally use only the text from documents (non-relational information) to infer a mixture of latent topics for each document.
Inter-document topic similarity can then be used as a similarity metric for link prediction \shortcite{chang2009relational}.
However, because many topic models are capable of performing joint transformation of the nodes and links,
we defer full discussion of such techniques to Section~\ref{sec:joint-disc}.

\subsection{Topology-based Link Prediction}\label{sec:topology-lp}
Topology-based link prediction uses the local relational neighborhood and/or the global graph structure to predict the existence of
unobserved links.
Table~\ref{table:metrics} summarizes some of the most common metrics that have been used for this task.
Below, we discuss many of these approaches, starting from the simplest local metrics and moving to the more complex techniques
based on global measures and/or supervised learning.  For a systematic study of many of these approaches applied to social network data, see \shortciteA{linkpred:liben07}.

\input{table-metrics}

\smallskip
\noindent
{\bf Metrics based on the Local Neighborhood of Nodes:}
The simplest approaches use only the local neighborhood of nodes in a graph to devise a measure of topology similarity, then use pairwise similarities between
nodes to predict the most likely links.
As shown in Table~\ref{table:metrics}, there are numerous such metrics, often based on the number of neighbors that two nodes share in common, with
varying strategies for normalization.

\shortciteA{zhou2009predicting} compares nine such local similarity measures on six datasets and finds that the simplest link predictor, common neighbors, performs the best overall. 
They also propose a new metric, RA, that outperforms the initial nine metrics on two of the datasets. 
This new metric is very similar to the Adamic/Adar metric, but uses a different normalization factor that yields better performance in networks with higher average degree.
They also propose a method that uses additional two-hop information to avoid degenerate cases where links are assigned the same similarity score. 
Their results highlight the importance of selecting the appropriate metrics for specific problems and datasets.
In another related investigation, \shortciteA{clauset2008hierarchical} evaluates a hierarchical random graph predictor against local topology metrics such as common neighbors, Jaccard's coefficient and the degree product on three types of networks: a metabolic, ecology and a social network.
They find that a baseline measure based on shortest paths performs best for the metabolic network, where the relationships are more homogeneous, but that
their hierarchical metric performs best when the links create more complex relationships, as in the predator-prey relationships found in the 
ecology network.

\shortciteA{liu2010link} proposed a local random-walk algorithm as an efficient alternative to the global random-walk predictors for large networks. This method is evaluated alongside other metrics (i.e., common neighbors, local paths, RA, and a few random-walk variants) and shown to perform better on most of the networks and more efficiently than the global random-walk models.

\smallskip
\noindent
{\bf Metrics based on the Global Graph Structure:}
More sophisticated similarity metrics are based on global graph properties,
often involving some weighted computation based on the number of paths between a pair of nodes.
For instance, the Katz measure~\shortcite{katz1953new} counts the number of paths between a pair of nodes, where shorter paths count more
in the computation.  
\shortciteA{rattAnomKDDExpl05} demonstrated that even this fairly simple metric could be effective for the task
of ``anomalous link prediction'', which is the identification of statistically unlikely links from among the links in the initial graph.

A related measure is the ``hitting time'' metric, which is the average number of steps
required for a random walk starting at node $x$ to reach node $y$.
\shortciteA{gallagher:08} use such random walks with restart
to estimate the similarity between every pair of nodes.  
They focus on sparsely labeled networks where unlabeled nodes may have only a few labeled nodes to support learning and/or inference in relational classification.
The prediction of new links improves the flow of information from labeled to unlabeled nodes,
leading to an increase in classification accuracy of up to $15\%$.
Note that adding teleportation probabilities to this random walk approach roughly yields the PageRank algorithm 
which is said to be at the heart of the Google search engine~\shortcite{page1998pagerank}.

The SimRank metric~\shortcite{jeh2002simrank} proposes that two nodes $x$ and $y$ are
similar if they are linked to neighbors that are similar.  Interestingly, they show that this approach
is equivalent to a metric based on the time required for two backwards, random walks starting from $x$ and
$y$ to arrive at the same node.  As with the other approaches based on random walks,
this metric could be computed via repeated simulations, but is more efficiently computed via
a recursive set-point approach.

\smallskip
\noindent
{\bf Meta-approaches and supervised learning approaches:}
The metrics above can be modified or combined in multiple ways.  
\shortciteA{linkpred:liben07} consider several such ``meta-approaches'' 
that use some local or global similarity metric as a subroutine.
For instance, the metrics discussed above can each be defined in terms of 
an arbitrary adjacency matrix $\mA$.  Given this formulation, we can
imagine first computing a low-rank approximation $\mA_k$ of this matrix using
a technique such as singular value decomposition (SVD), and then computing a local or global graph
metric using the modified $\mA_k$.  The idea is that $\mA_k$ retains the key structure
of the original matrix, but noise has been reduced.
\shortciteA{linkpred:liben07} also propose two other meta-approaches based on removing spurious links
suggested by a first round of similarity computation (the ``clustering'' approach) or based on
augmenting similarity scores for a node $x$ based on the scores for other nodes that are similar
to $x$ (the ``unseen bigrams'' approach).
They compare the performance of these three meta-approaches vs.\ multiple local and global metrics 
on the task of predicting future links in a social network.
The Katz measure and meta-approaches based on clustering and low-rank approximation perform the best on three of the five arXiv datasets,
but simple local measures such as common neighbors and Adamic/Adar also perform surprisingly well.

Supervised learning methods can also be used to combine or augment the similarity
metrics that we have discussed.  For instance,
\shortciteA{niteshchawlakdd2010} investigate several supervised
methods for link prediction in sparsely labeled networks, using many
of the metrics from Table~\ref{table:metrics}. These metrics are used
as features in simple classifiers such as C4.5, J48, and naive Bayes.
They find the supervised approach leads to a $30\%$ improvement in AUC
over the simple unsupervised link prediction metrics.  Similarly,
\shortciteA{abe:linkpred2006} propose a supervised
probabilistic model that assumes that a biological network has evolved
over time, and uses only topological features to estimate the model
parameters.  They evaluate the proposed method on protein-protein and
metabolic networks and report increased precision compared to simpler
metrics such as Adamic/Adar, Preferential Attachment, and
Katz.

\smallskip
\noindent
{\bf Discussion:}
In general, 
the local topology metrics sacrifice an amount of accuracy for computational gains while the global graph metrics may perform better but are costly to estimate and infeasible on huge networks.
Where appropriate, supervised methods that combine multiple local metrics may offer a promising alternative.
The next subsection discusses additional work on link prediction that has used supervised methods.

Link prediction using these metrics is especially sensitive to the characteristics of the domain and application.
For instance, many networks in biology, where the identification of links is costly, contain missing or incomplete links, while 
the removal of insignificant links is a more significant issue for social networks.
For that reason, researchers have analyzed and proposed many different
metrics when working in the domains of web analysis~\shortcite{kleinberg1999authoritative,broder2000graph}, social network
analysis~\shortcite{zheleva2008using,xiang:10,koren2007measuring}, citation analysis~\shortcite{borgman2002scholarly}, ecology communities~\shortcite{zhou2009predicting}, biological networks~\shortcite{jeong2000large}, and many others~\shortcite{barabasi2003linked,Newman03thestructure}.

\subsection{Hybrid Link Prediction}\label{sec:hybrid-lp}
In this subsection, we examine approaches that perform link prediction
using both the attributes and the graph topology.
For such approaches, there are two key questions.  First, what kinds of features 
should be used?
Second, how is the information from multiple features combined 
into a single measure or probability to be used for prediction?

We first consider the mix of non-relational and relational features that should be used.
As expected, the best features vary based on the domain and specific network.
For instance, \shortciteA{Taskar03linkprediction} studied link prediction for a network of web pages and found that simple local topology metrics
(which they called \textit{transitivity} and \textit{similarity}) were more important than non-relational features based
on the words presents in the pages.
Similarly, \shortciteA{Hasan06linkprediction} found that another topology metric (shortest distance) was the most useful for predicting
co-authorship links in a bibliographic network based on DBLP.

If only a single metric/feature, such as ``hitting time,'' will be used for link prediction, then 
we must ensure that the metric works well for all nodes and yields a consistent ranking.  
However, if multiple feature values will be combined in some way, then it may be more acceptable to use
a wider range of features, especially if a supervised learner will later select or weight the most important
features based on the training data.  Thus, hybrid systems for link prediction tend to have a more diverse
feature set.  For instance, \shortciteA{zheleva2008using} propose new features based on combining two different kinds
of networks (social and affiliation networks).  Features based on the groups and topology are constructed 
from the combined network and are used along with descriptive non-relational features, yielding an
improvement of 15-30\% compared to a system without the combined-network features.
A second example of more complex features is provided by \shortciteA{ben2005kernel}, who 
design a new pairwise kernel for predicting links between proteins (protein-protein interactions). 
The pairwise kernel is a tensor-product of two linear kernels on the original feature space,
and is especially useful in domains where two nodes might have only a few common features. 
This approach has also been applied for user preference prediction and recommender systems~\shortcite{Basilico04unifyingcollaborative}. 
\shortciteA{vert2005supervised} propose a related approach, where supervised learning
is used to create a mapping of the original nodes into a new
euclidean space where simple distance metrics can then be used for link prediction.

Given the great diversity of possible features for link prediction, an interesting approach
is a system that automatically searches for relevant features to use.
For example, \shortciteA{Popescul03SR} propose a unique link prediction approach that systematically generates and searches over a space of relational features to learn potential link predictors. They use logistic regression for link
prediction and consider the search space covering equi-joins, equality selections, and aggregation operations.
In their approach, the model selection algorithm continues to add one feature at a time to the model
as long the Bayesian Information Criterion (BIC) score over the training set can be improved.
They find that the search algorithm discovers a number of useful topology-based features, such as co-citation
and bibliographic coupling, as well as more complex features.  However, the 
complexity of searching a large feature space and avoiding overfitting present challenges.

We next consider the second key question: how should the information
from multiple features be combined into a single measure to be used
for link prediction?  Most prior work has taken a supervised learning
approach, where both non-relational and topology-based metrics are
used as features that describe each possible link. As with the supervised
techniques discussed in Section~\ref{sec:topology-lp}, a model is
learned from training data which can then be used to predict unseen
links.  

Most of these supervised approaches apply the classifier separately to each
possible link, using a classifier such as a support vector machine,
decision tree, or logistic
regression~\shortcite{Popescul03SR,ben2005kernel,Hasan06linkprediction}.
In these approaches, a ``flat'' feature representation for each link is created,
and the prediction made for each possible link is independent of the other predictions.

In contrast, early work on Relational Bayesian Networks (RBNs) \shortcite{getoor2003pmlink} and
Relational Markov Networks (RMNs) \shortcite{Taskar03linkprediction} involved a joint inference computation
for link prediction, where each prediction could be influenced by nearby link predictions (and sometimes
also by newly predicted node labels). Using a webpage network and a social network, 
\shortciteA{Taskar03linkprediction} demonstrated that joint inference
using belief propagation could improve accuracy compared to the independent inference approach.
However, this approach is computationally intensive, and they noted that getting the belief propagation algorithm
to converge was a significant problem.
A possible solution to this computational challenge is the simpler approach presented by \shortciteA{getoor:icdmw07}.  
Their method involved repeatedly predicting labels for each node, predicting links between the nodes using all available features
(including predicted labels), then re-predicting the labels with the new links, and so forth.
The link prediction 
was based on an independent inference step using logistic regression, as with the simpler approaches discussed above.
However, the repeated application of this step allows the possibility of link feature values changing in between
iterations based on the intermediate predictions, thus allowing link predictions to influence each other.

Recently, \shortciteA{backstrom2011supervised} proposed a novel approach that is supervised, but where the final predictions
are based on a random walk rather than directly on the output of some learned classifier.
Given a particular target node $v$ in a social network, along with nodes that are known to link to $v$, they study how to predict
which other links from $v$ are likely to arise in the future (or should be recommended).  
They define a few simple link features based on node profile similarity and messaging behavior, then use these features
to estimate initial link weights.  They show how to learn these weights (or transition probabilities) in a manner that 
optimizes the likelihood that a subsequent random walk, starting at $v$, will arrive at nodes already known to link to $v$.
Because the random walk is thus guided by the links that are already known to exist, they call this process a ``supervised random walk.''
They argue that this learning process greatly reduces the need to manually specify complex graph-based features, and show that it
outperforms other supervised approaches as well as unsupervised approaches such as the Adamic/Adar measure.

A final approach for link prediction is
to use some kind of unsupervised dimensionality reduction that yields
a new matrix that in some way reveals possible new links.  
For instance, \shortciteA{hoff2002latent} propose a latent space approach
where the initial link information is projected into a low-dimensional space.  Link existence can
then be predicted based on the spatial representation of the nodes in
the new latent space.
These models perform a kind of factorization of the link adjacency matrix and thus are often referred to
as matrix factorization techniques.
An advantage of such models
is that the spatial representation enables simpler visualization and
human interpretation.  
Related approaches have also been
proposed for temporal networks~\shortcite{sarkar2005dynamic}, for
mixed-membership models~\shortcite{nowicki:01,blei:jmlr08}, and for situations where the latent vector representing each node
is usefully constrained to be binary \shortcite{miller2009nonparametric}.
Typically, these models have the capability of including the 
attributes as covariates that affect the link prediction but are not directly part of the latent space
representation.  However, \shortciteA{zhu2007combining} demonstrated
how such attributes can also be represented in a related but distinct latent space.
More recently, \shortciteA{menon2011link} showed how a matrix factorization technique for link prediction can scale to much
larger graphs by training with stochastic gradient descent instead of MCMC.

\subsection{Discussion}\label{sec:discussion-lp}
Link prediction remains a challenge, in part because of the very large number of possible links (\ie, $N^2$ possible links given $N$ observed nodes),
and because of widely varying data characteristics.
Depending on the domain, the best approach may use only a single non-relational metric or topology metric, or it may use a richer set of features
that are evaluated by some learned model.  Future work may also wish to consider using an ensemble of link predictors to yield even better accuracy.

Our discussion of link prediction has focused on predicting new links
based on existing links and properties of the nodes.  In the context
of the web, however, ``link prediction'' has sometimes taken other
forms.  For instance, \shortciteA{rameshMarkovPath2000} use web server
traces to predict the next page that a user will visit, given their
recent browsing history.  In particular, they use Markov chains, which
are related to the random walks discussed in
Section~\ref{sec:topology-lp}, for this task that they also call
``link prediction.''  More recently,~\shortciteA{duboisevents} model
relational events (i.e., links) using latent classes where each
event/link arises from a latent class and the properties of the event
(i.e. sender, receiver, and type) are chosen from distributions over
the nodes conditioned on the assigned class.  In this work, the local
community of a node influences the distribution computed for each node,
in a way related to the computations of stochastic block
modeling~\shortcite{blei:jmlr08}.  DuBois \& Smyth's task 
is also a form of link prediction,
but where the goal is not to predict the presence or absence of a
static link, but the frequency of occurrence for each possible
event/link.

One might also be interested in deleting or pruning away noisy, less
informative links.  For instance, friendship links in Facebook are
usually extremely noisy since the cost of adding friendship links is
insignificant.  Most of the techniques used in this section could also
be used to remove existing links wherever the link prediction
algorithm yields a very low score (or weight) for an observed link in the original
graph.

Indeed, since most link prediction algorithms effectively assign a score 
to every possible link, they could also be used to assign a weight
to {\em just} the set of initial links in $G$.  This ``link weighting''
is one of the three subtasks of link interpretation shown in the taxonomy of
Figure~\ref{fig:rep-disc-taxonomy}.
However, in practice if weights are needed only for the initial links,
different features and algorithms will often be possible and/or more effective.
The next section discusses such link weighting algorithms, as 
well as link interpretation in general.
Also, in Section~\ref{sec:joint-disc} we discuss some additional methods
for link prediction that seek to jointly transform both nodes and links.

%% file: fig-lp.tex
\begin{figure}[t]
\centering
\subfigure[Initial Graph $G = \langle E, V \rangle$]{
\includegraphics[width=1.6in]{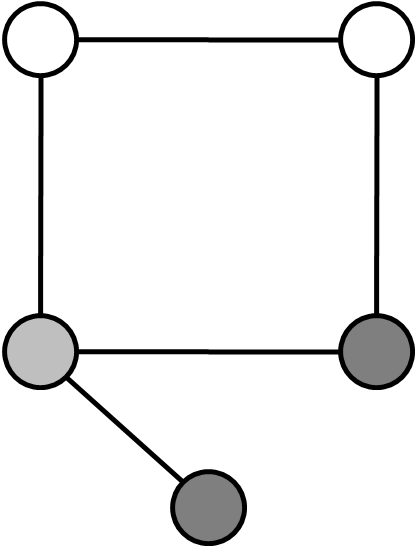}
\label{fig:lp-initial-graph}
}
\subfigure[Weighted Links $w_{ij} \in \tilde{E}$]{
\includegraphics[width=1.6in]{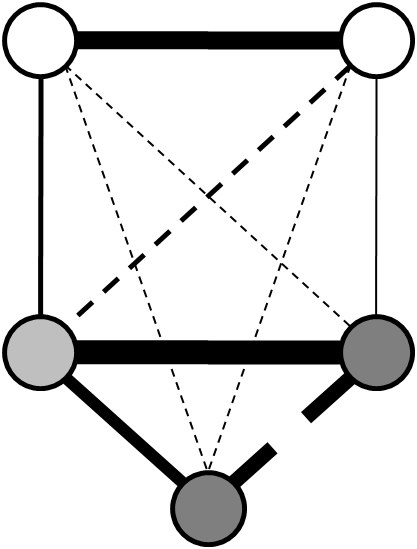}
\label{fig:lp-weighted-links}
}
\subfigure[Predicted Links ($\tilde{E} - E$)]{
\includegraphics[width=1.6in]{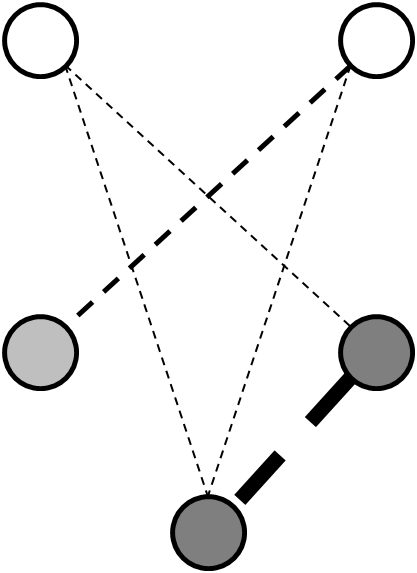}
\label{fig:lp-possible-links}
}
\subfigure[Pruning Predicted Links ($\tilde{E}^{> \delta}$)]{
\includegraphics[width=1.6in]{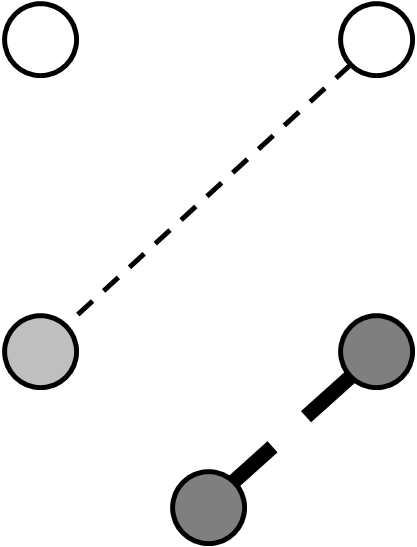}
\label{fig:lp-pruning-links}
}
\subfigure[$\tilde{E}^+ := \tilde{E}^{> \delta} + E$]{
\includegraphics[width=1.6in]{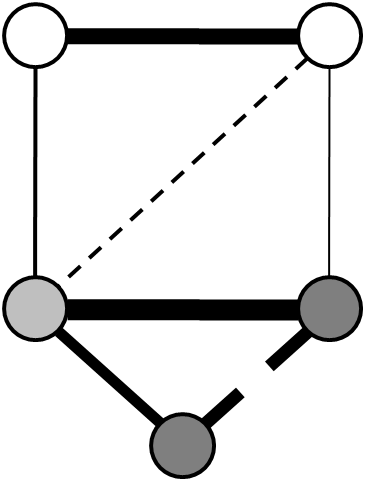}
\label{fig:lp-union-links}
}
\subfigure[$\tilde{E}^+$ with Uniform Link Weights]{
\includegraphics[width=1.6in]{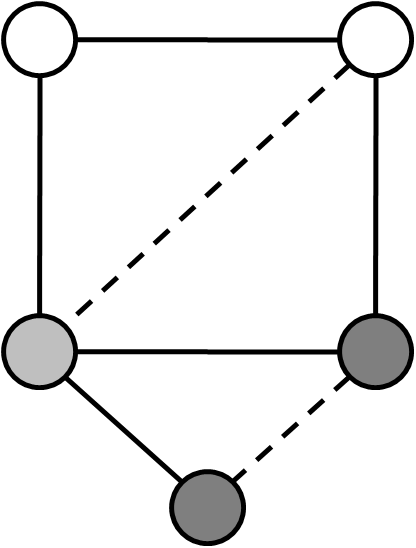}
\label{fig:lp-uniform-links}
}
\label{fig:lp}
\caption[Optional]{\textsc{Example demonstrating a general approach to link prediction:} 
The initial graph (a) is used as input to a link predictor, 
yielding a complete graph (b) where the weights $w_{ij}$ are estimated between all pairs of nodes.
The next step shows the removal of the initial (observed) links from consideration (c), 
followed by a pruning of all predicted links with a weight below some cut-off value $\delta$ (d).
The remaining predicted links are then combined with the initial links (e).  Often, the estimated 
weights on the initial and predicted links are then discarded, leaving a uniform weight
graph (f).}
\end{figure}

%% file: table-metrics.tex
\newcommand\TZZ{\rule{0pt}{2.4ex}}
\newcommand\BZZ{\rule[-1.4ex]{0pt}{0pt}}

\newcommand\TZ{\rule{0pt}{3.0ex}}
\newcommand\BZ{\rule[-1.4ex]{0pt}{0pt}}

\begin{table}
\caption{\textsc{Topology Metrics}: Summary of the most common metrics for link prediction. Notation: Let $\Gamma(x)$ be the neighbors of $x$ and $\mA$ be the adjacency matrix of $G$.}
\label{table:metrics}
\vspace{-4mm}
\begin{center}
\begin{small}
\scriptsize
\begin{tabular*}{1.00\textwidth}{@{\extracolsep{\fill}} l p{105mm}}
\toprule
\TZ \BZ  \normalsize \textbf{Local Node Metrics} & \textbf{Description}\\
\midrule
\TZ \BZ \multirow{1}{*}{Graph Distance} & Length of the shortest path between $x$ and $y$ \\ 
\TZ \BZ \multirow{1}{*}{Common Neighbors} & Number of common neighbors between $x$ and $y$, $w(x,y) = |\Gamma(x) \cap \Gamma(y)|$ \cite{newman2001clustering} \\
\TZ \BZ \multirow{1}{*}{Jaccard's Coefficient} & Probability that x and y share common neighbors (normalized), \quad $w(x,y) = \frac{|\Gamma(x) \cap \Gamma(y)|}{|\Gamma(x) \cup \Gamma(y)|}$ \cite{jaccard1901etude,salton1983introduction} \\
\TZ \BZ \multirow{1}{*}{Adamic/Adar} &  Similar to common neighbors, but assigns more weight to rare neighbors, \: 
$w(x,y) = \sum_{z \in \Gamma(x) \cap \Gamma(y)}{\frac{1}{\log|\Gamma(z)|}}$ \cite{Adamic01friendsand} \\
\TZ \BZ \multirow{1}{*}{RA} & Essentially equivalent to Adamic/Adar if $|\Gamma(z)|$ is small, \\ & $w(x,y) = \sum_{z \in \Gamma(x) \cap \Gamma(y)}{\frac{1}{|\Gamma(z)|}}$ \; \cite{zhou2009predicting} \\
\TZ \BZ \multirow{1}{*}{Preferential Attachment} & Probability of a link between $x$ and $y$ is the product of the degree of $x$ and $y$, \quad $w(x,y) = |\Gamma(x)| \cdot |\Gamma(y)|$  \cite{barabási1999emergence}\\
\TZ \BZ \multirow{1}{*}{Cosine Similarity} &  $w(x,y) = \frac{|\Gamma(x) \cap \Gamma(y)|}{\sqrt{|\Gamma(x)| \cdot |\Gamma(y)|}}$ \cite{salton1983introduction} \\
\TZ \BZ \multirow{1}{*}{Sorensen Index} &  $w(x,y) = \frac{2 \times |\Gamma(x) \cap \Gamma(y)|}{|\Gamma(x)| + |\Gamma(y)|}$ \cite{green1972latitudinal,zhou2009predicting} \\
\TZ \BZ \multirow{1}{*}{Hub Index} & Nodes with large degree are likely to be assigned a higher score, \\ &  $w(x,y) = \frac{|\Gamma(x) \cap \Gamma(y)|}{\min\{|\Gamma(x)|,|\Gamma(y)|\}}$ \cite{ravasz2002hierarchical} \\
\TZ \BZ \multirow{1}{*}{Hub Depressed Index} &  Analogous to Hub Index, $w(x,y) = \frac{|\Gamma(x) \cap \Gamma(y)|}{\max \{ | \Gamma(x)|,|\Gamma(y)| \} }$ \cite{ravasz2002hierarchical} \\
\TZ \BZ \multirow{1}{*}{Leicht-Holme-Newman} &  Assigns large weight to pairs that have many common neighbors, normalized by the expected number of common neighbors, $w(x,y) = \frac{|\Gamma(x) \cap \Gamma(y)|}{|\Gamma(x)| \cdot |\Gamma(y)|}$ \cite{leicht2006vertex} \\
\midrule
\TZ \BZ \normalsize \textbf{Global Graph Metrics} & \textbf{Description}\\
\midrule
\TZ \BZ \multirow{1}{*}{Katz} & Number of all paths between $x$ and $y$, exponentially damped by length thereby assigning more weight to shorter paths, $w(x,y) = [(\eye - \alpha \mA)^{-1}]_{xy}$ \cite{katz1953new} \\
\TZ \BZ \multirow{1}{*}{Hitting time} & Number of steps required for a random walk starting at $x$ to reach $y$ \cite{brightwell1990maximum} \\ 
\TZ \BZ \multirow{1}{*}{Commute Time} & Expected number of steps to reach node y when starting from x and then returning back to x, defined as $w(x,y) = L^{+}_{xx} + L^{+}_{yy} - 2L^{+}_{xy}$ where $\mL$ is the Laplacian matrix \cite{gobel1974random}\\
\TZ \BZ \multirow{1}{*}{Rooted PageRank} & Similar to Hitting time, but at each step there is some probability that the random walk will reset to the starting node $x$, $w(x,y) = [(\eye - \alpha \mP)^{-1}]_{xy}$ where $\mP = \mD^{-1}\mA$ \cite{page1998pagerank} \\
\TZ \BZ \multirow{1}{*}{SimRank} & x and y are similar to the extent that they are joined with similar neighbors, \\ & $w(x,y) = \frac{\sum_{u \in \Gamma(x)} \sum_{v \in \Gamma(y)} sim(u,v) }{|\Gamma(x)| \cdot |\Gamma(y)| }$ \; \cite{jeh2002simrank} \\ 
\TZ \BZ \multirow{1}{*}{K-walks} & Number of walks of length k from x to y, defined as $w(x,y) = [\mA^k]_{xy}$ \\ 
\midrule
\TZ \BZ \normalsize\textbf{Meta-Approaches} & \textbf{Description}\\
\midrule
\TZ \BZ \multirow{1}{*}{Low-rank Approximation} & Compute the rank-k matrix $\mA_{k}$ that best approximates $\mA$ (hopefully reducing ``noise''), then compute similarity over $\mA_k$ using some local or global metric \cite{eckart1936approximation,golub1970singular}\\
\TZ \BZ \multirow{1}{*}{Unseen Bigrams} & Compute initial scores using some local or global metric, then augment the scores $w(x,y)$ using values from $w(z,y)$ for nodes z that are similar to x  \cite{essen1992cooccurrence,lee1999measures}\\
\TZ \BZ \multirow{1}{*}{Clustering} & Compute initial scores using some local or global metric, discard links with the lowest scores, and then re-compute the scores on the modified graph \cite{johnson1967hierarchical,hartigan1979k} \\
\bottomrule
\end{tabular*}
\end{small}
\end{center}
\end{table}

%% file: link-interp.tex
\section{Link Interpretation}\label{sec:link-interp}

Link interpretation is the process of constructing weights, labels, or
general features for the links.  These three tasks of link
interpretation are related and somewhat overlapping.  First, link weighting is the task of assigning some weight to each link.  These
weights may represent the relevance or importance of each link, and
are typically expressed as continuous values. 
Thus the weights provide an explicit order over the links.
Second, link labeling is
similar, except that it usually assigns discrete values to each link.
This could represent a positive or negative relationship, or could be
used, for instance, to assign one of five topics to email
communication flows.  
Finally, link feature construction is the process
of generating a set of discrete or continuous features for the links.
For instance, these features might count the frequency of particular
words that appeared in messages between the two nodes connected by some
link, or simply count the number of such messages.

In a sense, link feature construction subsumes link weighting and labeling, since the weights and labels can be viewed simply as
possible link features to be discovered. However, for many
tasks it makes sense to compute one particular feature that
summarizes the relevance of each link (the weight) and/or one
particular feature that summarizes the type of each link (the label).
Such weights and labels may be especially useful to later processing,
for example with collective classification.
Moreover, the techniques used for general feature construction tend
toward simpler approaches such as aggregation and discretization,
whereas the best techniques for computing weights and labels may
involve much more complexity, including global path computations or
supervised learning.  For this reason, we treat link weighting
(Section \ref{sec:link-weight}) and link labeling (Section \ref{sec:link-label}) separately from
general link feature construction (Section \ref{sec:link-feature}).

\subsection{Link Weighting} \label{sec:link-weight}

Given the initial graph \ifullgraph, the task is 
to assign a continuous value (the weight) to each existing link in $G$,
representing the importance or influence of that link.
As previously discussed, link weighting could potentially be accomplished by
applying some link prediction technique and simply retaining the
computed scores as link weights.  For instance,
\shortciteA{lassez:latentlinks} perform link prediction and weighting
by applying singular value decomposition to the adjacency matrix, then
retaining only the $k$ most significant singular-vectors (similar to the
low-rank approximation techniques discussed in
Section~\ref{sec:topology-lp}).  They show that querying (\eg, with
PageRank) on the resultant weighted graph can
yield more relevant results compared to an unweighted graph.

Unlike with link prediction, however, most link weighting techniques
are designed to work only with links that already exist in the
graph. These techniques don't work for predicting unseen links because
they weight links based on known properties/features of the existing
links, or because they compute some additional link features that only
yield sensible results for links that already exist.

In the simplest case, link weighting can be just
aggregating an intrinsic property of links. For
example, \shortciteA{barabasi:07} defines link weights based on the
aggregated duration of phone calls between individuals in a
mobile communication network. In other cases, simply counting the
number of interactions between two nodes may be appropriate.

Thus, when link {\em features} like duration, direction, or frequency are known,
they can be aggregated in some way to generate link weights.  
If actual link {\em weights} are already known for some of the links,
then supervised methods can be used for weight prediction, using the known weights as training data.
For instance, \shortciteA{neville:aaai09} 
predict link strength within a Facebook dataset,
where stronger relationships are identified based on a user's 
explicit identification of their ``top friends'' via a popular
Facebook application.
\shortciteA{tiestrengthEricGilbert} also predict link strength for
Facebook, but form their training data from survey data collected from
35 participants (yielding strength ratings for about 2000 links).  Both of
these algorithms generate a large number (50-70) of features about each
link in the network, then learn a predictive model via regression or
some other technique such as bagged decision trees, which
\shortciteA{neville:aaai09} finds performs best among several
alternatives.  \shortciteA{tiestrengthEricGilbert} generate features
based on profile similarity (\eg, do two users have similar education
levels?) and based on user interactions
(\eg, how frequently and about what topics do two users communicate?).
They find the interaction
features to be most helpful, especially a feature based on the number
of days since the last communication event.
\shortciteA{neville:aaai09} use similar kinds of features, which they
term {\em attribute-based} and {\em transactional} features, and also
add {\em topological} features (such as the Adamic/Adar discussed in
Section~\ref{sec:topology-lp}) and {\em network-transactional} (NTR) features. 
NTR features are those that are based on communications between users (\eg,
the number of email messages exchanged) but moderated in some way by the larger
network context.  This moderation often takes the form of
normalization, for instance to dampen the influence of a node that has sent a
large number of messages to many different friends.  They find that
these NTR features are by far the most helpful for
prediction, but that many other features also contribute to
the overall predictive accuracy.

When training data with sample link weights is not available,
approaches based on a parameterized probabilistic model are
still possible.
However, since candidate link features can no longer be evaluated
against the training data, these approaches must (manually)
choose the features that they use much more carefully.  For instance,
\shortciteA{xiang:10} examine link weight prediction on two social
network datasets (Facebook and LinkedIn), but use only 5-11 features
for each link.  They hypothesize that relationship strength is a
hidden cause of user interactions, and propose a link-based latent
variable model to capture this dependence.  For inference, they use a
coordinate ascent optimization procedure to predict the strength of
each link.  Since the actual strength of each link is not known,
prediction tasks in this domain cannot directly evaluate accuracy.
However, \shortciteA{xiang:10} demonstrate that using the link
strengths produced by their method leads to higher autocorrelation
and higher collective classification accuracy when predicting user
attributes such as gender or relationship status.

A number of researchers have considered the importance of recency in evaluating
link weight, under the assumption that events or interactions that occurred
more recently should have more weight.
For instance, \shortciteA{suggestfriendsmaayanroth} propose the
``Interactions Rank'' metric for weighting a link based on the
messages between two nodes.  The formula separately weights incoming
and outgoing messages for each link, and imposes an exponential decay
on the importance of each message based on how old it is.
\shortciteauthor{suggestfriendsmaayanroth} use this metric to weight the links 
in what they call the ``implicit social network,'' where each node represents
a group of users.  They demonstrate
that this metric can be used to accurately predict users that are missing from
an email distribution list. However, the basic metric is simple to
compute and could be applied to many other tasks.

The Interactions Rank metric weights a link more
heavily if it connects two nodes that have frequently and/or recently
communicated.  Alternatively, \shortciteA{sharan:icdm08} have
considered how to weight links in a graph where the links (such as
hyperlinks or friendships) may themselves appear or disappear over
time.  In particular, they construct a summarized graph where all
nodes and links that have ever existed in the past are present.  Each
link in this new graph is weighted based on a kernel function that can
provide more weight to links that have been present more often or more
recently in the past.  They explain how to modify standard relational
classifiers to use these weighted links, and demonstrate that a
variety of kernels (including exponential and linear decay kernels)
produce weighted links that yield higher classification accuracy
compared to a non-weighted graph.  More recently,
\shortciteA{rossi:10,neville:tenc2} have extended this work to handle
time-varying attribute values, which may serve as a basis for incorporating
temporal dynamics into additional tasks.

\subsection{Link Labeling} \label{sec:link-label}

Given the initial graph \ifullgraph, the task is to construct some discrete label
for one or more links in $G$.
These labels can be used to describe the type of relationship that each link represents.
For instance, in the Facebook example, a link labeling algorithm may create
labels representing ``work'' or ``personal'' relationships.
Such labels would enable subsequent classification models to 
separately account for the influence of these different kinds of relationships.

Most prior work on link labeling has assumed that some text (such as a message) describes each link,
and has been based on unsupervised textual
analysis techniques such as Latent Dirichlet Allocation (LDA)~\shortcite{blei:03}, Latent Semantic
Analysis (LSA) \shortcite{lsa90}, or Probabilistic Latent Semantic Analysis (PLSA) \shortcite{plsa99}. 
Traditionally, these techniques have been used to assign one or more ``latent topics'' 
to each document in a collection of documents.
The ``topics'' that are formed are defined implicitly by a probability distribution over how likely 
each word is to appear, given that the topic is associated with a document.
These topics will not always be semantically meaningful, but often manual inspection reveals that
most prominent topics do represent sensible concepts such as ``advertising'' or ``government relations.'' 
However, even when such semantic associations are not obvious, 
inferring such topics for a set of links can still aid further analysis,
since the topics identify which links represent similar kinds of relationships.

These textual analysis techniques were developed with independent documents in mind, not inter-linked nodes,
but they can be adapted to label links in several ways.  For instance, 
\shortciteA{rossi:10} examined messages between developers contributing to an open-source software project.
They treat each message as a separate document, and use LDA to infer the single most likely latent topic 
for each message (i.e., a link label).  This technique could be used for any graph with textual content associated with the links.
\shortciteA{rossi:10} also go further, to consider the impact of time-varying topics and time-varying topic/word
associations, by running multiple iterations of LDA, one per time epoch.  
Using this model, they study the problem of predicting the effectiveness of different developers (nodes) in the network.
They demonstrate that the accuracy of
predictions is significantly improved by modeling the temporal evolution of the communication topics.

\shortciteA{mccallumrole:07} describe an alternative way of extending LDA-like approaches for link
labeling.  LDA is essentially a Bayesian network that models the probabilistic dependencies between documents, associated topics,
and words associated with those topics.  They propose to extend this model with the Author-Recipient-Topic (ART)
model, where the choice of topic for each document (message) depends on both the author and the recipient of the message.
Once parameters are learned for the model, inference (e.g., with Gibbs sampling) can be used to infer the most likely latent topics
for each message.  They make use of these topics to assign roles to people in an email communication network, and demonstrate
that it outperforms simpler models.

Supervised techniques can also be used for link labeling.  For instance,
\shortciteA{Taskar03linkprediction} study an academic webpage network and consider how to predict node
labels (such as ``Student'' or ``Professor'') while simultaneously predicting link labels (such as
``adviser-of'').  Given a labeled training graph, they learn a complex Relational Markov Network
(RMN) that can predict these labels and the existence of new links.
To make the link prediction tractable, only some candidate new links are considered,
such as those links suggested by a textual reference, inside a page, to some other entity in the graph.
The RMN utilizes text-based features, for instance based on the anchor text for known links or the heading
for the HTML section in which a possible link reference is found.  They demonstrate that the RMN's
joint inference over nodes and links improves performance compared to separate inference.  However,
learning and inference with RMNs can often be a significant challenge, which in practice limits the
number and types of feature that can be considered.

The RMN approach learns from some training data and then uses joint inference over the entire graph.  
A simpler supervised approach is to create a
set of features for each link and use these features for learning and inference with an arbitrary
classifier that treats each link separately.
\shortciteA{Les_predpos} study a particular form of this approach where there are only
two link labels, representing a positive or negative relationship (such as friendship vs.\
animosity).  They create link features based on the (signed) degree of the nodes involved in each
link and also based on transitivity-like properties computed from the known labels of nearby links.
They demonstrate this approach using data from Epinions, Wikipedia, and Slashdot, where users have
manually indicated positive or negative relationships to other users.  Given a network with almost
all edges labeled, the label classifier is able to predict the label (positive or negative) of a single
unlabeled edge with high accuracy.  Interestingly, they show that a classifier's predictive accuracy
for a particular dataset decreases only slightly when the classifier is trained on a different
dataset vs.\ being trained on the same dataset that is used for predictions.  They argue that
theories of balance and status from social psychology partially explain this ability of their
predictive models to generalize across datasets.  Unlike most of the other techniques discussed in
this section, this work does not make use of text-based features.  However, 
the general problem of predicting the ``sign'' of a link is related to sentiment analysis (or opinion mining) in natural language
processing~\shortcite{Godbole:sentiment,Pang08opinionmining}.
These sentiment analysis algorithms could be reformulated to predict the label (such as positive or negative) of a link given
its associated text.

Because a link between two nodes can be established based on many different kinds of relationships, there are many other
types of algorithms that could potentially be used for labeling links, even if the original
algorithm was not designed for this purpose.
For instance, Markov Logic Networks (MLNs) have been used to extract semantic networks from text,
yielding a graph where the nodes represent objects
or concepts~\shortcite{KokD08}.  This process produces relations such as ``teaches that'' or ``is written in'' between
the nodes, which could be used as link labels in further analysis.
Another example is the Group-Topic (GT) model proposed by \shortciteA{mccallumjoint:07}, which, like the
previously mentioned ART model, is a Bayesian network. 
The model is intended for graphs where two nodes (such as people) become connected when they both participate in the same ``event,''
such as both voting yes for the same political bill. 
Rather than directly labeling links (like ART), the GT model 
clusters these nodes (such as people) into latent groups 
based on textual descriptions of the events/votes.  However, the GT model also 
simultaneously infers a set of likely topics for each event, which could be used to label the implicit links between the nodes. 
The results of the model could also be used to add new nodes to
the graph that represent the latent groups that were discovered.

\subsection{Link Feature Construction}\label{sec:link-feature}

Link feature construction is the systematic construction of features on the links, typically for the purpose of improving the accuracy or understandability of SRL algorithms. 
Link feature construction can be important for many prediction tasks, but has received considerably less attention than node feature construction in the literature.
Fortunately, many of the computations that have been developed for node feature construction can also apply to link features.  
To avoid redundancy, we defer most of our analysis of feature construction to the discussion of node feature construction in Section~\ref{sec:node-feature}.
This section briefly discusses how such techniques for node feature construction can be applied to links,
then summarizes the major types of link features that can be computed.

\input{fig-link-aggr}

Section~\ref{sec:node-feature} will later describe how feature values for relational data are often based on {\em aggregating}
values from multiple nodes.  For instance, such a feature might compute the average or the most common feature value among all of the neighbors of a particular node.
Such aggregation-based features help to account for the varying number of neighbors that a node may have.
For links, aggregation is less essential, since (usually) each link has precisely two endpoint nodes.
However, aggregation can still be useful for computing features that collect information from a larger area of the graph.
For instance, in Figure~\ref{fig:link-aggr}, a link feature value is being computed for the link in the center of the subgraph (the ``target link'').
The computation considers the feature values (positive or negative signs) for all of the links that are adjacent to the target link.
In this case, the aggregation operator is {\sc Mode}, and the result is the new link feature value.  This example used link features as the input,
but node feature values (e.g., of the lightly-shaded nodes in Figure~\ref{fig:link-aggr}) could also be aggregated to form a new link feature.
In this way, all of the aggregation operators discussed for nodes in Section~\ref{sec:node-feature} can also be applied to links.

\input{fig-link-feat-taxonomy}

Figure~\ref{fig:link-feature-taxonomy} summarizes the kinds of features that can be constructed for
a link.  This figure is organized around the sources of information that go into computing a single
link feature (i.e., the inputs), rather than the details of the feature computation (such as the type of aggregation or other function used).
The bottom of the figure shows the four
types of link features, each represented by a subgraph.  In each case, the emphasized link at the
bottom of the subgraph is the target link for which a new feature value is being computed.
Each of the subgraphs shows varying amounts of information because each displays {\em only}
those features, nodes, and/or links that can be used as inputs for that kind of link feature.  

The simplest type is the {\em non-relational link feature}, which can be computed for each link
solely from information that is already known about that link.  Thus, Figure
\ref{fig:link-feature-taxonomy}A shows only the feature values which are already known for 
the target link, which can be used to construct a new feature value.
For instance, if a message is associated with each link, then a link feature could count the number of times that a
certain word occurs, or the number of distinct words.  Alternatively, if a date is associated with
the link, then a feature might compute the number of months since the link was formed.
\shortciteA{barabasi:07} computed this kind of feature when they aggregated the duration of all phone calls
between two people to form a new link feature (which they also used as a link weight).

The remaining feature types are all relational, meaning that they depend in some way on the graph (not just a single link).
First, {\em topology features} (Figure~\ref{fig:link-feature-taxonomy}B) are those that can be computed
using only the topology of the graph.  Such a feature might, for instance, compute the total number
of links that are adjacent to the target link.  Likewise,
\shortciteA{neville:aaai09} computed the clustering coefficient of a pair of linked nodes,
which measures the extent to which the two nodes have neighbors in common~\shortcite{Newman03thestructure},
as well as other topological features such as the Adamic/Adar measure discussed in Section~\ref{sec:link-weight}.  
They used these link features to help predict link strength, but they could also be used for other tasks.

Next, {\em relational link-value features} are those that are computed using the feature values of nearby links.  
For instance, Figure~\ref{fig:link-feature-taxonomy}C shows how link labels of personal (p) or work (w) might be identified from links
adjacent to the target link.  A new link feature could be formed by representing the distribution of these labels, by taking the most common
label, or (when the link features are numeric) by averaging.
\shortciteA{Les_predpos} used such link-value features when working with graphs where 
each link had a ``sign'' feature of positive or negative (as with Figure~\ref{fig:link-aggr}).
They computed features based on the signed-degree of the two nodes connected
by the target link as well as more complex measures based on other paths between these two nodes (e.g., to measure sign transitivity).

Finally, {\em relational node-value features} are those that are computed using the feature values of the nodes that are close to or are attached
to the target link.
For instance, Figure~\ref{fig:link-feature-taxonomy}D shows how node labels of conservative (C) or liberal (L) might be identified
for nodes close to the target link.  As with link-value features, these labels could be used to create a new feature value
by summarization or aggregation.  Often, only the two nodes that are directly attached to the target link are used.  For instance,
both \shortciteA{tiestrengthEricGilbert} and \shortciteA{neville:aaai09} construct link features based on the similarity of
two nodes' social network profiles.  However, the feature values of more distant nodes could also be used, for instance
to compute a new link feature based on how similar the {\em friends} of two people (nodes) are.

%% file: fig-link-aggr.tex
\begin{figure}[t]
\centering
\subfigure[Before link-aggregation]{
\includegraphics[width=2.2in]{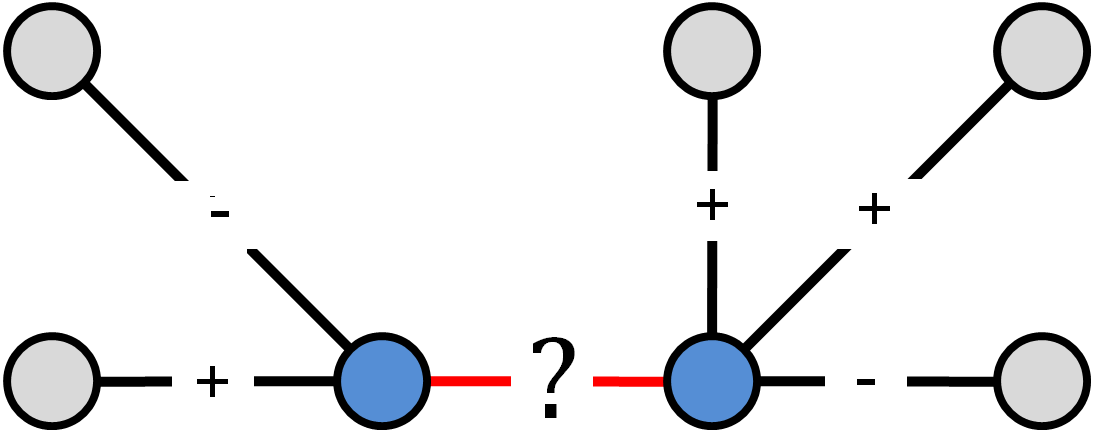}
\label{fig:link-aggr-before}
}
\hspace{0.5in}
\subfigure[After link-aggregation]{
\includegraphics[width=2.2in]{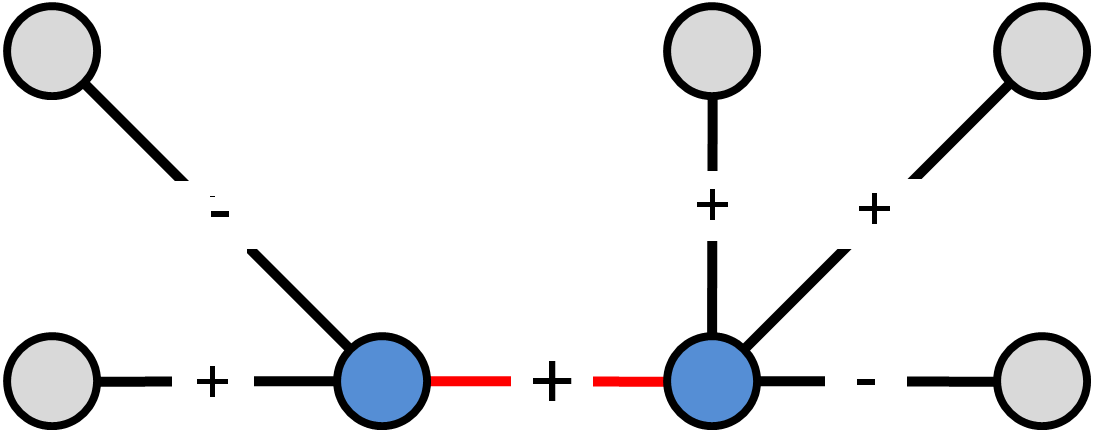}
\label{fig:link-aggr-after}
}
\caption[Optional]{\textsc{Link Feature Aggregation Example}: The figure demonstrates how an unknown link feature value can be computed by aggregating the link feature values
of surrounding links.  Here the aggregation operator is {\sc Mode}.}
\label{fig:link-aggr}
\end{figure}

%% file: fig-link-feat-taxonomy.tex
\begin{figure}[t]
\centering
\includegraphics[width=4in]{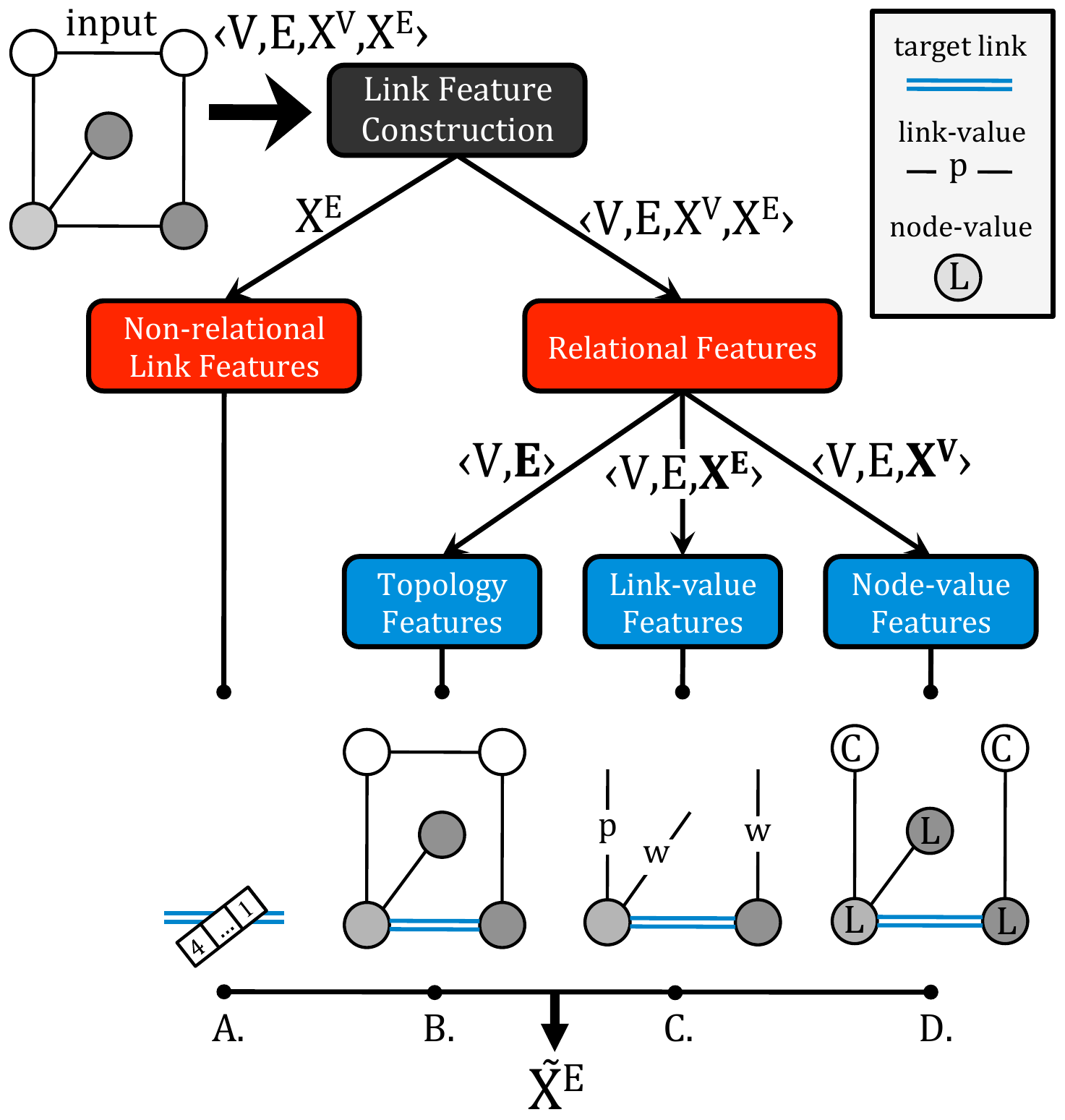}
\caption[Optional]{\textsc{Link Feature Taxonomy:}
The link feature classes are {\em non-relational features}, \textit{topology features}, \textit{relational link-value features}, and \textit{relational node-value features}. 
In the subgraphs at bottom, only the information that is potentially used by that class of link feature (i.e., nodes $V$, links $E$, node features $X^{V}$, and/or link features $X^{E}$)
is shown.  The emphasized link represents where the feature value is computed (i.e., the ``target link'').}
\label{fig:link-feature-taxonomy}
\end{figure}

%% file: node-pred.tex
\section{Node Prediction}\label{sec:node-pred}
Node transformation includes node prediction (e.g., predicting the existence of new nodes) and node interpretation (e.g., constructing node weights, labels, or features). 
This section focuses on node prediction, while Section~\ref{sec:node-interp} considers node interpretation.

Given a graph with existing nodes $V$, node prediction can be used in
two distinct ways.  First, a node prediction algorithm could be used
to discover additional nodes that are of the {\em same} type as those that are
already present in $V$.  For instance, given a set of people that
communicate via email, a simple algorithm might be used to create new
nodes that represent email recipients that are implied by the
messages, but not explicitly represented in the original graph.
Alternatively, supervised or unsupervised machine learning techniques
could be used to discover, for instance, new research papers or people from
information available on the
web~\shortcite{Craven00learningto,Cafarella08webtables}.  These techniques
are valuable, and can certainly be used to add new nodes to a graph.
However, most such work has been examined in the context of general
knowledge base construction, rather than relational learning.\footnote{
The recent work of \shortciteA{kimnetwork2011} is an exception.  Their technique uses EM to infer the existence
of missing nodes and links based on only the known topology of the graph.}

We focus on the second type of node prediction, which involves predicting nodes of a
{\em different} type than those that are already present in the graph.
These new nodes might represent locations, communities~\shortcite{kleinberg1999authoritative}, shared characteristics, social
processes~\shortcite{tang2009relational,hoff2002latent}, functions~\shortcite{letovsky2003predicting}, 
or some other kind of relationship.  
For instance, in the running Facebook 
example, a newly discovered node may represent a common interest or hobby that 
multiple people share.  
These nodes are usually referred to as ``latent nodes'' (and the nodes connected to each such node form a 
``latent group'').\footnote{Prior work sometimes refers to such nodes as ``hidden'' nodes, especially when they are thought to
represent concrete characteristics, such as geographic location, that could be measured but were, for some reason,
not observed in the data.}
The meaning of these nodes will depend upon what features and/or links were included as input to the node prediction
algorithm. For instance, including work-based friendships will lead to very different groups
than if only personal friendships are considered.

There are many advantages of this type of representation change with
regards to accuracy and understandability. For instance, nodes that
are not directly connected in the original graph but are similar in
some way become, because of the links to the new nodes, 
closer in graph space.  
Intuitively, nodes connected to a high level concept should share
some latent properties and representing that latent structure can
directly impact classification, network analysis, and many other tasks.  For instance, reducing the path
length between similar nodes enables influence from these nodes to
propagate more effectively if collective classification (CC) is performed on these nodes.
A model can still learn about and exploit these new nodes and relationships,
even if the semantic meaning of the new nodes is not precisely understood.

The most popular methods for predicting new nodes are based on
clustering, which in our context means the grouping of nodes such that
nodes within a group are more similar to each other than they are to
the nodes in other groups.  Typically, one new node is created for
each group, and then links are added between each existing node and its
corresponding group node (see right side of Figure~\ref{fig:node-feature-duality}).  Some techniques may also associate each
node with multiple groups, with link weights representing the affinity
to each group.

\input{fig-node-feature-duality}

When new groups are discovered, whether via clustering or via some
other technique, an alternative to creating new nodes and links is to
simply add new feature(s) to each node that represent the group
information.  The left side of Figure~\ref{fig:node-feature-duality}
demonstrates this alternative.  For instance, a new node feature
might represent having running as hobby, or it may simply represent
belonging to discovered group \#17, which is of unknown meaning.
\shortciteA{popescul:04} use the CiteSeer dataset to demonstrate that
this technique can derive features that can improve
predictive accuracy.
An advantage of this approach, as opposed to
adding new nodes, is that it potentially enables simpler,
non-relational algorithms to make use of the new information.
A potential disadvantage, though, is that it also does not allow for algorithms such as CC to propagate
influence between newly connected nodes, as discussed above.  
However, some such methods use this general strategy to 
generate much larger numbers of latent features that can be used for classification \shortcite{tang2009relational,menon2010predicting}.
Tang \& Liu demonstrate that, in some cases, the resultant large number of link-based features 
may make collective inference unnecessary for obtaining good accuracy.
Naturally, whether the information discovered from these clusterings
is best represented via new nodes or new features will depend upon the dataset
and the inference task.
In this
section, for simplicity we will discuss each algorithm assuming that new nodes will be
created (even if the algorithm was originally described in terms of
creating new features).  

As with our discussion of link prediction, we organize our discussion
around the kinds of information that are used for prediction.
Section~\ref{sec:feature-np} discusses non-relational (attribute-based)
node prediction, Section~\ref{sec:topology-np} discusses
topology-based node prediction, and Section~\ref{sec:hybrid-np}
discusses hybrid approaches that use both the node feature values and
the topology of the graph.

\subsection{Non-relational (Attribute-based) Node Prediction} \label{sec:feature-np}

There are many clustering algorithms that can be used to cluster existing nodes 
using only their non-relational features (attributes), which can then be used to add new nodes to a graph.
The two primary types are hierarchical clustering algorithms (e.g., agglomerative
or divisive clustering) and partitioning algorithms 
such as k-means, k-medoids~\shortcite{berkhin2006survey,zhu2006semi},
EM-based algorithms, and self-organizing maps~\shortcite{kohonen1990self}.
We do not discuss these algorithms further since they have been well studied
for non-relational data and can be easily applied to relational data if clustering
based only on attribute values is desired.

\subsection{Topology-based Node Prediction} \label{sec:topology-np}

The techniques described in this section link existing nodes to one or more new nodes (i.e., latent groups),
based only on the original link structure of the graph.  
In most cases, finding this grouping depends upon computing
some kind of similarity metric between every pair of nodes.  Two key questions thus serve to identify
these techniques.  First, what kind of similarity metric should be used?  Second, how should the metric
be used to predict groupings?  We address each question in turn.

\smallskip
\noindent
{\bf Types of metrics for group prediction:} Any type of topology-based link weighting metric (see Table~\ref{table:metrics})
could conceivably be used for latent node prediction.  A metric will be suitable so long as it produces high values
for pairs of nodes that should belong to the same group and lower values for other pairs.
For instance, a high value of the Katz metric (see Section~\ref{sec:topology-lp}) indicates that two 
nodes have many short paths between them, and
thus may belong to the same group.
Metrics representing distance rather than similarity can also be used after negating the metric.
For instance, \shortciteA{girvan:07} focus on detecting community
structure by extending the concept of node-betweenness to links.  
Intuitively, if a network contains latent groups that are only loosely
connected by a few intergroup links, then all shortest paths between
different groups must go along these links.  These links that
connect the different groups are assigned a high link-betweenness
value (which corresponds to a {\em low} similarity value).  The underlying group structure can then trivially be revealed
by removing the links with highest betweenness.

This idea of using link-betweenness for relational clustering has been
extended in a number of directions.  For instance,
\shortciteA{newman2004finding} introduced random-walk betweenness,
which is the expected number of times that a random walk between a
pair of nodes will pass down a particular link.  In addition,
\shortciteA{radicchi2004defining} proposed using a link-based
clustering coefficient metric.  They showed that this metric performs
comparably to the original link-betweenness metric of Girvan \&
Newman, but is much faster because it is a local graph measure instead
of a global graph measure.

~\shortciteA{zhou2003distance} describes a new metric, the ``dissimilarity index,'' 
which can be computed as follows.  For
 each node $i$, compute a vector $\vec{d_i}$ where each value $d_{ij}$
 represents the distance from node $i$ to node $j$ (Zhou
 measures distance based on the average number of steps needed for a
 random walk starting at node $i$ to reach node $j$, but any distance
 metric could be used).  If nodes $i$ and $k$ are very similar, they
 should have very similar distance vectors.  Thus, the dissimilarity
 index for nodes $i$ and $k$ is defined based on a Euclidean-like
 distance computation between vectors $\vec{d_i}$ and $\vec{d_k}$.  Zhou
 demonstrates that this technique outperforms the link-betweenness
 approach of Girvan \& Newman for some random modular networks.

Relatively simple metrics can often lead to useful results.  For
instance, \shortciteA{ravasz2002hierarchical} used a simple
clustering coefficient metric to study metabolic networks.  Their
study reveals that the metabolic networks of forty-three organisms are
organized into many small, highly-connected modules.  Furthermore, they
find that for E. coli, the hidden hierarchical modularity
closely overlaps with known metabolic functions.

\smallskip
\noindent
{\bf Using the metrics for group prediction:} The simplest
techniques for identifying new groups is to perform some kind of
hierarchical clustering.  For instance, after similarities or weights
have been computed for every pair of nodes, all links can be removed
from the graph.  Next, the weighted links are placed between the nodes
one by one, ordered by their weights.  The intuition is that varying
degrees of clusters are formed as more links are added.  In
particular, this approach forms a hierarchical tree where the leaves 
represent the finest granularity of clustering where every node
is a separate cluster. As we move up the tree larger clusters
are formed, until we reach the top where all the nodes are joined in
one large cluster.  This type of hierarchical approach was used by
~\shortciteA{zhou2003distance}.  \shortciteA{girvan:07} use a similar
strategy, but start instead with the original graph and iteratively
remove the less similar links from the graph to reveal the underlying
community structure.  A challenge with these approaches, as with
clustering in general, is to select the appropriate number of final
clusters, which corresponds to selecting a level in the clustering
tree.

Spectral clustering~\shortcite{dhillon2001co,ng2001spectral,kamvar2003spectral}
can also be used for group identification.
Spectral clustering relies upon computing a similarity matrix $S$ that
describes all the data points, then transforming the matrix in a way that
yields a new matrix $U$ where clustering the rows of $U$ using a simple
clustering algorithm (such as k-means) can trivially identify the
interesting groups in the data.  The matrix transformation has several
variants, but involves computing some kind of Laplacian of $S$, then
computing the eigenvectors of the resultant matrix and using those
eigenvectors to represent the original data.  The motivation for this
transformation can be seen as identifying good graph cuts in the original
graph (those that yield good separations of highly-connected nodes
into groups) or as identifying those nodes that are closely related in
terms of random walks; see \shortciteA{von2007tutorial} for an overview.  Spectral
clustering was originally applied to non-relational data, but, as with
the hierarchical techniques described above, it can be applied to relational data by using
link-based metrics for computing the similarity matrix.  For instance,
\shortciteA{neville:icdm05} use the node adjacency matrix and the spectral
clustering technique described by \shortciteA{shi2000normalized} to identify
latent groups in their graphs.  
They show that this technique enables simpler inference (since each group can be handled separately),
and ultimately yields more accurate classification compared to approaches that ignore the group structure.
\shortciteA{tang2011leveraging} also use spectral clustering on the link graph, but 
do so in order to create
a much larger number of latent features that are then used to learn a supervised classifier.  
Unlike the latent groups of Neville and Jensen, this technique allows each node to be associated with more than one cluster in
the output of the spectral clustering, which Tang \& Liu claim leads to improved classification accuracy.
Spectral clustering can also be used with more complex similarity metrics, as
described in the next subsection.

Techniques borrowed from web search can also be useful for node prediction.
For instance, given the adjacency matrix $\mA$ for a webpage graph, 
the \textsc{Hits} algorithm~\shortcite{kleinberg1999authoritative}
computes the first few eigenvectors of $\mA\mA^{T}$ and $\mA^{T}\mA$, which represent the most
authoritative nodes (the ``authorities'') as
well as prominent nodes that point to them (the ``hubs'').  Normally, this
algorithm is used to find only the single most prominent ``community''
of authorities and hubs (to assist with a web search), but secondary
communities can be discovered by also considering the
non-principal eigenvectors of $\mA\mA^{T}$ and $\mA^{T}\mA$~\shortcite{GibKleRag98}.  A node prediction algorithm
could then treat each such community as a latent group and add a new
node and links to represent this group.  
These techniques may be especially useful for detecting patterns of influence in a graph and adding more
explicit links to represent this influence.

\subsection{Hybrid Node Prediction} \label{sec:hybrid-np}

The techniques in the previous section added new nodes to the graph, often based on clustering, 
using only the topology of the graph.  In principle, a technique that also used the nodes' attributes
should produce more meaningful latent groups/nodes.  This section considers how to add such attribute
information to techniques for node prediction.

A simple approach is to define some kind of similarity metric that combines non-relational and topology-based
similarity into a single value, then provide that similarity metric to one of the previously mentioned
clustering algorithms.  
For instance, \shortciteA{neville:tr04} use a weighted combination of attribute and link information
$$S(i,j) = \alpha \cdot \frac{1}{k} \sum_{k} s_{k}(i,j) + (1 - \alpha) \cdot l$$
\noindent 
as a metric, where $s_{k}(i,j) = 1$ iff nodes $i$ and $j$ have the same value for the $k$th attribute, and
$l = 1$ iff a link exists between $i$ and $j$.  Here the constant $\alpha$ controls the relative importance of the
attributes vs.\ the links.
They use this metric with the NCut spectral clustering technique to add new nodes to the graph, and demonstrate that these
additional nodes increase the performance of relational classification.
A similar weighted combination of attribute and link-based similarity is used by ~\shortciteA{bhattacharya2005relational} for entity resolution.

Attribute-based information can also be incorporated on an ad-hoc basis.
For instance, \shortciteA{adibi2004kojak} describe a group finding algorithm
where an initial seed set of clusters is formed based on a handcrafted set of logical rules,
and then these clusters are refined using a probabilistic system based on mutual information.
In their system, the logic-based component primarily uses the attributes about each node (person),
while the probabilistic system primarily uses the links that describe connections between the people.
However, both components make some use of both attributes and links.

A more principled approach is to define some kind of generative model 
that represents the dependence of the observed attributes and links on some latent group nodes,
then use that model to estimate group membership.
For instance, \shortciteA{kubica2002stochastic} define a
generative model where each node belongs to one or more groups, and
group members tend to link to each other.  In particular, they use a
group membership chart to track whether each node belongs to each
group, and do a local search over possible states of the chart (using
stochastic hill climbing) to try to identify membership changes that
would better explain the known data. At each step, maximum likelihood is used to
estimate the parameters of the model.  They demonstrate the usefulness
of their technique on news articles, webpages, and some synthetic
data.

Generative models can also be used with more sophisticated inference.
For example, \shortciteA{taskar2001probabilistic} treats group membership as a latent variable
and then uses loopy belief propagation to implicitly perform a clustering of the nodes.
Likewise, Mixed Membership Relational Clustering
(MMRC)~\shortcite{long2007probabilistic} uses EM variants to
estimate group memberships.  In particular, it uses a first round of
hard clustering (where each object is assigned to exactly one
cluster), following by a round of soft clustering where continuous
strength values are associated with each membership assignment.
Mixed membership stochastic blockmodels~\shortcite{blei:jmlr08} also assign continuous group membership values
to each node, but use only topological information (not attributes) for their group assignments and 
use variational inference techniques with the generative model.
Finally, \shortciteA{Long06spectralclustering} demonstrates how node clustering can be
performed instead using spectral clustering, and focuses particularly on how to simultaneously 
cluster multiple types of nodes (e.g., to simultaneously cluster web pages and web users into two
distinct sets of groups).

Most group prediction algorithms assume that links are more likely to connect nodes that belong to the same group.
An exception is \shortciteA{anthony:07}, which also uses a generative model where the links and attributes depend on 
some latent group memberships, but where some types of links are more likely to occur between 
nodes that do {\em not} belong to the same group.  
For instance, they note that if groups in a social network are defined by gender, then a link
representing ``dating'' is more likely to connect two nodes from different groups.

\subsection{Discussion}\label{sec:discussion-np}

Most of the techniques described above produce a single clustering of the nodes, usually 
based on assigning every node to a single group.  
In contrast, multi-clustering is an emerging research area that aims to provide multiple orthogonal clusterings of complex data~\shortcite{strehl2003cluster,topchy2004analysis}.   
For instance, individuals in Facebook might be clustered in multiple ways where latent node types might represent friend groups, work relations, socioeconomic status, locations, or family circles. 
A type of multi-clustering is performed by \shortciteA{mccallumrole:07} where latent nodes are created based on roles and topics.
In addition, \shortciteA{kok:icml07} propose Statistical Predicate
Invention (SPI), a node transformation approach based on Markov Logic
Networks~\shortcite{richardson2006markov}. SPI clusters nodes, features and
links forming the basis for the prediction of predicates (or potential
nodes).  SPI considers multiple relational clusterings based on the
observation that multiple distinct clusterings may be necessary to,
for instance, group individuals based on their friendships and their
work relationships.  They demonstrate that MLN inference can estimate
these clusters and improves performance compared to two simpler
baselines.  A similar node prediction approach applies MLNs for role
labeling \shortcite{mlnrolelabeling}.

Node {\em deletion} may also be useful in some cases.  For
instance, node deletion might be beneficial for removing outdated or
spurious nodes from the graph.  Alternatively, there may be multiple
nodes that represent the same real-world object or concept, in which
case deletion for the purposes of entity resolution can be
important~\shortcite{pasula:03,bhattacharya:07,Singla06entityresolution}.

\begin{figure}[t]
\centering
\includegraphics[width=3in]{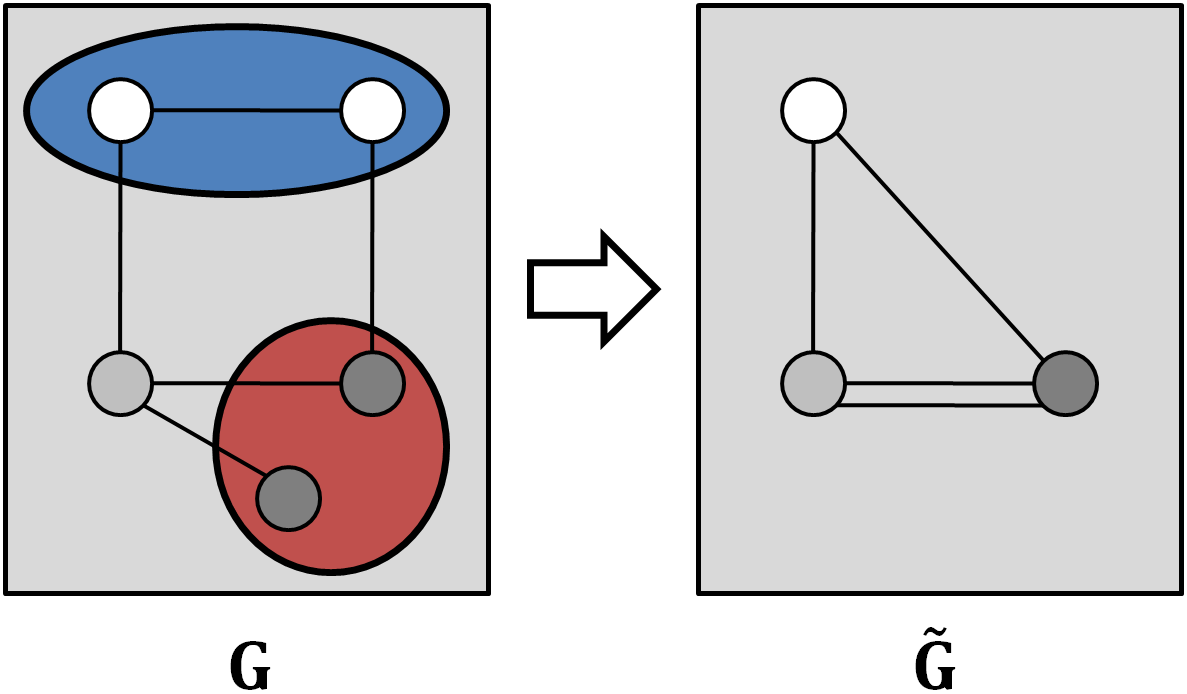}
\caption[Optional]{\textsc{Lifted graph representation:} The initial graph G is clustered and transformed into a lifted graph representation $\tilde{G}$. The lifted graph representation is created by clustering nodes, links, or both.}
\label{fig:lifted-graph}
\end{figure}

Finally, node representation changes can be used to not only to improve accuracy, but also to yield graphs that can
be processed more efficiently or that have other desirable properties.
Section~\ref{sec:topology-np} already discussed how \shortciteA{neville:icdm05} used the addition of latent nodes
to enable simpler inference.  Another possibility is the creation of ``super-nodes'' that represent more than one of the original nodes.
For instance, Figure~\ref{fig:lifted-graph} demonstrates how five original nodes can, after clustering, be collapsed into three super-nodes,
yielding a ``lifted graph'' representation.
This kind of representation change can be used for more efficient inference in Markov Logic Networks (see Section~\ref{sec:node-feature}) 
and for network anonymization (see Section~\ref{sec:anon}).

%% file: fig-node-feature-duality.tex
\begin{figure}[t]
\centering
\includegraphics[width=3in]{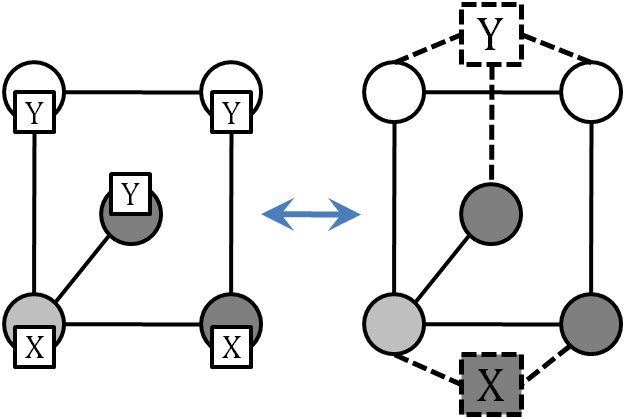}
\caption[Optional]{\textsc{Alternative representations for newly predicted groups:}
The left figure shows how a new feature (with value X or Y) could be added to each node, while the right figure
demonstrates the creation of two new nodes to represent the groups.}
\label{fig:node-feature-duality}
\end{figure}

%% file: node-interp.tex
\section{Node Interpretation}\label{sec:node-interp}

Node interpretation is the process of constructing weights, labels, or general features for the
nodes.  As with the symmetric tasks for link interpretation, node weighting seeks to assign a
continuous value to each node, representing the node's importance, while node labeling seeks
to assign a discrete value to each link, representing the type, group, or class of a node.  Likewise,
node feature construction is the process of systematically generating general-purpose node features based on, for
instance, aggregation, dimensionality reduction, or subgraph patterns.

As discussed in Section~\ref{sec:link-interp} for links, node feature construction could be viewed as subsuming
node weighting and node labeling, since general feature construction could always be used
to construct feature values that are treated as weights or labels for the nodes.
In practice, however, the techniques used tend to be rather different.   For instance, PageRank is often
used for node weighting and supervised classification is often used for node labeling, but these
techniques are rarely used for general feature construction.
Nonetheless, for node interpretation (more so than with link interpretation) there is some substantial overlap between the techniques actually used
for weighting and labeling vs.\ those used for general feature construction.  Below, we first discuss 
node weighting in Section~\ref{sec:node-weight} and labeling in Section~\ref{sec:node-label}.  Section~\ref{sec:node-feature}
then discusses node feature construction, mentioning only briefly the relevant techniques that were previously discussed
for weighting and labeling.

\subsection{Node Weighting}\label{sec:node-weight}

Given the initial graph \ifullgraph, ~the task is to assign a continuous value (the weight)
to each existing node in $G$, representing the importance or influence of that node.
Node weighting techniques have been used for information retrieval, search engines, social network analysis, 
and many other domains as a way to discover the most important nodes with respect to some defined measure. 
As with node prediction they can be classified based on whether they use only the node attributes, only the graph topology, or both to construct
a weighting.  

\medskip
\noindent
{\bf Non-relational (attribute-based) node weighting:}
The simplest node weighting techniques use only the \textit{node features} \infeatures (i.e., the attributes).  For instance,
nodes representing documents might be weighted based on the number of query-relevant words they contain, while nodes representing
companies might be ranked based on their gross annual sales.
Many more sophisticated strategies have also been considered.  
For instance, Latent Semantic Indexing \shortcite{lsa90} can be used to identify the most important semantic concepts in a corpus of text, then 
nodes can be ranked based on their connection to these concepts.
These methods have been extensively applied to quantify or rank the importance of scientific publications~\shortcite{egghe1990introduction}.
However, because these techniques have been extensively studied elsewhere and also ignore graph structure (such as citations), we do not discuss
them further here.

\medskip
\noindent
{\bf Topology-based node weighting:}
Several node weighting algorithms that use only the topology of the graph were developed to support early search engines.
Examples of this kind of algorithm include PageRank \shortcite{page1998pagerank}, \textsc{Hits} \shortcite{kleinberg1999authoritative},
and SALSA \shortcite{lempel2000stochastic}. 
Each of these algorithms rank the relative importance of web sites, conceptually based on some kind of eigenvector analysis~\shortcite{langville2005survey},
though in practice iterative computation may be used. 
For instance, PageRank models the web as a Markov Chain and is implemented by systematically computing the principal eigenvector of $\lim_{k\to\infty} \mA^{k}e$ where $\mA$ is the adjacency matrix and $e$ is the unit vector.
\textsc{Hits}, as previously described, instead computes the principal eigenvectors of $\mA\mA^{T}$ and $\mA^{T}\mA$.
These algorithms continue to be very important for webpage ranking, but can also be applied to many other kinds of graphs~\shortcite{kosala2000web}.

In social network analysis, the objective of topology-based node weighting is typically to
identify the most influential or significant individuals in a social network. There have been a
variety of centrality measures devised that use the local and global network structure to
characterize the importance of individuals~\shortcite{wasserman1994social}.
Examples of these metrics include node degree, clustering coefficient~\shortcite{watts1998collective}, betweenness~\shortcite{betweennessfreeman}, closeness (i.e., distance/shortest paths), eigenvector centrality~\shortcite{bonacich2001eigenvector}, and many others~\shortcite{jackson2008social,newman2010networks,sabidussi1966centrality}.  
In addition, \shortciteA{white2003algorithms} considered how to compute {\em relative} node rankings, i.e., rankings relative to a set
of particularly interesting nodes.  They show how to compute such relative rankings both for metrics based on shortest paths as well as for Markov chain-based techniques (e.g., to produce ``PageRank with priors''). In addition, some of the similarity metrics described in Table~\ref{table:metrics} can alternatively be formulated for computing weights on nodes.

More recently, node weighting techniques have been extended to measure the relative importance of
nodes in temporally-varying data.  For instance, both \shortciteA{kossinets2008structure} and
\shortciteA{tang2009temporal} define notions of temporal distance based on an analysis of how
frequently information is exchanged between nodes.  This information can be used to define a range
of new graph metrics, such as global temporal efficiency, local temporal efficiency, and the
temporal clustering coefficient~\shortcite{tang2009temporal}. More recently,
\shortciteA{tang2010analysing} define notions of temporal betweenness and temporal closeness.  They
argue that incorporating temporal information with these metrics provides both a better understanding of dynamic processes in the network and
more accurately identifies the most important nodes (people).  All of these metrics primarily concern
networks that have time-varying interactions (e.g., communications between people), but they could also
be applied to other types of data with intermittent interactions between nodes or where nodes/link
join and leave the network over time.  Some of these metrics also apply to links, and could possibly be used
to improve link prediction algorithms.

\medskip
\noindent
{\bf Hybrid node weighting:} There are also \textit{hybrid} node weighting approaches that
use both the attributes and the graph topology~\shortcite{bharat1998improved,cohn2001missing}.  For
instance, there are various approaches that modify \textsc{Hits}
\shortcite{chakrabarti1998automatic,bharat1998improved} and PageRank \shortcite{haveliwala2003topic} to
construct node weights based on both content and links.  Topic-Sensitive PageRank
\shortcite{haveliwala2003topic} seeks to compute a biased set of PageRank vectors using a set of
representative topics.  Alternatively, \shortciteA{kolda2005higher} propose TOPHITS, a hybrid approach
that adds anchor text (i.e., the clickable text on each hyperlink) to the adjacency matrix representation used by
\textsc{Hits}.  They then use a higher-order analogue of SVD known as Parallel Factors (PARAFAC)
decomposition \shortcite{harshman1970foundations} to identify both the key topics in the graph as well as the most
important nodes.
Other hybrid approaches have been proposed such as
SimRank~\shortcite{jeh2002simrank}, Topical methods
\shortcite{haveliwala2003topic,nie2006topical,kolda2006tophits}, Probabilistic HITs
\shortcite{cohn2000learning}, and many others \shortcite{Richardson02theintelligent,lassez:latentlinks}.  
Section~\ref{sec:joint-disc} discusses further relevant work in the context of joint node and link transformation techniques.

Recently, node weighting approaches have been applied in Adversarial Information
Retrieval (AIR) to detect or moderate the influence of spam web sites.
Typically, these techniques produce weights using both the topology of the graph and some other information, but not necessarily
the kind of attribute information that is used by the techniques discussed above.
For instance, TrustRank~\shortcite{gyngyi2004combating} is based on PageRank and uses a set of trusted sites evaluated
by humans to propagate the trust to other locally reachable sites.  On the other hand, SpamRank~\shortcite{benczr-spamrank}
measures the amount of undeserved PageRank by analyzing the backlinks of a site. There are other
algorithms that try to identify link farms and link spam
alliances~\shortcite{wu2005identifying}, given a seed set of known link farm pages. Among these AIR methods, TrustRank is the most widely known
but suffers from biases where the human-selected set of trustworthy sites may favor certain communities
over others.

\subsection{Node Labeling}\label{sec:node-label}
Given the initial graph \ifullgraph, the task is to assign some discrete label for some or all of the nodes in $G$.
We first discuss labeling techniques based on classification, then consider unsupervised textual analysis techniques.

In many cases, node labeling  may be considered an end in itself.  For instance, in our running Facebook example, the stated goal is to 
predict the political affiliation of each node where that label is not already known. In other cases, however, node labeling is
more properly understood as a representation change that supports the desired task.  For instance, for some definitions of anomalous link
detection~\shortcite{rattAnomKDDExpl05}, having estimated node labels would allow us to identify links between nodes whose labels indicate
they should rarely, if ever, be connected.  Alternatively, for some datasets estimating node labels may enable us to subsequently partition the data 
based on node type, enabling us to learn more accurate models for each type of node.

Even when node labeling is the final goal, as with our Facebook example, intermediate label
estimation may still be useful as a representation change.  In particular,
\shortciteA{kou2007stacked} describe a ``stacked model'' for relational
classification that relabels the training set with estimated node labels using a non-relational
classifier.  They then use these estimated labels to learn a new classifier (one that uses both
attributes and relational features), and use the new classifier to perform relational classification
on the test graph.  This approach yields high accuracy, comparable to that of much more complex
algorithms for collective classification (CC).  \shortciteA{fast2008stacked} analyze this result and
discuss how it can be explained by a natural bias in most
CC algorithms:  training is performed with the given node labels but the inference depends in part
on estimated labels~\shortcite{aha:09}.  Stacked models compensate for this bias by instead
training with the relabeled (estimated) training set. In addition, inference with the new
classifier needs only a single pass over the test graph, yielding much faster inference than CC techniques
like Gibbs sampling or belief propagation.
More recently,
\shortciteA{maes2009simulated} extend these ideas of node relabeling in order to generate a larger training
set via multiple simulated iterations of classification.  They show that in some cases this
approach can outperform stacked models and other CC algorithms like Gibbs sampling.

Thus, there are multiple reasons for creating new labels for the nodes in a graph.  This labeling can be accomplished by relational-aware algorithms like those described above as well as by earlier
algorithms used for relational or collective
classification~\shortcite{chakra:98,neville:srl00,taskar2001probabilistic,getoorlinkbased2003,macskassy:03}.  Node labeling can of course also be done by traditional, non-relational algorithms such as SVM, decision
trees, kNN, logistic regression, and Naive Bayes, among various others
\shortcite{lim2000comparison,Michie94machinelearning,burges1998tutorial,cristianini2000introduction,joachims1998text}. These
methods simply use features \nfeatures and do not exploit topology or link-structure.

The above techniques all assign new labels via supervised learning.  Labels can also be
assigned via unsupervised techniques for textual analysis.  
There are many networks in the real-world that contain
textual content such as social networks, email/communication networks, citation networks, and many others.
Traditional textual analysis models such as LSA~\shortcite{lsa90}, PLSA~\shortcite{plsa99} and LDA~\shortcite{blei:03} can be used to assign each node a topic representing an abstraction of the textual information.
More recent techniques such as Link-LDA \shortcite{erosheva2004mixed} and Link-PLSA
\shortcite{cohn2001missing} aim to incorporate the link structure into the traditional techniques in
order to more accurately discover a node's type.\footnote{The names for Link-LDA and Link-PLDA come from \shortciteA{nallapati2008joint}, not from the original papers describing
the techniques.}
In particular, \shortciteA{cohn2001missing} demonstrate that their technique can produce more accurate node labels
than techniques that use only the node attributes or only the link topology.
There have also been more sophisticated topic models that
have been developed for specific tasks such as social tagging~\shortcite{topicperspectivecaimeilu} or
temporal data~\shortcite{huhfienbergtopic,danhetopicstreams}.

\subsection{Node Feature Construction}\label{sec:node-feature}

Node feature construction is the systematic construction of features for the nodes, typically for the
purpose of improving the accuracy or understandability of SRL algorithms.  Feature construction is the
most common relational representation change, and is very frequently done before performing a task
such as classification.  For instance, before performing CC to classify the nodes in our example Facebook political 
affiliation task, we are likely to compute some new features representing the information about each node (\eg, age bracket?) and the
known information about each node's neighbors (\eg, how many are liberal?).

Different techniques for node feature construction have been described by many previous investigations,
though feature construction was not necessarily the focus of many of those investigations.  In this section, we summarize and
explain the different aspects of feature construction.  In particular, Section~\ref{sec:rel-feature-space} presents and discusses a
taxonomy of features based on what kinds of {\em inputs}, such as topology information or link
feature values, they use for computing the new feature values.  Next, Section~\ref{sec:rel-ops}
describes the possible {\em operators}, such as aggregation or discretization, that can be applied
to these inputs.  Finally, Section~\ref{sec:search-select-features} examines how to
perform automatic {\em feature search and selection} to support a desired computational task.

\subsubsection{Relational Feature Inputs}\label{sec:rel-feature-space}

\begin{figure}[t]
\centering
\includegraphics[width=4in]{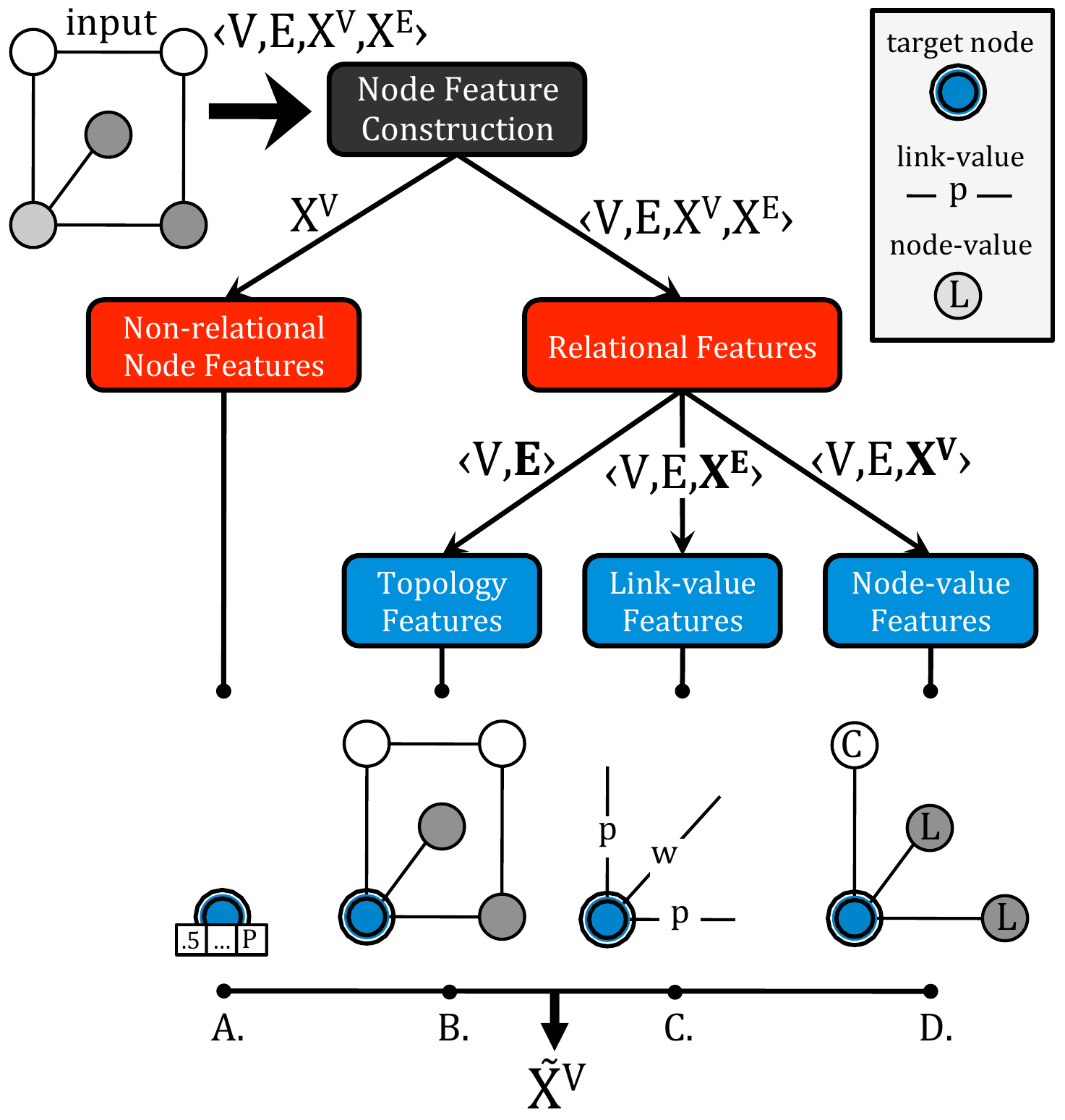}
\caption[Optional]{\textsc{Node Features Taxonomy based on Inputs Used:} The classes of node features are non-relational features, topology features, relational link-value features, and relational node-value features. 
These classes are defined with respect to the relational information used in the construction of the features (i.e., nodes $V$, links $E$, node features $\mathbf{X}^{V}$, link features $\mathbf{X}^{E}$).
The double-lined ``target'' node represents where the new feature value is being computed.
Parts C and D show only a single feature value for each link or node for simplicity, but in general more than one such feature may exist and be used.
}
\label{fig:node-feature-space-taxonomy}
\end{figure}

A node feature can be categorized according to the types of information that it uses for computing feature values.
The possible information to use includes the set of nodes $V$ or links $E$, the node
features $\mathbf{X}^{V}$, and the link features $\mathbf{X}^{E}$. 
Figure~\ref{fig:node-feature-space-taxonomy} shows our taxonomy of node features based on which of these sources of information (the ``inputs'')
they use.  This taxonomy is consistent with some distinctions that have been previously made in the literature
(e.g., between non-relational and relational features), but to the best of our knowledge this more complete taxonomy has
never been previously described.  
The taxonomy consists of four basic types:  non-relational features and three types of relational features 
(topology features, relational link-value features, and relational node-value features).  Below we describe and give examples of each.

\begin{list}{$\diamond$}{}
\item \textbf{Non-relational Features}: 
A node feature is considered a non-relational feature if the value of the feature for a particular node is computed using only the
non-relational features (i.e., attributes) of that node, ignoring any link-based information.
For instance, Figure~\ref{fig:node-feature-space-taxonomy}A shows a node and the corresponding node's feature vector. 
A new feature value might be constructed from this vector using some kind of dimensionality reduction, by adding together several feature values,
by thresholding a particular value, etc.

\item \textbf{Topology Features}: 
A feature is considered a topology-based feature if values of the feature are computed using only the nodes $V$ and links $E$,
ignoring any existing node and link feature values.
For instance, in Figure~\ref{fig:node-feature-space-taxonomy}B, a new feature value is being computed for the node in the bottom left of the figure (the ``target node''),
using only the topological information shown.  In particular, the new feature value might count the number of adjacent nodes, or count how many shortest paths in the graph pass through the target node.

\item \textbf{Relational Link-value Features}: 
A feature is considered a relational link-value feature if the feature values of the {\em links} that are adjacent to the
target node are used for computing the new feature.  
Typically, some kind of aggregation operator is applied to these values, such as count, mode, average, proportion, etc.
For instance, in Figure~\ref{fig:node-feature-space-taxonomy}C, the values on the links shown represent communication topics 
(work or personal), and a new link-value feature might compute the mode of these values (p).
Usually this computation will include only the links directly connected to the target node, but links a few hops away could also be used.

\item \textbf{Relational Node-value Features}: 
A feature is considered a relational node-value feature if the feature values of nodes linked to the target node are used in
the construction.  Links are used only for identifying these nodes, although nodes more than one hop away from the target node may also be included.
For instance, Figure~\ref{fig:node-feature-space-taxonomy}D shows the feature values of adjacent nodes (C or L) which could, for instance, be used
to compute a new node-value feature based on the mode (L) of those values. Alternatively, one feature might count the number of adjacent ``C'' nodes and another might count the number of adjacent ``L'' nodes.
 
\end{list}{}{}

Feature computation may also be applied recursively.  
For instance, the ReFeX system~\shortcite{refex2011kdd} first computes features for every node based on their degree (a topology-based feature),
then considers recursive combinations of these features (such as the mean out-degree of a node's neighbors).  
\shortciteauthor{refex2011kdd} show that such recursive features can often improve classification accuracy for datasets where the network structure
is predictive.  Alternatively, a topology-based feature such as betweenness might be
computed, then a relational node-value feature might compute the
average betweenness of the nodes that are neighbors of the target and have a label of ``C.''  This is
an example of a hybrid feature that uses both node-value and topology-based information.  

Another interesting aspect of relational features is the potential for feature value
re-computation.  In particular, many techniques for collective classification involve computing a
node feature (such as the number of neighbors currently labeled ``C'') where that feature depends on
other feature values that are estimated (e.g., the predicted node labels) and thus may change \shortcite{jensen:kdd04,sen2008collective}.
In addition, \shortciteA{mcdowell2010meta} describe features that have a similar need for recomputation, because
the ``meta-features'' they use depend upon the estimated label probabilities for each node in the
neighborhood of the target node.  In contrast, this kind of feature re-computation has much less
applicability for non-relational data, where the nodes are assumed to be independent of each other.
However, it can occur with techniques such as semi-supervised learning or co-learning.

\subsubsection{Relational Feature Operators}\label{sec:rel-ops}

The previous section described features according to the different
kinds of {\em inputs} that they use during feature value computation,
whereas this section describes the different {\em operators} that can be
used for this computation.  Table~\ref{table:rel-operators} summarizes
these operators.  In some cases, an operator can be used for many
different types of relational input.  For instance, aggregation operators can be
computed using the graph topology, relational node-value inputs,
and/or relational link-value inputs, as indicated by the appropriate checkmarks
in Table~\ref{table:rel-operators}.  In contrast, 
path or walk-based operators generally use only the graph topology; 
for these operators, the lighter colored checkmarks in Table~\ref{table:rel-operators}
indicate that path/walk-based operators {\em could} sensibly be used in conjunction with relational
link-value or node-values inputs, but this has been rarely if ever done. 
Below we discuss each of the operators from Table~\ref{table:rel-operators} in more detail.

\input{table_opers}

\medskip
\noindent
{\bf Relational Aggregates:}
Aggregation refers to a function that returns a single value from a collection of input values such
as a set, bag, or list. The most classical statistical aggregation operators are \textsc{Average},
\textsc{Mode}, \textsc{Exists}, \textsc{Count}, \textsc{Max}, \textsc{Min}, and 
\textsc{Sum}~\shortcite{neville:srl00,getoorlinkbased2003}.
  For SRL, another
frequent operator is \textsc{Proportion}, which computes, for instance, the fraction of a node's
neighbors that meet some criteria such as having the label ``C''~\shortcite{mcdowell2007cautious}.  These operators may also be
combined with thresholds, e.g., to evaluate whether the \textsc{Count} of a node's neighbors labeled
``C'' is at least 3.  The thresholding turns the numerical aggregate into a Boolean feature, which
is needed for tree-based algorithms~\shortcite{neville:kdd03}.  
\shortciteA{perlich:03} describe a set of more complex relational aggregates that depend on the distribution of attribute values that are
associated with each node (e.g., via links or a relational join).   For instance, these aggregates may use a function such as the edit distance to compare each node's distribution to a reference distribution
computed from the training data.  \shortciteauthor{perlich:03} demonstrate that these aggregations can in some cases improve performance compared to simpler
alternatives.
There are also aggregate operators that use only topology-based information.  For instance, the operator \textsc{Degree}, which simply
counts the number of adjacent links, can be a predictive feature, but should be applied carefully to
relational data to avoid bias~\shortcite{jensen:icml03}.

\medskip
\noindent
{\bf Temporal Aggregates:}
Relational information might also contain temporal information in the form of timestamps or durations for 
the links, node, or features. 
In general, such data can be handled by defining special temporal-aggregation features computed over 
the raw data~\shortcite{mcgovern2008spatiotemporal} or by defining a graph that summarizes all of the temporal information
(usually by decreasing the importance of less recent information) \shortcite{sharan:icdm08,rossi:10}.
\shortciteA{rossi:10} discuss an example of the latter approach, where they explore the impact of using various temporal-relational information and various kernels
for summarization.  
Alternatively, Section~\ref{sec:node-weight} discusses how notions of temporal distance can be used to modify path/walk-based metrics
such as node betweenness and closeness.

\medskip
\noindent
{\bf Set Operators:}
The traditional domain-independent set operators such as set union,
intersection, and difference can be applied to construct
features~\shortcite{kohavi1997wrappers}.  For instance, if there are two
attributes that both represent the presence of some word in a page
(node), a new feature might represent the case where a page contains both of those words 
(i.e., feature intersection).  For relational data, more complex set-based features are possible.  
For instance, a feature for collective classification might represent the union of all the class labels
of the nodes adjacent to the target node.  \shortciteA{neville:icdm03} propose a more complex approach where the feature value
is a multiset that represents the complete distribution of adjacent nodes' labels (e.g., {\tt \{3C, 2M, 5L\}} to indicate the labels
of ten adjacent nodes).  Using this feature representation, they show that
the ``independent-value'' approach that assumes that the labels are independently drawn from the same distribution 
yields the most effective relational classification.  Recently, \shortciteA{aha:09} showed that, for CC, this ``multiset'' approach usually outperformed other types of features such as the proportion or count-based aggregates discussed above.

\medskip
\noindent
{\bf Clique Potentials:} Some probabilistic models such as Relational Markov Networks
(RMNs)~\shortcite{taskardpm:02} perform inference over related nodes without computing aggregates.
Instead, they use clique-specific potential functions to represent the probabilistic
dependencies, and a product term in the probability computation naturally expands to accommodate a
varying number of neighbors for each node. In one sense, this is a ``featureless'' approach, since
there is no need to choose a relational aggregation function.  However, different kinds of
dependencies can still be represented by different cliques.  For instance, \shortciteA{taskardpm:02} consider
different sets of cliques for webpage classification: one based only on hyperlinks, the other
including information based on where links appear within a page.  Likewise, later work added
additional types of cliques to enable link prediction~\shortcite{Taskar03linkprediction}.
Thus, even with these models there remain important feature choices to be made.

Other probabilistic models also use link-based information without computing explicit features, such as the
random walk-based classifier of \shortciteA{limMRW10} or the weighted-neighbor approach of \shortciteA{macskassy2007classification}.
Even in these cases, however, choices remain about what types of links to use.  For instance, in webpage graphs, ``co-citation'' links
may be more predictive of class labels than direct links~\shortcite{macskassy2007classification,aha:09}.

\begin{figure}[t]
\centering
\includegraphics[width=4in]{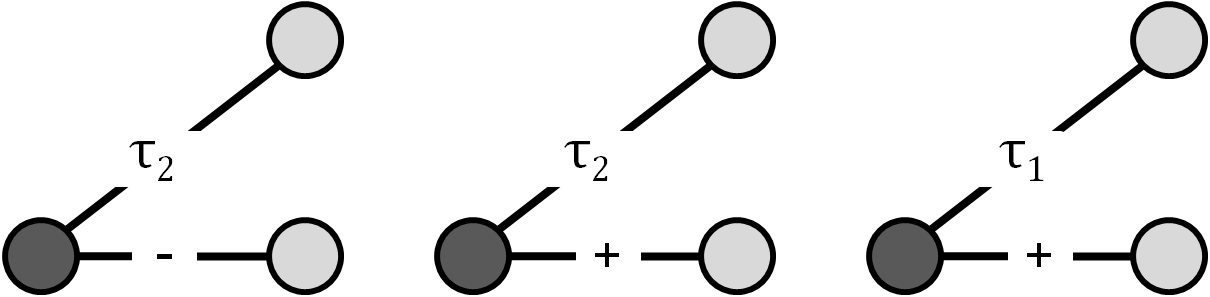} 
\caption{Subgraph Patterns with Link Labels. Each subgraph represents a possible pattern that a particular feature could look
for in relation to the target node (the bottom-left node in each case).}
\label{fig:link-type-patterns}
\end{figure}

\medskip
\noindent
{\bf Subgraph Patterns:}
A subgraph pattern feature is one that is based on the existence of a particular pattern
in the graph adjacent to the target node.
Such a feature might count how many times a particular pattern exists for the target node,
or produce a value of {\tt true} if at least one such pattern exists.
The simplest such pattern is called {\em reciprocity}; it is true when the target node $i$ links to node $j$ and
$j$ links back to $i$.  In most cases, however, the patterns are more complex and involve more nodes.  
\shortciteA{robins:sn07} define many such patterns including {\em two-star} (a node with at least two links),
{\em three-star} (a node with at least three links), and {\em triangle} (also known as transitivity, where
 $i \rightarrow j \rightarrow k$ and $i \rightarrow k$).  Most such patterns can be defined for both directed
and undirected links.

Many other patterns are possible. For instance, \shortciteA{Robins06ERGM} use subgraph patterns for probabilistically modeling graphs.
They argue that using more complex patterns such as the {\em alternating k-triangle} 
(based on finding $k$ triangles that all share a common side) can help to avoid degeneracy that might
otherwise arise during graph generation.  
Furthermore, subgraph patterns can also be extended to exploit labels on the links and/or nodes.
For instance, assume some links are labeled with $\tau_{1}$ or $\tau_{2}$ (representing different topics)
and some links are labeled with a plus or minus sign (representing positive or negative relationships).
Figure~\ref{fig:link-type-patterns} demonstrates three possible subgraph patterns, based on different link labelings,
relative to the target node shown at the bottom left of each subgraph.
A subgraph feature could compute, for each node, the number of matches for one of these patterns, and this feature
could be used for later analysis.

\medskip
\noindent
{\bf Dimensionality Reduction}
The goal of dimensionality reduction is to find a lower $k$-dimensional representation of the
initial $n$ features~\shortcite{sarwar2000application,fodor2002survey}.  More formally, given an initial
$n$-dimensional feature vector $\mathbf{x} = \{x_{1}, x_{2}, ..., x_{n}\}$, find a lower
$k$-dimensional representation $\mathbf{\tilde{x}}$ such that $\mathbf{\tilde{x}} = \{\tilde{x}_{1}, \tilde{x}_{2}, ...,
\tilde{x}_{k}\}$ with $k$ $\leq$ $n$ where the most significant information of the original data is
captured, according to some criterion. There are many dimensionality reduction methods such as
Principal Component Analysis (PCA), Principal Factor Analysis (PFA), and Independent Component
Analysis (ICA).

Dimensionality reduction techniques can be applied on the adjacency matrix $\mA$ of the graph $G$ to create a 
low-dimensionality graph representation; Section~\ref{sec:hybrid-lp} explained how this can be used
for link prediction.
These techniques can also be useful for feature computation.
For instance, \shortciteA{bilgicActiveLearning10} investigate active learning to improve the accuracy of collective
classification.  Their technique involves both non-relational and relational features, but they demonstrate that
first applying dimensionality reduction (with PCA) to the non-relational features simplifies learning, 
leading to substantial gains in accuracy.

\medskip
\noindent
{\bf Other operators:}  We mention only briefly those operators that have already been discussed extensively elsewhere.
{\bf Path-based measures} (such as betweenness and distance)
and {\bf walk-based measures} (such as PageRank) were discussed in Sections~\ref{sec:node-weight}.  These types of
measures have been used as features in a classifier to predict links~\shortcite{niteshchawlakdd2010} as
well as for validating relational sampling
techniques~\shortcite{leskovec:kronecker2010,moreno2009prop,ahmed2010time}.
These measures typically use only the topology (not the features), but one could easily imagine computing metrics based, for instance, only on paths where each edge had a particular label or type.
{\bf Textual analysis} techniques were discussed in Sections \ref{sec:link-label} and \ref{sec:node-label}, and {\bf relational clustering} techniques were discussed in Section~\ref{sec:node-pred}.
These operators were used specifically for node/link prediction, weighting, or labeling, but can also be used for more general feature construction.

Finally, there are operators based on {\bf similarity measures}.  Similarity between two nodes is often computed, for instance 
for link prediction (Section~\ref{sec:link-pred}) or weighting (Section~\ref{sec:link-weight}).  Such computations can easily lead
to a feature value for a {\em link}, since the link obviously refers to two endpoint nodes that can be compared.  However, for computing a {\em node} 
feature value, there
is usually no obvious other node for comparison, so similarity measures are not typically used for node feature values.
Such measures can, however, be used for node prediction, and Section~\ref{sec:node-pred} discusses how in some cases newly discovered nodes/groups can 
be used to create new node features.
As a particular instance of relational similarity functions, graph kernels for structured data~\shortcite{gartner2003survey} can also be used.
Such kernels can be used either between the nodes of a single graph \shortcite{kondorKernels2002} or to compute the similarity between two graphs \shortcite{vishwanathan2010graph}.
For instance, the former type of kernel is another technique that could also be used for link or group prediction.

\medskip
\noindent
{\bf Discussion:} Many of the feature operators discussed can naturally be used to compute feature values for links in additions to nodes.
For instance, textual analysis can be applied to links if there is text associated with each link, and most node-centered path-based measures have analogous
formulations for links.  One difference is that nodes naturally may link to many other nodes, whereas we assume links with just two endpoints. 
Thus, relational aggregates such as \textsc{Count} do not initially seem as useful for computing link features.  However, Figure~\ref{fig:link-aggr} previously demonstrated
how link-aggregation can be accomplished by broadening the computation to include the multiple links or nodes that are logically connected to each
endpoint node of the target link.  Naturally, some feature inputs and operators are better suited for computing node features vs.\ for computing link features. 
The next section examines how to select the most appropriate features for a given task.

\subsubsection{Searching, Evaluating, and Selecting Relational Features}\label{sec:search-select-features}

Given the large number of possible features that {\em could} be used for some task (such as the example Facebook classification task), which 
features should {\em actually} be used to learn a model?
In some cases, such selection is done manually based on prior experience or trial and error.
In many situations, though, more automatic feature selection is desirable.
For non-relational data, this has been a widely studied topic in machine learning~\shortcite{guyon2003introduction,koller1996toward,yang1997comparative,dash1997feature,jain1997feature,pudil1994floating},   
but selecting \textit{relational} features has received considerably less attention.
Given the large number of possible features, efficient strategies for searching over and evaluating the possible features is needed.
In this section, we first summarize these two key problems of feature search and feature evaluation, then give examples of how these issues have been
resolved in actual SRL systems.

\medskip
\noindent
{\bf Search:}
The first step in searching over the relational features is to define the possible relational
feature space by specifying the possible raw feature inputs (e.g., node and link feature values) and
operators to consider.  The possible operators can include \textit{domain-independent operators}
(e.g., mode, count) and/or \textit{problem-specific operators} (e.g., count the number of
friends divided by the number of groups). Domain-independent operators are obviously more general
and easier to apply, while the problem-specific operators can reduce the number of possibilities
that must be considered but require more effort and expert knowledge. However, both approaches are
vulnerable to selection biases~\shortcite{jensen:icml03,jensen:icml02}.  The second step is to pick an
appropriate search strategy, usually either \textit{exhaustive}, \textit{random}, or
\textit{guided}.  An exhaustive strategy will consider all features that are possible given the
specified inputs and operators, while a random strategy will consider only a fraction of this space.
A guided strategy will use some heuristic or sub-system to identify the features that should be
considered.  
In all three cases, each feature that is considered is subjected to some evaluation
strategy that assesses it usefulness; these strategies are described next.

\medskip
\noindent
{\bf Evaluation and Selection:}
Each feature that is considered must be evaluated in some way to determine if it will be retained for use in the final model.
For instance, a candidate feature may be evaluated by adding it to the current classification model; if it improves accuracy on a holdout set, then it is
immediately (and greedily) added to the set of retained features~\shortcite{davis2005integrated,davis:ijcai07}.
In other cases, every candidate feature is assigned some score and then only the best scoring feature is retained~\shortcite{neville:kdd03},
or features are added to the model based on decreasing score, so long as the new features continue to improve the model~\shortcite{mihalkova2007bottom}.
Simpler techniques that do not require evaluating the overall model can also be used.
For instances, metrics such as correlation or mutual information can be used to estimate how useful the feature is for the desired task.
Other metrics or strategies that could be used include Akaike's information criterion (AIC) \shortcite{akaike1974new}, Mallows $C_{p}$~\shortcite{mallows2000some}, Bayesian information criterion (BIC)~\shortcite{hannan1979determination,schwarz1978estimating} and many others~\shortcite{shao1996bootstrap,george1993variable}.
Frequently, a possible feature may have a particular parameter whose value must be set (such as a threshold); selecting the best value for a given feature
can use the same evaluation metrics or may use a simpler estimation technique, e.g., based on maximum likelihood.

\input{table-rel-feat-systems}

\medskip
\noindent
{\bf Examples:} Table~\ref{table:rel-feat-systems} summarizes the strategies used by a number of SRL
systems that automatically search for features.  The columns of the table describe how each system
searches for features and how the features are evaluated. For instance, Relational Probability Trees (RPTs)~\shortcite{neville:kdd03}
are an extension of
probability estimation trees for relational data that use an {\em exhaustive} search strategy for feature
selection.  In particular, RPT learning involves automatically searching over the space of possible
features using aggregation functions such as \textsc{Mode}, \textsc{Average}, \textsc{Count},
\textsc{Proportion}, \textsc{Min}, \textsc{Max}, \textsc{Exists}, and \textsc{Degree}.  These
aggregations can involve node and link feature values (\eg, for \textsc{Average}) or just topology
information (\eg, for \textsc{Degree}).  These features are used for classification tasks, such as
predicting the class label for a document.  Each feature is evaluated based on using the chi-square
statistic to measure the correlation between the feature and the class label; this yields a feature
score and an associated p-value.  Features with p-values below the level of statistical significance
are discarded, then the remaining feature with the highest score is chosen for inclusion in the
model.  This selection process has also been extended to use randomization tests to adjust for
biases that are common in relational data~\shortcite{jensen:icml03,jensen:icml02}.  RPTs have also been
extended for temporal domains~\shortcite{sharan:icdm08,neville:tenc2}.
 
RPTs represent the conditional probability distributions using a single tree.
In contrast, \shortciteA{natarajan2012gradient} propose using gradient boosting~\shortcite{friedman2001greedy} such that each conditional probability distribution is 
represented as a weighted sum of regression trees grown in a stage-wise optimization.
The features for each tree are selected via a depth-limited, exhaustive search, though they note that domain knowledge could also be used to guide this search.
\shortciteauthor{natarajan2012gradient} argue that the resultant set of multiple, relatively shallow trees allows efficient learning of complex structures, and demonstrate that
this technique can outperform alternatives based on single trees or the Markov Logic Networks discussed below.

Another system that uses exhaustive search is ReFeX~\shortcite{refex2011kdd}, which uses aggregates of \textsc{Sum} and \textsc{Mean} operators
to recursively generate features based on the degree of a node and its local neighborhood.  
To prune the resultant large set, ReFeX uses logarithmic binning of the feature values, clusters features based on their similarity in the binned space, and then retains only one feature from each cluster.
The logarithmic binning is chosen because it favors features that are more discriminative for high-degree nodes.
This recursive approach has also been modified for constructing features over dynamic networks~\shortcite{rossi2012dynamics}.

Alternatively, spatiotemporal RPTs \shortcite{mcgovern2008spatiotemporal} use a {\em random} search strategy.
In particular, these RPTs add temporal and spatial-based features to the set of possible features.
The resultant feature space is too large for exhaustive search, so instead random sampling is used.
After a pre-defined number of features have been considered, the best scored feature is added to the model.

The remaining systems that we will discuss all use a {\em guided} search strategy, where some
heuristic or sub-system provides candidate features that are considered.  
For instance, several such
systems~\shortcite{davis2005integrated,landwehr2005nfoil} use an ILP system to generate candidate
features, then evaluate those features and select some for ultimate use.  In particular,
SAYU~\shortcite{davis2005integrated} uses the ILP system Aleph~\shortcite{srinivasan1999aleph} to generate a
candidate feature (which they consider to be a new ``view'' on the original data).  Aleph creates
candidates features based on positive examples, from the training data, of the concept which is being predicted.
Each proposed feature is evaluated by learning a new model that includes the feature and then computing 
the area under the precision-recall curve (AUC-PR).  If a feature improves the AUC-PR score, it is
permanently added to the model and the feature search continues.  SAYU-VISTA~\shortcite{davis:ijcai07}
retains this same general approach but extends the types of features that can be considered, in
particular adding the ability to dynamically link together objects of different types and to
recursively build new features from other constructed features.  \shortciteauthor{davis:ijcai07}
demonstrate that the link connections are especially helpful in improving performance compared to
the original SAYU system.  
\shortciteA{landwehr2005nfoil} describe the nFOIL system which is very similar to SAYU but was developed independently, 
while \shortciteA{de2010probabilistic} describe how ProbFOIL upgrades a deterministic rule learner like FOIL to be probabilistic.
\shortciteA{landwehr2010fast} describes the related kFOIL system which integrates FOIL with kernel methods. They also consider the impact of 
several different feature scoring functions.

A number of systems have considered how to perform structure learning for Probabilistic Relational Models (PRMs)~\shortcite{getoor:icml01}
or for Markov Logic Networks (MLNs)~\shortcite{Domingos04markovlogic}, which is a more general case of the feature selection
problems described above.  For instance, a MLN is a weighted set of first-order formulas; structure
learning corresponds to learning these formulas while weight learning corresponds to learning the
associated weights.  The first MLN structure learning approaches systematically construct candidate
clauses by starting from an empty clause, greedily adding literals to it, and testing the resulting
clauses fit to the training data using a statistical measure~\shortcite{kok:05,biba2008discriminative}.
However, these ``top-down'' approaches are inefficient because the initial proposal of clauses
ignores the training data, resulting in a large number of possible features being considered and
possible problems with local minima.  In response, a number of ``bottom-up'' approaches have been
proposed.  In particular, \shortciteA{mihalkova2007bottom} uses a propositional Markov network
structure learner to construct template networks to guide the construction of features based on the training data.  More recent
work has examined how to enable bottom-up approaches to learn longer clauses based on constraining
the search to only consider features consistent with certain patterns or {\em
motifs}~\shortcite{kokMotifs2010}, or by clustering the input nodes to create a ``lifted graph'' representation,
enabling feature search over a smaller graph~\shortcite{Kok09structlearning}.  

\shortciteA{khosravi2010structure} perform MLN structure learning by first learning the structure of a simpler
Parametrized Bayes Net (PBN)~\shortcite{poole2003first}, then converting the result into a MLN.
For data that contains a significant number of descriptive attributes, they show that this approach
dramatically improves the runtime of structure learning and also improves predictive accuracy.
\shortciteA{schulte2011tractable} has given a theoretical justification for this approach.
Another alternative, proposed by ~\shortciteA{khot2011learning}, is to extend the previously mentioned work of \shortciteA{natarajan2012gradient} on gradient boosting 
to MLNs.  
Essentially, the problem of learning MLNs is transformed into a series of relational regression problems where the functional gradients are represented as clauses
or trees.  For several datasets they demonstrate faster MLN structure learning that is as accurate or better than baselines including
the algorithms of \shortciteA{mihalkova2007bottom} and \shortciteA{kokMotifs2010}.

The above techniques for MLNs all seek to learn a network structure that best explains the training data as a whole.
In contrast, for situations where the prediction of a specific predicate is desired (\eg, to predict the political 
affiliation in our Facebook example), \shortciteA{huynhDiscimStruct2008} and \shortciteA{biba2008discriminative}
both propose discriminative approaches to MLN structure learning.  For instance, Huynh and Mooney use a modified version of 
Aleph~\shortcite{srinivasan1999aleph} to compute a large number of candidate clauses, then use a form of $L_1$-regularization
to force the weights that are subsequently learned for these clauses to be zero when the clause is not very helpful for
predicting the predicate.  This regularization, in conjunction with an appropriate optimization function, effectively
leads to selecting a smaller set of features that are useful for the desired task.

\medskip
\noindent
{\bf Discussion:}
We focus in this article on graph-based data representations (see Section~\ref{sec:scope}).  However,
many of the examples discussed above use a logical representation instead.  We include
them in this section because the techniques used for constructing and searching for features or rules
are very similar in both settings.
For instance, both RPTs (a graph-based approach) and RDN-Boosting (a logical approach)
use an exhaustive search over probabilistic decision trees, with different feature scoring strategies.

\shortciteA{Popescul03SR} examine how to automatically learn new relational features for links (to
support link prediction), but their techniques could also be applied to constructing node features.
In particular, they treat each feature as a relational database query, and use the concept of
refinement graphs~\shortcite{shapiro1982algorithmic} to consider refining an initial query with
equi-joins, equality selections, and statistical aggregates.
After each refinement, further refinements can be considered; this search is guided by sampling
over some possible further refinements and proceeding only if the results of a particular refinement
or type seems promising.  The features chosen are combined with a logistic regression classifier.
For evaluation of the specific features, they use the Bayesian Information Criterion (BIC)~\shortcite{schwarz1978estimating},
which includes a term than penalizes feature complexity to reduce the danger of overfitting.

We discussed multiple systems that include notions of aggregation including RPTs, SAYU-VISTA, and
the work of \shortciteA{Popescul03SR} discussed above.  There are also other aggregate-based
learning approaches such as Crossmine~\shortcite{yin2006crossmine}, CLAMF~\shortcite{frank2007method},
Multi-relational Decision Trees (MRDTL) \shortcite{leiva2002mrdtl}, Confidence-based Concept Discovery
(C$^{2}$D) ~\shortcite{kavurucu-aggregation}, and many
others~\shortcite{perlich:06,krogel2001transformation,knobbe2002involving}.  There are also other
possibilities for feature evaluation.  For instance, GleanerSRL~\shortcite{goadrich2007combining} uses
Aleph~\shortcite{srinivasan1999aleph} to search for clauses and then uses a metric of $precision$
$\times$ $recall$ for evaluating the clauses.

%% file: table_opers.tex
\newcommand\JT{\rule{0pt}{2.8ex}}
\newcommand\JB{\rule[-1.8ex]{0pt}{0pt}}

\begin{table}
\caption{\textsc{Relational Feature Operators}: Summary of the most popular types of relational feature operators. 
A check is used to indicate the classes of inputs (see Section~\ref{sec:rel-feature-space}) that each operator most naturally uses for constructing feature values, while a lighter check 
indicates that the operator {\em could} sensibly be used with that input but that this combination has rarely if ever been used.} 
\label{table:rel-operators}
\vspace{-0.3in}
\begin{center}
\begin{small}
\begin{tabular}{l||p{78mm}||c|c|c|c||}

\multicolumn{1}{c||}{} \JT \JB
& 
& \multicolumn{4}{c||}{\textbf{Inputs}}
\\ 
\multicolumn{1}{l||}{\textbf{Relational Operators}} \JT \JB
& \textbf{Example Techniques}
& \rotatebox{90}{\textbf{Non-relational}}
& \rotatebox{90}{\textbf{Topology}}
& \rotatebox{90}{\textbf{Relational Link-value}}
& \rotatebox{90}{\textbf{Relational Node-value}}
\\
\toprule
\bottomrule

\textbf{Relational aggregates} & 
\textsc{Mode}, \textsc{Average}, \textsc{Count}, \textsc{Proportion}, \textsc{Degree}, ...
& \JT \JB           & \mycheck & \mycheck & \mycheck
\\ \hline

\textbf{Temporal aggregates} &  
Exponential/linear decay, union,  ...
& \JT \JB \mycheck  & \mycheck & \mycheck & \mycheck
\\ \hline

\textbf{Set operators} & 
Union, intersection, multiset, ...
& \JT \JB \mycheck  &          & \mycheck & \mycheck
\\ \hline

\textbf{Clique potentials} & 
Direct link cliques, co-citation cliques, triads, ...
& \JT \JB           & \mylight & \mycheck & \mycheck
\\ \hline

\textbf{Subgraph patterns} & 
Two star, three-star, triangle (i.e., transitivity), ...
& \JT \JB           & \mycheck & \mylight & \mylight
\\ \hline

\textbf{Dimensionality reduction} & 
PCA, SVD, Factor Analysis, Principal Factor Analysis, Independent Component Analysis, ...
& \JT \JB \mycheck  & \mycheck & \mycheck & 
\\ \hline

\textbf{Path/walk-based measures} &  
Betweenness, common neighbors, Jaccard's coefficient, Adamic/Adar, shortest paths, random-walks, ...
& \JT \JB           & \mycheck & \mylight & \mylight
\\ \hline

\textbf{Textual analysis} & 
LSA, LDA, PLSA, Link-LDA, Link-PLSA, ... 
& \JT \JB \mycheck  &          & \mycheck & \mycheck
\\ \hline

\textbf{Relational clustering} &
Spectral partitioning, Hierarchical clustering, Partitioning relocation methods (k-means, k-medoids), ... 
& \JT \JB \mycheck  & \mycheck & \mylight & \mylight
\\ \hline

\toprule
\bottomrule
\end{tabular}
\end{small}
\end{center}
\end{table}

%% file: table-rel-feat-systems.tex
\newcommand\TE{\rule{0pt}{3.0ex}}
\newcommand\BE{\rule[-1.1ex]{0pt}{0pt}}
\begin{table} 
\caption{\textsc{Systems for Searching for and Selecting Node Features}: A summary of some of the systems that can be used to automatically
search for and select the most appropriate features for a given task.
Note that, depending on the context, these papers may be describe
their function in terms of learning the best {\em rules} for a system
or of learning the {\em structure} (\eg, of a MLN). Only some of the MLN-based systems are described; 
for some of these, WPLL is the ``weighted pseudo log-likelihood.''}
\label{table:rel-feat-systems}
\vspace{-4mm}
\begin{center}
\begin{small}
\begin{tabular}{p{70mm}cc}
\toprule
\TT \BB 
\large \textbf{Proposed System}
& \large \textbf{Search method}
& \multicolumn{1}{c}{\large\textbf{Feature Evaluation}}
\\
\midrule

\TE \BE \textbf{\normalsize \textbf{RPT}}~\cite{neville:kdd03}
& Exhaustive 
& Chi-square statistic/p-value
\\

\TE \BE \textbf{\normalsize \textbf{RDN-Boosting}}~\shortcite{natarajan2012gradient}
& Exhaustive
& Weighted variance
\\

\TE \BE \textbf{\normalsize \textbf{ReFeX}}~\shortcite{refex2011kdd}
& Exhaustive
& Log-binning disagreement
\\

\TE \BE \textbf{\normalsize \textbf{Spatiotemporal RPT}}~\cite{mcgovern2008spatiotemporal} 
& Random 
& Chi-square statistic/p-value
\\

\TE \BE \textbf{\normalsize \textbf{SAYU}}~\cite{davis2005integrated}  
& Aleph 
& AUC-PR
\\

\TE \BE \textbf{\normalsize \textbf{nFOIL}}~\cite{landwehr2005nfoil}
& FOIL 
& Conditional Log-Likelihood
\\

\TE \BE \textbf{\normalsize \textbf{SAYU-VISTA}}~\cite{davis:ijcai07}
& Aleph 
& AUC-PR
\\

\TE \BE \textbf{\normalsize \textbf{ProbFOIL}}~\shortcite{de2010probabilistic}
& FOIL
& {\em m}-estimate
\\

\TE \BE \textbf{\normalsize \textbf{kFOIL}}~\shortcite{landwehr2010fast}
& FOIL
& Kernel target alignment
\\

\TE \BE \textbf{\normalsize \textbf{PRM struct. learning}}~\shortcite{getoor:icml01}
& Greedy hill-climbing 
& Bayesian model selection
\\

\TE \BE \textbf{\normalsize \textbf{TSDL}}~\cite{kok:05}  
& Beam search
& WPLL
\\

\TE \BE \textbf{\normalsize \textbf{BUSL}}~\cite{mihalkova2007bottom} 
& Template-based
& WPLL
\\

\TE \BE \textbf{\normalsize \textbf{PBN Learn-And-Join}}~\shortcite{khosravi2010structure}
& Level-wise search
& Pseudo-likelihood
\\

\TE \BE \textbf{\normalsize \textbf{Discriminative MLN structure learning}}~\shortcite{huynhDiscimStruct2008,biba2008discriminative} 
& Aleph++
& {\em m}-estimate
\\

\bottomrule
\end{tabular}
\end{small}
\end{center}
\end{table}

%% file: joint-disc.tex
\section{Jointly Transforming Nodes and Links}\label{sec:joint-disc}

In the previous sections, we primarily discussed relational
representation transformation techniques that are applied independently of
one another.  For instance, one technique might be used to predict
links, while another builds on the transformed representation by
applying a node labeling technique.  This section instead
examines ``joint'' transformation tasks that combine node and link
transformation in some way, for instance to label the nodes and weight the
links simultaneously.  Such techniques may enable each subtask to
influence the other in helpful ways, and avoids any bias that might be
introduced by requiring the serialization of two tasks
(such as link weighting and node labeling) that might usefully be performed jointly.

One recent approach proposed by~\shortciteA{namata2011collective} collectively performs link prediction, node labeling,
and entity resolution (which can be seen as a form of node deletion/merging).
They present an iterative algorithm that solves all three tasks simultaneously by propagating information among solutions to
the above three tasks. In particular, they introduce the notion of {\em inter-relational} features, which are relational features
for one task that depend upon the predicted values for another.  Their results show that using such features can improve accuracy,
and that inferring predicted values for all three tasks simultaneously can significantly improve accuracy compared to performing the three
tasks in sequence, even if all possible orderings are considered.

\input{tab-joint-disc}

Techniques that model the full distribution across links and attributes such as RMNs~\shortcite{taskardpm:02}, PRMs~\shortcite{Friedman99:PRMs},
and MLNs~\shortcite{Domingos04markovlogic} can also be used in this scenario, for instance to jointly predict node and link labels.
In this section, however, we focus particularly on recent techniques that all presume the existence of 
some textual content
that is associated with the nodes or links of the graph (although the
basic algorithms would also work with other kinds of features).  We consider three types of
techniques, based on what kind of input text they use: stand-alone
text documents (e.g., legal memos with no links), 
text documents connected by links (e.g., webpages with hyperlinks),
or entities connected by links that have associated text (e.g., people connected by email messages).
Table~\ref{table:ldtypes} lists some of the most prominent models, grouped according to these three types.
The columns of this table indicate what kinds of input the models use (middle section) and the
types of transformation they can perform (right-hand section).  
The text documents corresponds to node features in this table, while text associated with links yields link features.
Below we discuss each of the three types of techniques in more detail.

\medskip
\noindent
{\bf Using text documents with no links:}
First, many techniques can be used to assign topics or labels to the
nodes when those nodes (such as documents) have associated text.  For
instance, the first row of Table~\ref{table:ldtypes} indicates that
LDA and PLSA use only the nodes and node features and can perform node
prediction, weighting, and labeling.  Section~\ref{sec:node-interp}
already mentioned how these techniques can be used to label each node
with one or more discovered topics, which is their more typical use.
However, these techniques can also perform node weighting (using the
weights associated with the topics) and/or node prediction (by
converting the discovered topics to new latent nodes as discussed in
the introduction to Section~\ref{sec:node-pred}).
In Table~\ref{table:ldtypes}, we use lighter checkmarks to represent 
these kind of situations where a transformation task {\em could} be performed by
a particular model but is not its primary use/output.

LDA and PLSA treat each document as a bag of words and seek to assign
one or more topics (labels) to each document based on the words.  In
contrast, Nubbi \shortcite{chang2009connections} designs an approach based
on LDA where a graph is defined based on objects (nodes) that are
referenced in a set of documents, then links are predicted based on the 
relationships that are implied in the text of the documents.
In addition, the nodes and links are associated with their most likely topic(s) 
based on these relationships.
Thus, this model simultaneously performs
link prediction, link labeling, and node labeling.
A similar result is produced by the semantic network extraction of ~\shortciteA{KokD08} that
was discussed in Section~\ref{sec:link-label}.

\medskip
\noindent
{\bf Using text documents with links:} The second type of joint
transformation also uses text documents, but adds known links between the
documents to the model.  For instance, Section~\ref{sec:node-interp}
discussed how Link-LDA and Link-PLSA add link modeling to LDA and PLSA
in order to perform node labeling; as discussed above for LDA and PLSA
this can be modified to also achieve node prediction and weighting.
As shown in Table~\ref{table:ldtypes}, Link-LDA and Link-PLSA can also
be used for link prediction and weighting by learning a model from a
training graph and then using it to predict unseen links on a new test
graph~\shortcite{nallapati2008joint}.

Link-LDA and Link-PLSA model links in a way that is very similar to how they model the presence of words in a document (node).  For instance,
in Link LDA's generative model, to generate one word, each document chooses a topic, then chooses a word from a topic-specific multinomial.  
The identical process (using a topic-specific multinomial) is used to generate, for a particular document, one target document to link to.
Thus, Link-LDA and Link-PLSA directly extend the original LDA and PLSA models to add links.  

\shortciteA{nallapati2008joint} argue that Link-LDA's and Link-PLSA's extensions for links, while pragmatic,
do not adequately capture the topical relationship between two documents that are linked together.
Instead, they propose two alternatives.  The first, Pairwise Link-LDA, replaces the link model of Link-LDA
with a model based on mixed membership stochastic blockmodels~\shortcite{blei:jmlr08}, where each
possible link is modeled as a Bernoulli variable that is conditioned on a topic chosen based on the
topic distributions of each of the two endpoints of the link.  The second approach, Link-PLSA-LDA,
retains the link generation model of Link-LDA, but changes the word generation model for some of the
documents (the ones with incoming links) 
so that the words in such a document depend on the topics of other documents that link to it.
The downside of this latter
approach is that it only works when the nodes can be divided into a set with only outgoing links and
a set with only incoming links.  However, \shortciteauthor{nallapati2008joint} argue that this limitation can be largely
overcome by duplicating any nodes that have both incoming and outgoing links.  Moreover, this
approach is much faster and more scalable than Pairwise Link-LDA.  \shortciteauthor{nallapati2008joint} demonstrate that
both models outperform Link-LDA on a likelihood ranking task, and that Link-PLSA-LDA also
outperforms Link-LDA on a link prediction task.  They also show that Link-PLSA-LDA and Link-LDA were
comparable in terms of execution time, but that Pairwise Link-LDA was much slower.

Changes to the generative model used by each of these approaches encode different assumptions about the data
and can lead to significant performance differences.  For instance, \shortciteA{chang2009relational} introduce
the Relational Topic Model (RTM) and compare it to the Pairwise Link-LDA model discussed above.  Both models allow similar
flexibility in terms of how links are defined, but \shortciteauthor{chang2009relational} argue that their model 
forces the same topic assignments that are used to generate the words in the documents to also generate the links, which is not
true of Pairwise Link-LDA.  They then demonstrate that RTM provides more accurate predictions and link suggestions than
Pairwise Link-LDA and several other baselines.

Another possible change to the model is to add other types of objects.
For instance, Topic-Link LDA \shortcite{liu2009topic} models not only
documents, links, and the most likely topics associated with each
document, but also explicitly considers the author of each document
and clusters these authors into multiple ``communities.''  Creating
this new clustering is not equivalent to finding per-document topics
because each author is associated with more than one document.  They
argue that this approach is analogous to unifying the separate
tasks of (1) assigning topics to documents and (2) analyzing the
social network of authors.  They show that their approach can in some
cases outperform LDA and Link-LDA.

\medskip
\noindent
{\bf Using text associated with links:} 
The final type of joint transformation techniques form link features based on text associated with links, such as the text of email
messages \shortcite{mccallumrole:07} or scientific abstracts that relate to a particular protein-protein interaction~\shortcite{balasubramanyan-block}.
Several such techniques were discussed previously in the context of link interpretation.
For instance, Section~\ref{sec:link-label} discussed how models such as the Author-Recipient-Topic (ART) model~\shortcite{mccallumrole:07}
and the Group-Topic (GT) model~\shortcite{mccallumjoint:07} extend LDA to perform link labeling; the strength of these predicted labels (topics) can also be used to weight the links.
In addition, the GT model directly assigns nodes to groups (i.e., node labeling), while the labels that ART associates with each link could also be used to label the associated nodes.
The RART model~\shortcite{mccallumrole:07} extends ART by allowing a node to have multiple roles.
More recently, Block-LDA \shortcite{balasubramanyan-block} merges the ideas from these latent variables models with stochastic blockmodels. More specifically, 
the Block-LDA shares information through three components: the link model shares information with a block structure which is then shared by the topic model. 
Unlike GT and ART, however, Block-LDA focuses on labeling the nodes rather than the links.
\shortciteauthor{balasubramanyan-block} evaluate Block-LDA on a protein dataset and the Enron email corpus and demonstrate that it outperforms Link-LDA and several other baselines on the
task of protein functional category prediction.

\medskip
\noindent
{\bf Discussion:} 
Most of the techniques discussed above are variants of latent group models that focus on node and/or link label prediction,
 but they can also be used for node prediction where the new nodes represent newly discovered topics or latent groups.
These models have also been extended to incorporate notions of time~\shortcite{Dietz07unsupervisedprediction,wang2008continuous,wang2006topics}, topic hierarchies~\shortcite{li2006pachinko}, 
and correlations between topics~\shortcite{blei2007correlated}.  
In addition, links are usually assumed to be generated based on the overall topic(s) of a node or link.  In contrast,
the Latent Topic Hypertext Model (LTHM) \shortcite{gruber2008latent} models each link as originating from some specific word in a document.
Somewhat surprisingly, they show that this approach leads to a model with fewer parameters than models like Link-LDA, and demonstrate that their approach
outperforms both Link-LDA and Link-PLSA when evaluated on a link prediction task.

\input{fig-joint-disc}

If new nodes are added to the graph to represent discovered topics,
then links are invariably added to connect existing nodes to the new nodes.  However, some
models may also learn information about how the discovered
topics are related to each other.  For instance, Figure~\ref{fig:joint-disc} shows how
two new topics are discovered in a graph and how they are connected to the
existing nodes.  In addition, the topics are connected to each other with
new links where the weight of each link represents how frequently a
document from that topic cites a document representing a different
topic.  Adding these additional links to the graph lets the original nodes be
connected more closely not only to their primary topics but also to related topics.

%% file: tab-joint-disc.tex
\begin{table}
\caption{\textsc{Summary of the Joint Transformation Models}:  The middle section of the table indicates what types of graph features are used as inputs to the model, while the right side of the table indicates what types of link or node transformation can be performed by the model.
Lighter checkmarks indicate that the output of the model can be transformed to perform a particular transformation task (e.g., to use the node labels
to create new latent group nodes), but where that task was not the primary goal of the specified model.}
\label{table:ldtypes}
\begin{center}
\begin{small}
\begin{tabular}{l||c|c|c|c||c|c|c|c|c|c}

\multicolumn{5}{c||}{} \JT \JB
& \multicolumn{3}{c|}{\textsc{Links}}
& \multicolumn{3}{c}{\textsc{Nodes}}
\\ 
\multicolumn{1}{l}{} \JT \JB
& \multicolumn{4}{c||}{\textsc{\Large Input}}
& \rotatebox{90}{\textbf{Link Prediction}}
& \rotatebox{90}{\textbf{Link Weighting}}
& \rotatebox{90}{\textbf{Link Labeling}}
& \rotatebox{90}{\textbf{Node Prediction}}
& \rotatebox{90}{\textbf{Node Weighting}}
& \rotatebox{90}{\textbf{Node Labeling}}
\\
\toprule
\bottomrule
\textbf{Joint Transformation Model} \JT \JB & $E$ & $\mathbf{X^{E}}$ & $V$ & $\mathbf{X^{V}}$

& $\tilde{E}$ & $\mathbf{\tilde{X}^{E}}$ & $\mathbf{\tilde{X}^{E}}$ & $\tilde{V}$ & $\mathbf{\tilde{X}^{V}}$ & $\mathbf{\tilde{X}^{V}}$ \\
\toprule
\bottomrule

LDA/PLSA & \JT \JB 
 &  & \checkmark & \checkmark &
 & & & \mylight & \mylight & \checkmark \\ \hline

Nubbi & \JT \JB 
 &  & \checkmark & \checkmark &
\checkmark &            & \checkmark & \mylight  & \mylight  & \checkmark \\ \hline \hline

Link-LDA, Link-PLSA & \JT \JB 
\checkmark & & \checkmark & \checkmark &
\checkmark & \checkmark & & \mylight & \mylight & \checkmark \\ \hline

Pairwise-Link-LDA & \JT \JB 
\checkmark &  & \checkmark & \checkmark &
\checkmark & \checkmark & & \mylight & \mylight & \checkmark \\ \hline

Link-PLSA-LDA & \JT \JB 
\checkmark &  & \checkmark & \checkmark &
\checkmark & \checkmark & & \mylight & \mylight & \checkmark \\ \hline
 
Relational Topic Model (RTM) & \JT \JB 
\checkmark & & \checkmark & \checkmark &
\checkmark & \checkmark & & \mylight & \mylight & \checkmark \\ \hline
 
Topic-Link LDA & \JT \JB 
\checkmark & & \checkmark & \checkmark &
\checkmark & \checkmark & & \mylight & \mylight & \checkmark \\ \hline \hline
 
\hline

Group-Topic (GT) & \JT \JB 
\checkmark & \checkmark & \checkmark & &
& \checkmark & \checkmark  & \mylight & \mylight & \checkmark \\ \hline
 
Author-Recipient-Topic (ART) & \JT \JB 
\checkmark & \checkmark & \checkmark &  &
& \checkmark & \checkmark  &  & & \mylight \\ \hline

Block-LDA & \JT \JB 
\checkmark & \checkmark  & \checkmark &  &
\checkmark & \checkmark & & \mylight & \mylight & \checkmark \\ \hline

\toprule
\bottomrule
\end{tabular}
\end{small}
\end{center}
\end{table}

%% file: fig-joint-disc.tex
\begin{figure}[t]
\centering
\subfigure[Initial Graph]{
\includegraphics[width=1.6in]{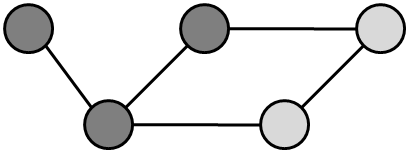}
\label{fig:joint-before}
}
\subfigure[Joint Transformation]{
\includegraphics[width=3.6in]{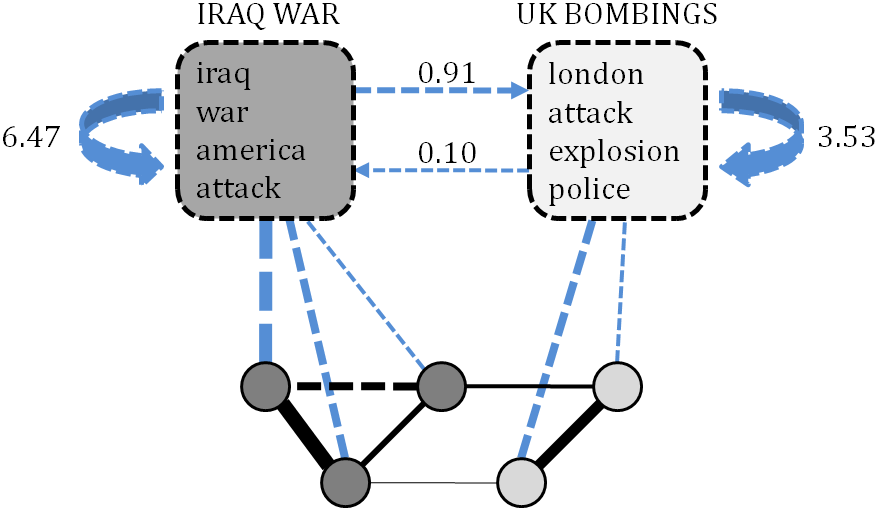}
\label{fig:joint-after}
}
\caption[Optional]{\textsc{Example of Joint Transformation:} In this example, adapted from results in \shortciteA{nallapati2008joint},
new latent nodes are added to represent discovered topics, and weighted links are added from each original node to a new latent node.
In addition, weighted links are added between the latent nodes, representing connection strength between these topics.  Finally, new links
between the original nodes may be also be predicted.}
\label{fig:joint-disc}
\end{figure}

%% file: discussion.tex
\section{Discussion and Challenges}\label{sec:discussion}

In this section we discuss additional issues that are related to relational representation transformation
and highlight important challenges for future work.

\subsection{Guiding and Evaluating Representation Transformation}

The goal of representation transformation is often to ``improve'' the data representation in some way 
that leads to better results for a subsequent task or possibly to a more understandable representation.
How can we evaluate whether a particular transformation technique has accomplished this goal?
We first address this question, then consider when the final goal can be used to more directly guide the initial transformation.

For some tasks, representation evaluation is straightforward provided that ground truth values are
known for a hold-out data set.  For instance, to test if a technique for
link prediction is effective, accuracy can be measured for links predicted for the hold-out set
\shortcite{Taskar03linkprediction,liu2009topic}.  The particular evaluation metric can be modified as
appropriate for the domain.  For instance, \shortciteA{chang2009relational} evaluate the precision
of the twenty highest-ranked links suggested for each document, while
\shortciteA{nallapati2008joint} consider a custom metric called ``RKL'' that measures the rank of
the last true link suggested by the model.  Likewise, if the desired task involves classification, then
a classification algorithm can be run on the hold-out data, with and without the representation
change, to see if the change increases classification accuracy.

In other cases, it may be difficult to directly measure how well a representation change has
performed, but classification can be used as a surrogate measure: if accuracy increases, the change
is assumed to be beneficial.  For instance, classification has been used to evaluate link
prediction~\shortcite{gallagher:08}, link weighting~\shortcite{xiang:10}, link labeling~\shortcite{rossi:10,SofusMinedLinks}, and node prediction~\shortcite{neville:icdm05}.  
In addition, node labeling is naturally a classification problem, while node weighting is usually evaluated in
other ways, e.g., based on query relevance.

Other techniques can be used when direct evaluation is not feasible, but there exists some other 
metric that is believed to be related.
For instance, higher autocorrelation in a graph can be associated with the presence of more sensible links,
and algorithms such as collective classification typically perform better when the level of autocorrelation is higher.  
Thus, \shortciteA{xiang:10} demonstrate the success of their technique for estimating relationship
strengths (link weights) based in part on showing an increase in autocorrelation when measured for several attributes in a social network.
Likewise, increased information gain for some of the attributes could be used to demonstrate an improved representation~\shortcite{lippi2009relational},
or link perplexity could be used to assess topic labelings~\shortcite{balasubramanyan-block}.
Naturally, the most appropriate evaluation techniques vary based upon the task, 
and a comparison of transformation techniques may yield different results depending upon what metric is chosen.

Ideally, representation transformation would be guided more directly by the final goal as it is executed, rather than only being evaluated when the
transformation is complete.  
This is often the case for the feature selection and structure learning algorithms discussed in
Section~\ref{sec:node-feature}: task accuracy (or a surrogate measure) is evaluated with a particular feature
added, and it is retained if accuracy has improved.
In other cases, the transformation is even more directly specified by the desired end goal.
For instance, the ``supervised random walk'' approach discussed in Section~\ref{sec:hybrid-lp} uses a gradient descent method 
to obtain new link weights such that links predicted by a subsequent random walk (their final goal) will be more accurate. 
Likewise, \shortciteA{menon2010predicting} show how to add supervision to methods for generating latent features
(see introduction to Section~\ref{sec:node-pred}) so that the features learned would be more relevant to their final classification task.  
They show, however, that adding such supervision is not always helpful.
As a final example, \shortciteA{shicollective2011} use a quadratic program to optimize a linear combination
of link weights such that the final link weights will lead directly to more accurate classification via a label propagation algorithm.

In general, ensuring that a particular transformation will improve performance on the final SRL task remains challenging. 
Many transformations cannot be directly guided by the final goal, either because suitable supervised data is not available, or because
it is not clear how to modify the transformation algorithms to use such information (e.g., with the latent topic models of Section~\ref{sec:joint-disc}
or the group detection algorithms of Section~\ref{sec:node-pred}).

\subsection{Causal Discovery}
Causal discovery
refers to identifying cause-and-effect relationships (i.e., smoking causes cancer) from either online
experimentation~\shortcite{aral-creating} or from observational data.  The challenge is to distinguish
true causal relationships from mere statistical correlations.
One approach is to use quasi-experimental designs (QEDs), which
take advantage of circumstances in non-experimental data to identify situations that provide the
equivalent of experimental control and randomization.  \shortciteA{jensen2008automatic} propose a system to 
discover knowledge by applying QEDs that were discovered automatically.  More
recently, \shortciteA{oktay2010causal} applies three different QEDs to
demonstrate how one can gain causal understanding of a social media system.  There is also another
causal discovery technique for linear models proposed by \shortciteA{zhenxingcausal}.  The challenge
remains of how to extend these techniques to apply to a broader range of relational data.

\subsection{Subgraph Transformation and Graph Generation}

The majority of this article focused on transformation tasks centered around the nodes or links of the graphs.
However, there are also useful tasks for {\em subgraph transformation} which seek
to identify frequent/informative substructures in a set of graphs or to create features or classify
such subgraphs \shortcite{inokuchi2000apriori,deshpande2005frequent}.
For instance, \shortciteA{xiangnangraphclassif} consider how to use semi-supervised techniques to perform feature selection
for subgraph classification given only a few labeled subgraphs.  
As with nodes and links, for subgraphs the tasks of prediction, labeling, weighting, and feature generation can all be described.
Many of the techniques that we described for node-centered features can also be used in this context, but a full discussion
of subgraph transformation is beyond the scope of this article.

Recently, graph generation algorithms have attracted significant interest.  These algorithms use
some model to represent a family of graphs, and present a way to generate multiple samples from this
family.  Two prominent models are Kronecker Product Graph Models (KPGMs)
\shortcite{leskovec:kronecker2010} and those based on {\em preferential attachment} \shortcite{price1976general,barabási1999emergence}.
These graph generation methods take advantage of global (with KPGMs) and local (with preferential attachment models) graph
properties to generate a distribution of graphs that can potentially include attributes.  Sampling
from these models can be useful for creating more robust algorithms, for instance by training a classifier on a
family of related graphs instead of on a single graph.  
\shortciteA{Newman03thestructure} surveys additional network models and properties
that are relevant to graph generation.

\subsection{Model Representation}
\label{sec:modelrep}

In SRL there is also the notion of model
representation: what kind of statistical model is learned to represent the relationship between the
nodes, links, and their features?  Some of the most prominent models for SRL are Probabilistic
Relational Models (PRMs) \shortcite{Friedman99:PRMs}, Relational Markov Networks (RMNs)
\shortcite{taskardpm:02}, Relational Dependency Networks (RDNs) \shortcite{neville:jmlr07}, Structural
Logistic Regression \shortcite{Popescul03SLR}, Conditional Random Fields (CRFs) \shortcite{lafferty:01}, and
Markov Logic Networks (MLNs) \shortcite{Domingos04markovlogic,richardson2006markov};  full discussion of
these models is beyond the scope of this article.  In many cases techniques for relational
representation transformation, such as link prediction, can be performed regardless of what kind of
statistical model will be subsequently used.  However, the choice of statistical model does strongly
interact with what kinds of node and link features are useful (or even possible to use); 
Section~\ref{sec:node-feature} describes some of these connections.
While a number of relevant comparisons have already been published~\shortcite{jensen:kdd04,neville:jmlr07,macskassy2007classification,sen2008collective,aha:09,craneMLN11},
more work is needed to evaluate the interaction between the choice of statistical model and feature selection, and to evaluate which
statistical models work best in domains with certain characteristics.

\subsection{Temporal and Spatial Representation Transformation}

Where appropriate, we have already discussed multiple techniques that can incorporate temporal
information from graph data (see especially Sections~\ref{sec:link-label}, \ref{sec:node-weight},
and \ref{sec:node-feature}).  These techniques focused on solving particular problems such as node
classification, but dealing with such data invariably requires studying how to represent the
time-varying elements.  However, more work is needed to examine the general tradeoffs involved with
different temporal representations.  For instance, \shortciteA{hill2006building} provide a generic
framework for modeling any temporal dynamic network where the central goal is to build an
approximate representation that satisfies pre-specified objectives.  They focus on summarization
(representing historical behavior between two nodes in a concise manner), simplification (removing
noise from both edges and nodes, spurious transactions, or stale relationships), efficiency
(supporting fast analysis and updating), and predictive performance (optimizing the representation
to maximize predictive performance).  This work provides a number of useful building blocks, but
more comparisons are needed to, for instance, evaluate the merits of using summarized networks with
general-purpose algorithms vs.\ using more specialized algorithms with data that maintains the
temporal distinctions.

Temporal data is one particular kind of data that can be represented as a relational sequence.  
\shortciteA{kersting2008relational} surveys the area of {\em relational sequence learning} and explains multiple tasks related to such data, such
as sequence mining and alignment.
These tasks often involve the need to identify relevant features or structure, such as identifying frequent patterns or useful similarity functions.
Thus, the set of useful techniques for feature construction and search in this domain overlap with those
discussed in Section~\ref{sec:node-feature}.

\subsection{Privacy Preserving Representation} \label{sec:anon}

There is sometimes a desire to make private graph-based data publicly available (e.g., to support research or public policy)
in a way that preserves the privacy of the individuals described by the data.
The goal of \textit{privacy preserving
representation} is to transform the data in a way that minimizes information loss while maximizing anonymization, e.g., to prevent individuals
in the anonymized network from being identified.
Naive approaches to anonymization operate by
simply replacing an individual's name (or other attributes) with arbitrary and meaningless unique
identifiers.  However, in social networks there are many adversarial methods through which the true identity
of a user can often be discovered from such an anonymized network.
In particular, the adversarial methods can use the network structure and/or remaining attributes to discover the identities of users within the
network \shortcite{Liu08towardsidentity,Zhou08anonsurvey,Narayanan09deanon}.

An early approach by \shortciteA{getoor07anon} examines how a graph may be modified to prevent
sensitive relationships (a particular kind of labeled link) from being disclosed.  They describe
their approach in terms of node anonymization and edge anonymization. Node anonymization
clusters the nodes into $m$ equivalence classes based on node attributes only, while most of the edge anonymization
approaches are based on cleverly removing sensitive edges. \shortciteA{lars07WWW}
address a related family of attacks where an adversary is able to learn whether an edge exists between
targeted pairs of nodes.

More recently, \shortciteA{geromeVLDB08} study privacy issues in graphs that contain no attributes.
Their goal is to prevent ``structural re-identification'' (i.e., identity reconstruction using graph topology information) by anonymizing a graph via
creating an aggregate network model that allows for samples to be drawn from the model. The approach generalizes a graph by
partitioning the nodes and then summarizing the graph at the partition level. 
This approach differs from the other approaches described above because
it drastically changes the representation as opposed to making more incremental changes.
However, this method enforces privacy while still preserving enough of the network properties to allow for a wide variety
of network analyses to be performed.

In each of these investigations the key factors are the information available in the graph, the resources of the attacker, and the type
of attacks that must be defended against.  In addition, if an attacker can possibly obtain additional information related to the
graph from other sources, then the challenges are even more difficult.  More work is needed to provide strong privacy guarantees while still enabling
partial public release of graph-based information.

%% file: conclusion.tex
\section{Conclusion} \label{sec:conclusion}

Given the increasing prevalence and importance of relational data, this article has surveyed the
most significant issues in relational representation transformation.  After presenting a new taxonomy of the
key representation transformation tasks in Section~\ref{sec:overview}, we next discussed the four primary
tasks of link prediction, link interpretation, node prediction, and node interpretation.
Section~\ref{sec:joint-disc} considered how some of these tasks can be accomplished simultaneously via
techniques for joint transformation.  Finally, Section~\ref{sec:discussion} considered how to perform
representation evaluation and key challenges for future work.

The taxonomy presented in Section~\ref{sec:overview} highlighted the symmetry between the possible transformation tasks for links and those for nodes.
This symmetry helped to organize this survey, and also suggests areas where techniques developed for one of these entities can be used for an analogous
task with the other.  For instance, \shortciteA{linkpred:liben07} reformulated traditional node weighting algorithms to 
weight links.  Likewise, topic discovery techniques based on LDA can be used both for node labeling and for link labeling.
Finally, many of the techniques used to create node features can also be used to create link features, and vice versa, although node features
have been studied much more thoroughly.

As discussed in Section~\ref{sec:discussion}, there remains much work to do.  For instance, link
prediction remains a very difficult problem, especially for the general case where any two arbitrary
nodes might be connected together.  Even more significantly, while we have described a wide range of
techniques that can address each of the transformation tasks, at the end of the day the practitioner is
left with a wide range of choices without many guarantees about what might work best.  For instance,
node weighting may improve classification accuracy for one dataset but decrease it on another.  This
challenge is made all the more difficult because the techniques that we have described come from a
wide range of areas, including graph theory, social network analysis, matrix factorization, metric
learning, information theory, information retrieval, inductive logic programming, statistical
relational learning, and probabilistic graphical models.  While the breadth of techniques relevant
to relational transformation is a wonderful resource, it also means that evaluating the representation
change techniques that are relevant to a particular task is a time-consuming, technically
challenging, and incomplete process.  Therefore, much more work is needed to establish a
theoretical understanding of how different representation changes affect the data, how different data
characteristics interact with this process, and how the combination of these techniques and the data
characteristics affect the final results of an analysis with relational data.